\documentclass[letterpaper]{article} 
\usepackage{aaai2026}  
\usepackage{times}  
\usepackage{helvet}  
\usepackage{courier}  
\usepackage[hyphens]{url}  
\usepackage{graphicx} 
\usepackage{natbib}  
\usepackage{caption} 
\usepackage{algorithm}
\usepackage{algorithmic}
\usepackage{booktabs}
\usepackage{tabularx}
\usepackage{multirow}
\usepackage{subcaption} 
\usepackage{xcolor}
\usepackage{amsmath}
\usepackage{array}
\usepackage{amssymb}
\usepackage{colortbl}
\usepackage{tcolorbox}                   
\tcbuselibrary{breakable}                 
\usepackage{enumitem}
\usepackage{lipsum}      
\usepackage{siunitx}
\usepackage[utf8]{inputenc}
\usepackage{newfloat}
\usepackage{listings}
\DeclareCaptionStyle{ruled}{labelfont=normalfont,labelsep=colon,strut=off} 
\floatstyle{ruled}
\newfloat{listing}{tb}{lst}{}
\floatname{listing}{Listing}
\title{LLMTM: Benchmarking and Optimizing LLMs for Temporal Motif Analysis in Dynamic Graphs}

\author {
    Bing Hao\textsuperscript{\rm 1,\rm 2}\footnotemark[1],
    Minglai Shao \textsuperscript{\rm 1,\rm3}\thanks{Equal contribution.},
    Zengyi Wo \textsuperscript{\rm 1},
    Yunlong Chu
    \textsuperscript{\rm 1},
    Yuhang Liu
    \textsuperscript{\rm 1},
    Ruijie Wang
    \textsuperscript{\rm 2}\thanks{Corresponding author.}
}
\affiliations {
    \textsuperscript{\rm 1}School of New Media and Communication, Tianjin University, China\\
    \textsuperscript{\rm 2}School of Computer Science and Technology, Beihang University, China\\
    \textsuperscript{\rm 3}Key Lab of Education Blockchain and Intelligent Technology, Ministry of Education, Guangxi Normal University, China\\
    ruijiew@buaa.edu.cn, 
    \{haobing, shaoml, wozengyi1999, cyl2024245030, liuyuhang\_13\}@tju.edu.cn
}

\usepackage{bibentry}

\begin{document}

\maketitle

\begin{abstract}
\label{sec:abstract}
The widespread application of Large Language Models (LLMs) has motivated a growing interest in their capacity for processing dynamic graphs. Temporal motifs, as an elementary unit and important local property of dynamic graphs which can directly reflect anomalies and unique phenomena, are essential for understanding their evolutionary dynamics and structural features. However, leveraging LLMs for temporal motif analysis on dynamic graphs remains relatively unexplored. In this paper, we systematically study LLM performance on temporal motif-related tasks. Specifically, we propose a comprehensive benchmark, LLMTM (Large Language Models in Temporal Motifs), which includes six tailored tasks across nine temporal motif types. We then conduct extensive experiments to analyze the impacts of different prompting techniques and LLMs (including nine models: openPangu-7B, the DeepSeek-R1-Distill-Qwen series, Qwen2.5-32B-Instruct, GPT-4o-mini, DeepSeek-R1, and o3) on model performance. Informed by our benchmark findings, we develop a tool-augmented LLM agent that leverages precisely engineered prompts to solve these tasks with high accuracy. Nevertheless, the high accuracy of the agent incurs a substantial cost. To address this trade-off, we propose a simple yet effective structure-aware dispatcher that considers both the dynamic graph's structural properties and the LLM's cognitive load to intelligently dispatch queries between the standard LLM prompting and the more powerful agent. Our experiments demonstrate that the structure-aware dispatcher effectively maintains high accuracy while reducing cost.
\end{abstract}

\begin{links}
    \link{Benchmark, Code and Extended version}{https://github.com/Wjerry5/LLMTM}
\end{links}

\section{Introduction}
The success of Large Language Models (LLMs) has motivated exploration into their capabilities on complex structured data, such as web data~\cite{asurveyofLLMandgraph}. A key frontier is the application of LLMs to dynamic graphs, aiming at capturing evolution patterns of temporal graphs. Recent works study the LLMs’ spatial-temporal understanding abilities on dynamic graphs, highlighting the immense potential of LLMs as a new paradigm for dynamic graph analysis. ~\cite{zhang2024llm4dyglargelanguagemodels,huang2025largelanguagemodelsgood}

Temporal motifs, as elementary units reflecting important local properties of dynamic graphs~\cite{motif_in_temporal_network,liu2021temporalnetworkmotifsmodels}, are typically defined as a set of nodes that interact in a specific temporal sequence within a short period of time.
Therefore, temporal motifs play a critical role in revealing the functionality and characterizing the key features of dynamic graphs~\cite{Seshadhri_2013, jha2014pathsamplingfastprovable, pinar2016escapeefficientlycounting5vertex, conf/wsdm/0002C00P17, jain2018fastprovablemethodestimating}. Thus, mining temporal motifs is essential for numerous real-world applications, such as fraud detection~\cite{zhang2025atmgadadaptivetemporalmotif}, friendship prediction~\cite{9525268}, vendor identification~\cite{liu2025temporalmotifsfinancialnetworks}, knowledge graph reasoning~\cite{wang-metahkg,wang-mpkd,wang-metatkgr,liu-2025-hyperkgr,unifying_knowledge,kgreasoning_survey_kdd}, among others. 

Traditional temporal motif detection methods are typically designed for specific motifs and cannot handle diverse motifs in a unified manner~\cite{cai2024efficienttemporalbutterflycounting}. Deep learning-based approaches, which often require supervised training, perform poorly on this task (experiments in Appendix ~\ref{app:deep_learning}). However, the capability of LLMs to solve temporal motif-related problems in dynamic graphs remains underexplored. Different from other existing benchmarks (Table~\ref{tab:benchmark_comparison}), this paper starts by exploring the following research question:

\noindent \textit{\underline{RQ1: Can Large Language Models Solve Temporal Motif }}
\noindent \textit{\underline{
Problems on Dynamic Graphs?}}

\begin{figure*}[t]
    \centering
    \includegraphics[width=0.95\textwidth]{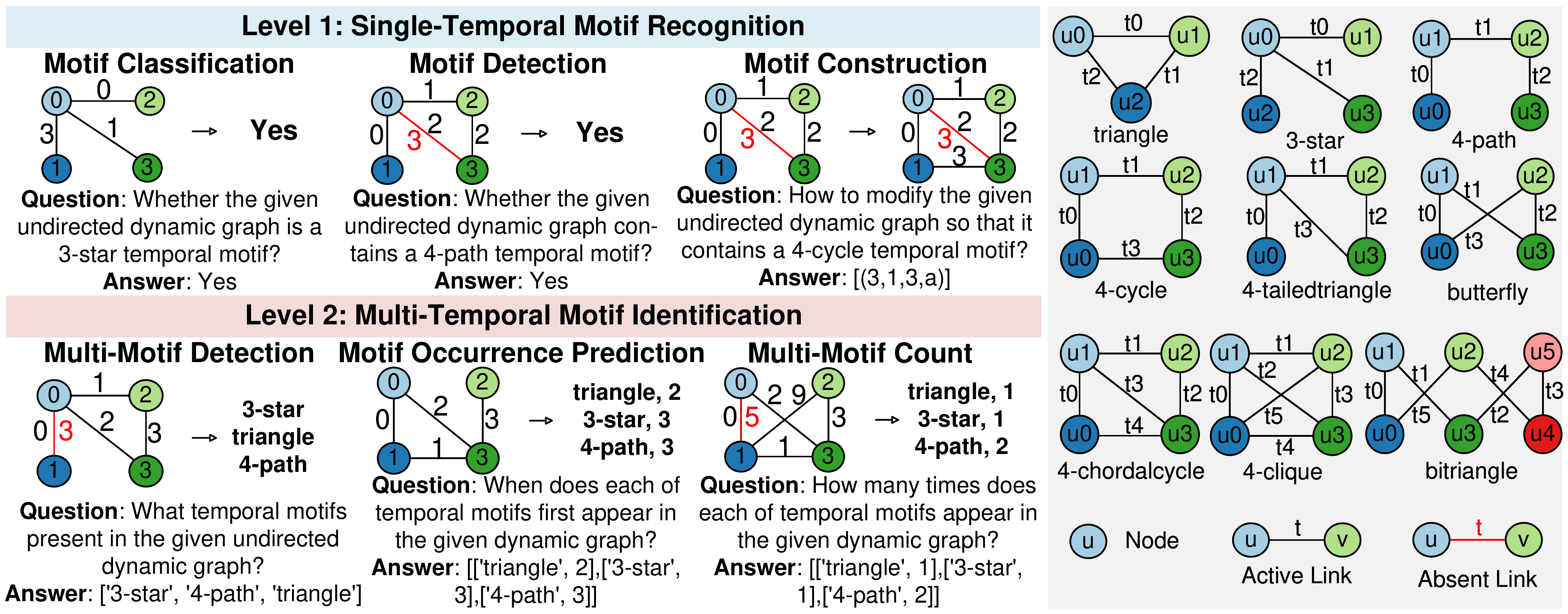}
    \caption{An overview of our LLMTM Benchmark, which includes six tasks (the question is a shortened version due to space) and nine temporal motifs. The tasks (left) are organized into two levels of increasing complexity: (1) single-temporal motif recognition and (2) multi-temporal motif identification. The nine motifs are illustrated on the right.}
    \label{fig:tasks_and_motifs}
\end{figure*}

Addressing this question is non-trivial and presents three key challenges:
\begin{itemize}
    \item How to design a benchmark that can rigorously assess an LLM's understanding and reasoning on temporal motifs?
    \item How to generate dynamic graph datasets with a balanced distribution of positive and negative motif instances for fair evaluation?
    \item How to formulate prompting scheme that can precisely instruct LLMs to understand and process the complex spatio-temporal characteristics of temporal motifs?
\end{itemize}

\begin{table}[h] 
    \centering

    \footnotesize
    \setlength{\tabcolsep}{2.5pt} 
    \begin{tabular}{l c c c c}
        \toprule
        \textbf{Bench.} & 
        \textbf{LLM} & 
        \textbf{\begin{tabular}[c]{@{}c@{}}Dyn.\\Graph\end{tabular}} & 
        \textbf{\begin{tabular}[c]{@{}c@{}}Temp.\\Motif\end{tabular}} & 
        \textbf{\begin{tabular}[c]{@{}c@{}}\#Ev.\\LLMs\end{tabular}} \\
        \midrule

        LLMGP~\cite{dai2025largelanguagemodelsunderstand} 
         & $\checkmark$ & $\times$ & $\times$ & 7 \\

        LLM4DyG~\cite{zhang2024llm4dyglargelanguagemodels} 
         & $\checkmark$ & $\checkmark$ & $\times$ & 5 \\

         TNM~\cite{liu2021temporalnetworkmotifsmodels} 
         & $\times$ & $\checkmark$ & $\checkmark$ & 0 \\
        
        \textbf{LLMTM} (Ours)
        & \textbf{$\checkmark$} & \textbf{$\checkmark$} & \textbf{$\checkmark$} & \textbf{9} \\
        
        \bottomrule
    \end{tabular}
    \caption{A comparison of related benchmarks by their core research areas. Prior work has focused on LLMs for static graphs (LLMGP), general dynamic graph problems (LLM4DyG), or temporal motifs without considering LLMs (TNM). In contrast, our LLMTM benchmark is the first to specifically evaluate the capabilities of LLMs on temporal motifs, thereby filling a critical research gap.}
    \label{tab:benchmark_comparison}
\end{table}

\noindent\textbf{The Benchmark.}  To address these challenges, we introduce \textbf{LLMTM}, a comprehensive benchmark for evaluating LLMs on temporal motif problems (Table~\ref{tab:benchmark_comparison}). Unlike prior work that only considers incremental graph changes~\cite{zhang2024llm4dyglargelanguagemodels}, our benchmark uses a quadruplet representation, $(u, v, t, op)$, to fully capture both edge appearance (add) and disappearance (delete). It comprises six tailored tasks organized into two levels of increasing complexity: (1) single-temporal motif recognition, and (2) multi-temporal motif identification (Figure~\ref{fig:tasks_and_motifs}).

Furthermore, by analyzing the relationship between temporal motifs frequency, time window size, and dynamic graph scale, we determined random dynamic graph generation settings that ensure our datasets are balanced. To comprehensively assess the capabilities of LLMs, we designed a well-defined prompting scheme (Appendix~\ref{app:all_prompts}) and conducted extensive experiments, evaluating four prompting techniques (including zero/one-shot prompting, zero/one-shot chain-of-thought (CoT) prompting ~\cite{wei2023chainofthoughtpromptingelicitsreasoning}) and nine LLMs (including closed-source o3, DeepSeek-R1, GPT-4o-mini and open-source openPangu-7B~\cite{shi2025deepdiveradaptivesearchintensity}, DeepSeek-R1-Distill-Qwen-7B, 14B, 32B~\cite{deepseekai2025deepseekr1incentivizingreasoningcapability}, Qwen2.5-32B-Instruct~\cite{yang2024qwen2technicalreport}, QwQ-32B~\cite{qwen2025qwen25technicalreport}).

Our extensive experiments reveal a key limitation of LLMs. We observe that LLMs perform poorly on complex tasks such as "Motif Detection" and all Level 2 multi-motif tasks, primarily due to excessive cognitive load (The long-context reasoning capability required for LLMs to extract dynamic graph and temporal motif from natural language) (Observations 3 \& 5). This suggests that the reasoning depth of current LLMs remains shallow, and they are likely to fail on problems that require complex, multi-step reasoning.

Tool learning with large language models (LLMs) has emerged as a promising paradigm for augmenting the capabilities of LLMs to tackle highly complex problems ~\cite{Qu_2025}. We further consider:

\noindent\textit{\underline{RQ2: How can agentic capability help to solve temporal}}
\noindent\textit{\underline{motif problems on dynamic graphs?}}

\noindent\textbf{A Tool-Augmented LLM Agent.} Motivated by this, we design a tool-augmented LLM agent, which leverages five algorithms and precisely engineered prompts to solve all six tasks in our benchmark with high accuracy (Table~\ref{tab:Performance of the Tool-augmented LLM agent}). Our experiments show that the proposed agent achieves exceptional performance across all tasks. However, despite high accuracy, the agent incurs greater costs (longer token length and slower response time). This trade-off makes it critical to determine when the agent's intervention is truly necessary, where some simple query can be processed by direct LLM prompt without tool usage. This leads to a key question:

\noindent\textit{\underline{RQ3: How to automatically determine agentic tool usage }}
\noindent\textit{\underline{and direct LLM prompting to balance the trade-off between}}
\noindent\textit{\underline{accuracy and cost?}}

\noindent\textbf{Trade-off Between Accuracy and Cost.} To this goal, we propose the \textbf{structure-aware dispatcher} (Figure~\ref{fig:trade-off}), an agent that strategically routes a given problem to either a standard LLM or a tool-augmented agent. The core of this dispatcher is to predict a problem's intrinsic difficulty using five novel features extracted based on dynamic graph properties and LLM cognition capability: cyclomatic complexity, number of edges, edge locality score, and the proportion of nodes with specific degrees. Empirically, we demonstrate that the structure-aware dispatcher accurately predicts problem difficulty, determines the optimal path to balance accuracy and cost, and exhibits strong generalization to unseen motif-related tasks (Table~\ref{tab:smart dispatcher}).

In summary, our key findings are threefold:

\begin{itemize}
    \item \textbf{LLMs have a performance bottleneck.} While they perform well on simple temporal motif tasks, their performance is bottlenecked by cognitive load on more complex ones. Among mainstream LLMs, DeepSeek-R1 essentially achieves the best performance.
    \item \textbf{Agents are accurate but costly.} The tool-augmented agent solves all temporal motif tasks with high accuracy, but at a significant computational cost.
    \item \textbf{The Structure-Aware Dispatcher effectively balances this trade-off.} Our structure-aware dispatcher strategy works by first predicting a problem's intrinsic difficulty and then intelligently routing it to either the standard LLM or the tool-augmented agent, thereby optimizing the accuracy-cost balance.
\end{itemize}

\section{Formulations and Background}
\subsection{Notations}

\begin{figure*}[t]
    \centering
    \includegraphics[width=0.7\textwidth]{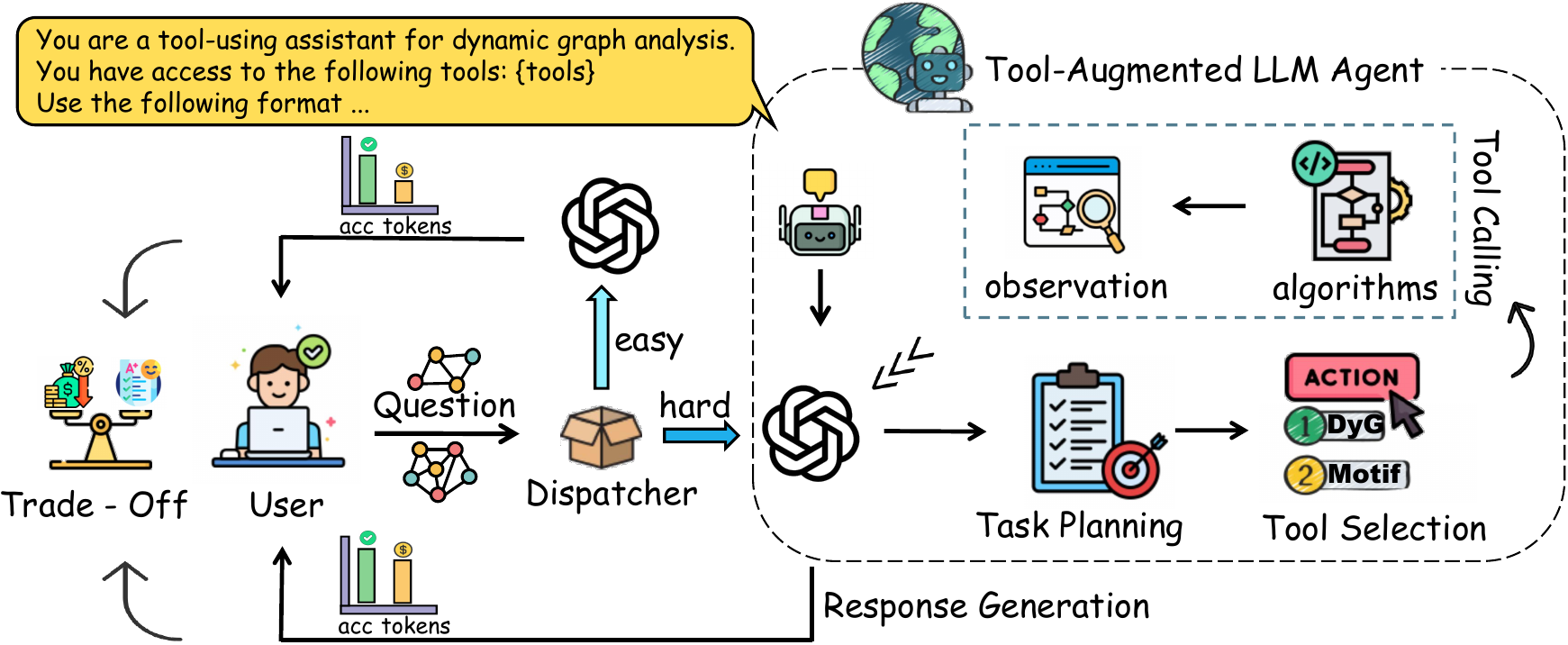}
    \caption{An overview of our framework for balancing the accuracy-cost trade-off. A "Structure-Aware Dispatcher" first extracts the dynamic graph from natural language, predicts the query's difficulty, and then strategically routes simple queries to a standard LLM (for low cost); complex ones to our tool-augmented agent. The agent follows a workflow of task planning, tool selection, tool calling, and response generation to achieve high accuracy, albeit at a greater computational cost.}
    \label{fig:trade-off}
\end{figure*}

\noindent\textbf{Dynamic Graph.} Prior representations of dynamic graphs often use triplet, $(u, v, t)$, that overlooks edge deletions, a common event in real-world networks. To address this limitation, we introduce a quadruplet representation, $(u, v, t, op)$, to fully capture both edge appearance and disappearance. Here, $u$, $v$ are nodes, $t$ is the timestamp, and the operator $op \in \{a, d\}$ denotes an "add" or "delete" operation.

\noindent\textbf{Temporal Motif.} A temporal motif is a sequence of interconnected edge events that form a specific structural pattern within a constrained time duration. Formally, a ($k, l, \delta$)-temporal motif is a time-ordered sequence of $l$ distinct edge events, $M = \{ (u_1, v_1, t_1, a), \dots, (u_l, v_l, t_l, a) \}$, involving $k$ unique nodes that satisfies the following four constraints:
\begin{itemize}
    \item \textbf{Structural Constraint:} The set of static edges corresponding to the events must form a predefined pattern (e.g., a star or a triangle).
    \item \textbf{Temporal Constraint:} The event timestamps must be strictly increasing, i.e., $t_1 < t_2 < \dots < t_l$.
    \item \textbf{Duration Constraint:} The entire sequence must occur within a time window of duration $\delta$, meaning $t_l - t_1 \leq \delta$.
    \item \textbf{Connectivity Constraint:} Consecutive events must be connected, meaning each event $e_{i+1}$ shares at least one node with the set of nodes involved in the preceding events $\{e_1, \dots, e_i\}$.
\end{itemize}

\section{The LLMTM Benchmark}

\subsection{Dataset Construction}
Aiming to create robust benchmarks with balanced positive and negative temporal motif instances, we first analyzed how graph scale ($N$), time span ($T$), and window size ($W$) collectively influence motif distribution to inform our data generation. (Appendix~\ref{app:analysis})

\noindent\textbf{Generation Process.} We first generate a static graph using the Erdős–Rényi (ER) model, $G_0 = \text{ER}(N, p)$, assigning each edge a random timestamp uniformly distributed over the total time span $T$. To model edge dynamics, all edges are initially designated as "add" operations ($a$), with "delete" operations ($d$) subsequently introduced for a random subset. A fixed random seed ensures full reproducibility.

\noindent\textbf{Task-Specific Settings.} For Level 1 tasks, we generate 20 instances per motif type, with two tasks being further refined: "Motif Classification" matches the graph scale to the query motif's size, while "Motif Detection" uses settings that ensure an approximate 50\% presence rate. For Level 2 multi-motif tasks, we generate 100 instances with settings ($N=20, p=0.3, T=15$) chosen to ensure a non-zero probability for all motif types. Detailed settings in Appendix~\ref{app:exp_settings}.

\noindent\textbf{Real-world dataset.} We randomly sample ego-graphs from real graphs for evaluation. Experiments in Appendix ~\ref{app:real-world}.

\subsection{Task Definitions}
\noindent\textbf{Temporal Motifs.} We investigate nine distinct types of temporal motifs: the 3-star, triangle, 4-path, 4-cycle, 4-chordalcycle, 4-tailedtriangle, 4-clique, bitriangle, and butterfly. Figure~\ref{fig:tasks_and_motifs} illustrates the structural patterns and valid temporal orderings for each of these motifs.

\noindent\textbf{Benchmark Tasks.} Building on these nine motifs, we design six tailored tasks to systematically evaluate the capabilities of LLMs on temporal motif problems. As illustrated in Figure~\ref{fig:tasks_and_motifs}, these tasks are organized into two levels of increasing complexity: (1) single-temporal motif recognition and (2) multi-temporal motif identification.

\noindent\textbf{Level 1: Single-Temporal Motif Recognition.}
This level assesses the ability of LLMs to recognize and reason about individual temporal motif instances within a dynamic graph.

\begin{itemize}
    \item \noindent\textbf{Motif Classification.} Requires the model to classify if a given dynamic graph, as a whole, is an exact instance of a temporal motif. The balanced input data includes positive cases and negative cases derived by individually violating the structural, temporal, or duration constraints.

    \item \noindent\textbf{Motif Detection.} Tasks the model with detecting if a dynamic graph contains a specific temporal motif as a subgraph. A positive instance is true if any subset of the graph's events satisfies all four motif constraints.
    
    \item \noindent\textbf{Motif Construction.} Requires the model to complete a nearly-formed temporal motif by adding the single missing edge. The input graph is constructed to contain a pattern missing only its final ($l$-th) edge; the model must identify and output the complete quadruplet for this edge.
\end{itemize}

\noindent\textbf{Level 2: Multi-Temporal Motif Identification.}
This level challenges the LLM's ability to comprehend and resolve problems involving multiple, co-occurring temporal motifs in a dynamic graph.

\begin{itemize}
    \item \noindent\textbf{Multi-Motif Detection.} Requires the model to determine, for each of the nine predefined motif types, whether it is contained as a subgraph in a given dynamic graph. The evaluation uses a fine-grained scoring rule that awards partial credit but penalizes false positives.
    
    \item \noindent\textbf{Motif Occurrence Prediction.} Asks the model to detect all present motif types and report the first occurrence time for each. The time is defined as the timestamp of the final edge completing the motif's first instance. Partial credit is awarded for incomplete but correct responses.
    
    \item \noindent\textbf{Multi-Motif Count.} Tasks the model with detecting and counting the total number of occurrences for each of the nine predefined motif types. For evaluation, the final score is calculated by summing the ratio of the predicted count to the ground-truth count for each motif type.
\end{itemize}

We also conduct a broader evaluation covering four fundamental dynamic graph tasks in the Appendix~\ref{sec:level_0_results}.

\subsection{Prompt and Key Findings}

\noindent\textbf{Prompt Structure.} Our prompts comprise six sequential components: dynamic graph, temporal motif, task-specific , and answer format instructions, exemplars, and the final question. To prevent ambiguity, we use a distinct notation for motifs (e.g., $(u0, u1, t0, a)$), although both dynamic graphs and motifs are represented by quadruplets. (Appendix~\ref{app:all_prompts})

\noindent\textbf{Prompting Strategies.} We investigate the effect of four prompting strategies on model performance: zero-shot, few-shot~\cite{brown2020languagemodelsfewshotlearners}, CoT~\cite{wei2023chainofthoughtpromptingelicitsreasoning}, and a combination of few-shot with CoT. For CoT-based methods, we use custom-designed prompts with reasoning steps carefully tailored to each task.

\noindent\textbf{Bottleneck.} LLMs perform poorly on "Motif Detection" and all Level 2 multi-motif tasks, due to excessive cognitive load (Observations 3 \& 5). Strikingly, some LLMs self-diagnose this limitation, responding that the problems exceed their text-based capabilities and require specialized algorithmic tools. This finding directly motivates our work on the tool-augmented agent presented in the next section.

\section{Exploring Agentic Capability on Temporal Motif-Related Tasks}

Our methodology addresses the limitations of standard LLMs in two stages. We first introduce a tool-augmented agent that, while highly accurate, reveals a significant accuracy-cost trade-off. To resolve this, we then present our primary contribution: the \textbf{Structure-aware Dispatcher}. This predictive framework assesses a problem's intrinsic difficulty using a novel set of metrics that quantify graph structure and LLM cognitive load. It then strategically routes each query to either a standard LLM or the powerful agent, creating a hybrid system that optimally balances accuracy and computational cost with strong generalization.

\subsection{Tool-Augmented LLM Agent}
\begin{table*}[t]
    \centering
    \footnotesize
    \setlength{\tabcolsep}{3pt} 

    \begin{tabular}{@{}lccccccccc@{}}
        \toprule
        Motif Classification & 3-star & triangle & 4-path & 4-cycle & 4-chordalcycle & 4-tailedtriangle & 4-clique & bitriangle & butterfly \\ 
        \midrule
        openPangu-7B          & 55\%             & 60\%             & 70\%             & 70\%             & 55\%              & 75\%              & 65\%             & 70\%             & 60\%             \\ 
        DeepSeek-Qwen-7B                  & 50\%             & 50\%             & 65\%             & 50\%             & 65\%              & 60\%              & 65\%             & 65\%             & 50\%             \\ 
        DeepSeek-R1-Distill-Qwen-14B                 & \textbf{100\%}          & 90\%             & \underline{95\%}            & \textbf{100\%}          & 90\%                 & \textbf{100\%}                 & \textbf{100\%}          & \textbf{100\%}          & \underline{90\%}             \\
        DeepSeek-R1-Distill-Qwen-32B                 & \underline{95\%}            & \underline{95\%}            & \underline{95\%}            & 90\%             & \underline{95\%}                 & \textbf{100\%}                 & \textbf{100\%}          & \textbf{100\%}          & 85\%             \\
        Qwen2.5-32B-Instruct              & \textbf{100\%}          & 80\%             & 90\%             & 90\%             & \underline{95\%}                 & \underline{90\%}                 & 75\%             & 70\%             & \underline{90\%}             \\
        QwQ-32B                  & \textbf{100\%}          & \underline{95\%}            & \underline{95\%}            & 90\%             & \textbf{100\%}          & \underline{90\%}                 & \underline{95\%}            & \underline{90\%}            & 85\%             \\
        GPT-4o-mini              & \textbf{100\%}          & 90\%             & \textbf{100\%}          & 90\%             & \underline{95\%}                 & \underline{90\%}                 & 90\%             & 85\%             & 85\%             \\
        DeepSeek-R1              & \textbf{100\%}          & \underline{95\%}            & \textbf{100\%}          & \underline{95\%}            & 90\%                 & \textbf{100\%}                 & \textbf{100\%}          & \textbf{100\%}          & \textbf{100\%}          \\
        o3                       & \textbf{100\%}          & \textbf{100\%}          & \textbf{100\%}          & \textbf{100\%}          & \underline{95\%}                 & \underline{90\%}                 & 90\%             & \textbf{100\%}          & \textbf{100\%}          \\ 
        
        \midrule
        DeepSeek-Qwen-7B-One-shot         & 60\%             & 50\%             & 65\%             & 50\%             & 65\%              & 60\%              & 35\%             & 50\%             & 55\%             \\
        DeepSeek-Qwen-7B-Zero-shot+CoT    & 50\%             & 50\%             & 65\%             & 50\%             & 65\%              & 60\%              & 65\%             & 65\%             & 50\%             \\
        DeepSeek-Qwen-7B-One-shot+CoT     & 90\%             & \underline{95\%}            & \underline{95\%}            & 65\%             & 65\%              & \underline{90\%}                 & 45\%             & 80\%             & \underline{90\%}             \\ 
        \bottomrule
    \end{tabular}
    \caption{Performance comparison on the "Motif Classification" task against the random baseline. 5-run average results are reported. The best and second-best performance in each column are marked in \textbf{bold} and \underline{underlined}, respectively.}
    \label{tab:judge_is_motif_result}
\end{table*}

To overcome the limitations of standard LLMs, we introduce a tool-augmented agent designed to solve the tasks in our benchmark. It is equipped with a specialized suite of algorithmic tools and guided by a precisely engineered prompt.

\noindent\textbf{Tools.} The agent utilizes a suite of five algorithmic tools for motif reasoning (e.g., Motif\_Detection). The core implementation identifies valid motifs via a two-step process: (1) we use the classic GraphMatcher algorithm~\cite{1323804} to find all subgraph isomorphisms satisfying the \textbf{Structural} and \textbf{Connectivity} constraints. (2) Each resulting topological mapping is then verified against the \textbf{Temporal} and \textbf{Duration} constraints. Detailed in Appendix~\ref{app:algorithms}.

\noindent\textbf{Agent Workflow and Prompting.} Our agent adapts the classical Reason-Act (ReAct) paradigm in a four-stage workflow (Figure~\ref{fig:trade-off}): \textbf{Task Planning} (interpreting the query and system prompt), \textbf{Tool Selection} (choosing a tool and parameters), \textbf{Tool Calling} (executing the tool), \textbf{Response Generation} (synthesizing the output). A key challenge is designing a single, universal prompt that can reliably steer this process. Through iterative refinement, we developed a unified prompt that compels the model to follow: (1) a structured reasoning framework and (2) precise formatting for robust tool calls, as detailed in Appendix~\ref{app:agent_prompt}.

\noindent\textbf{Trade-Off.} While our tool-augmented agent consistently outperforms direct LLM inference, it is costly, consuming, on average, at least three times more tokens, as shown in Table~\ref{tab:Performance of the Tool-augmented LLM agent} and Figure~\ref{fig:Performance of the Tool-augmented LLM agent}. This clear trade-off between the agent's high accuracy and its significant computational cost motivates our subsequent investigation into how to automatically determine agentic tool usage
and direct LLM prompting to balance the trade-off between accuracy and cost.

\subsection{Structure-Aware Dispatcher}
To address the inherent accuracy-cost trade-off in using tool-augmented agents, we introduce our primary contribution: the \textbf{Structure-Aware Dispatcher} (Figure~\ref{fig:trade-off}), an agent designed to strategically route a given problem to either a standard LLM prompting or a tool-augmented agent by first predicting the problem's intrinsic difficulty.

Informed by our LLMTM findings (Observation 3) that an LLM's reasoning is primarily affected by graph structure and cognitive load, we focused on the "Motif Detection" task and designed five novel metrics to form the core of our structure-aware dispatcher. We group these into three conceptual levels: (1) \textbf{Structural Scale}, measured by the number of edges; (2) \textbf{Structural Complexity}, measured by cyclomatic complexity and node degree proportions (ratio\_nodes\_eq\_2 / ratio\_nodes\_ge\_3); and (3) \textbf{LLM Cognitive Load} is the long-context reasoning capability required for LLMs to extract dynamic graph and temporal motif from natural language, measured by a new metric, the edge locality score, which calculates the dispersion of edges in their sequential representation. Detailed formulas in Appendix~\ref{formula}.

Based on the metrics described above, We train a lightweight XGBoost classifier, encapsulated within the Structure-Aware Dispatcher. The classifier is trained on a dataset with varying graph scales and difficulties across five motif types: 3-star, 4-cycle, 4-clique, 4-chordalcycle, and bitriangle. As shown in Table~\ref{tab:smart dispatcher}, our approach is highly effective: the structure-aware dispatcher accurately predicts problem difficulty, determines the optimal path to balance accuracy and cost, and exhibits strong generalization.

\section{Experiments}
\subsection{LLMTM Benchmark (RQ1)}
\begin{table*}[t]
    \centering
    \footnotesize
    \setlength{\tabcolsep}{3pt} 

    \begin{tabular}{@{}lccccccccc@{}}
        \toprule
        Motif Detection & 3-star & triangle & 4-path & 4-cycle & 4-chordalcycle & 4-tailedtriangle & 4-clique & bitriangle & butterfly \\ \midrule
        openPangu-7B         & 25\%             & 35\%             & 15\%             & 20\%             & 20\%              & 20\%              & 35\%             & 30\%             & 10\%             \\ 
        DeepSeek-R1-Distill-Qwen-7B                  & 50\%             & 20\%             & 35\%             & 30\%             & 15\%              & 20\%              & 30\%             & 5\%              & 0\%              \\
        DeepSeek-R1-Distill-Qwen-14B                 & \textbf{90\%}          & 80\%             & 75\%             & 30\%             & 30\%              & 35\%              & 25\%             & 35\%             & 40\%             \\
        DeepSeek-R1-Distill-Qwen-32B                 & \textbf{90\%}          & \underline{85\%}            & \underline{80\%}            & 45\%             & 10\%              & \underline{55\%}             & 40\%             & 35\%             & 35\%             \\
        Qwen2.5-32B-Instruct              & 75\%             & 70\%             & 75\%             & 50\%             & \underline{55\%}             & 45\%              & 20\%             & 45\%             & 35\%             \\
        QwQ-32B                  & \underline{85\%}            & \underline{85\%}            & 70\%             & 50\%             & 5\%               & 25\%              & 5\%              & 10\%             & 15\%             \\
        GPT-4o-mini              & 60\%             & 45\%             & 45\%             & 30\%             & \underline{55\%}             & 30\%              & 45\%             & \textbf{55\%}          & 40\%             \\
        DeepSeek-R1              & \textbf{90\%}          & \textbf{95\%}          & \textbf{90\%}          & \textbf{75\%}          & \textbf{65\%}          & \textbf{85\%}          & \textbf{55\%}          & \textbf{55\%}          & \textbf{80\%}          \\
        o3                       & \textbf{90\%}          & \textbf{95\%}          & \underline{80\%}            & \underline{65\%}            & 20\%              & \underline{55\%}             & \underline{50\%}            & \underline{50\%}            & \underline{55\%}            \\ 
        \bottomrule
    \end{tabular}
    \caption{Performance comparison on the "Motif Detection" task against the random baseline. 5-run average results are reported.}
    \label{tab:Judge_cotain_motif_result}
\end{table*}
We conduct extensive experiments to evaluate the capabilities of LLMs (Large Language Models) on temporal motif problems. Beyond quantitative performance, we also analyze the LLMs' reasoning processes, finding that while they perform well on simple tasks, their performance is bottlenecked by cognitive load on more complex ones.

\noindent\textbf{Models.} We evaluate a diverse set of LLMs, including closed-source models (GPT-4o-mini, DeepSeek-R1, o3) and several open-source models. These include  
openPangu-7B-DeepDiver ~\cite{shi2025deepdiveradaptivesearchintensity}, the DeepSeek-R1-Distill-Qwen series at 7B, 14B, and 32B parameter sizes ~\cite{deepseekai2025deepseekr1incentivizingreasoningcapability}, Qwen2.5-32B-Instruct~\cite{yang2024qwen2technicalreport}, and QwQ-32B~\cite{qwen2025qwen25technicalreport}. Across all experiments, we set the decoding temperature to $\tau=0$ for reproducibility and use accuracy as evaluation metric.

 \noindent\textbf{Abbreviations.} For presentation, in certain locations, we use "Multi Detect", "Motif Occur", and "Multi Count" as shorthand for the Multi-Motif Detection, Motif Occurrence Prediction, and Multi-Motif Count tasks; and the DeepSeek-R1-Distill-Qwen-7B, 14B, and 32B and Qwen2.5-32B-Instruct models are hereafter referred to as the DeepSeek-Qwen-7B, 14B, 32B, and Qwen2.5-32B models, respectively.

\noindent\textbf{Observation 1: LLMs identify motifs via textual pattern matching and are challenged by multiple constraints.} LLM's primary mechanism is textual matching. For motifs with intuitive names (e.g., "4-cycle" or "3-star"), the LLM also leverages its semantic understanding to guide the matching process, leading to higher accuracy on these types. (Table \ref{tab:judge_is_motif_result}). Nevertheless, the requirement to concurrently satisfy four constraints remains the core challenge. Without explicit guidance, the LLM tends to overlook some constraints, leading to errors.

\noindent\textbf{Observation 2: CoT improves performance by providing semantic guidance and enforcing step-by-step reasoning.} We designed CoT prompts that guide the LLM to first describe a motif's structural characteristics in words (activating semantic understanding), and then sequentially check each constraint (pattern, temporal order, and so on). As shown in Table~\ref{tab:judge_is_motif_result}, this strategy is highly effective because it leverages the model's semantic strengths to aid its weaker structural reasoning, while task decomposition mitigates the difficulty of handling multiple constraints simultaneously. Detailed examples in Appendix~\ref{app:cot_prompts}.

\noindent\textbf{Observation 3: Cognitive Load is the key bottleneck for motif detection in complex graphs.} The performance drop from the simple "Motif Classification" task (Table~\ref{tab:judge_is_motif_result}) to the more complex "Motif Detection" task (Table~\ref{tab:Judge_cotain_motif_result}) highlights this limitation. The latter task's core challenge stems from numerous irrelevant ”distractor” edges, notably increasing its cognitive load, thereby degrading detection accuracy.

\noindent\textbf{Observation 4: Generative tasks are far more challenging than discriminative ones.} The "Motif Construction" task highlights this gap. Despite demonstrating a conceptual understanding, models perform poorly because their inability to perform direct, algorithmic construction forces a highly inefficient trial-and-error strategy that fails in larger search spaces, as shown in Table~\ref{tab:motif_modification_performance}.

\begin{table}[h]
    \centering
    \footnotesize
    \setlength{\tabcolsep}{1.2pt} 

    \begin{tabular}{@{}lccccc@{}}
        \toprule
        
        \text{Motif Construction} & 
        \text{\begin{tabular}[c]{@{}c@{}}4-chordal \\ cycle\end{tabular}} & 
        \text{4-clique} & 
        \text{4-cycle} & 
        \text{\begin{tabular}[c]{@{}c@{}}4-tailed \\ triangle\end{tabular}} & 
        \text{bitriangle}  \\ 
        \midrule
        openPangu-7B          & 60\%             & 45\%             & 50\%             & 50\%             & 45\%               \\ 
        DeepSeek-Qwen-7B                  & 5\%                  & 15\%                & 10\%                & 5\%                 & 25\%                 \\
        DeepSeek-Qwen-14B                 & 40\%                 & 65\%                & 70\%                & 35\%                & 80\%             \\
        DeepSeek-Qwen-32B                 & 55\%                 & 65\%                & 80\%            & 60\%                & 50\%                 \\
        Qwen2.5-32B              & 65\%                 & 20\%                & 30\%                & 60\%                & 55\%                 \\
        QwQ-32B                  & 45\%                 & 45\%                & 70\%                & 40\%                & 30\%                 \\
        GPT-4o-mini              & \textbf{90\%}            & \underline{90\%}            & 80\%            & 75\%                & 80\%             \\
        DeepSeek-R1              & \textbf{90\%}            & \textbf{95\%}           & \underline{95\%}            & \textbf{100\%}          & \underline{85\%}             \\
        o3                       & \underline{85\%}             & \textbf{95\%}           & \textbf{100\%}          & \underline{95\%}            & \textbf{90\%}            \\ 
        \bottomrule
    \end{tabular}
    \caption{Performance comparison on "Motif Construction" task across various temporal motifs.} 
    \label{tab:motif_modification_performance}
\end{table}

\noindent\textbf{Observation 5: LLMs exhibit a "Capability Collapse" on multi-motif identification tasks.} On Level 2 tasks, we observe a sharp performance decline. While LLMs understand the high-level procedure—sequentially checking each motif type, they fail due to a compounding cognitive load. This requires simultaneously tracking multiple, independent constraints for all nine motif types, which causes a severe combinatorial explosion and a diffusion of attention. Strikingly, the Qwen2.5-32B-Instruct model self-diagnosed the limitation, stating the "Multi-Motif Count" task required a "specialized algorithm". This metacognitive observation, where the model recognizes that the task exceeds its text-based limitations, directly explains the poor accuracies in Table~\ref{tab:performance_Level_3}.

\begin{table}[h]
    \centering
    \footnotesize
    \setlength{\tabcolsep}{3pt} 

    \begin{tabular}{@{}lccc@{}}
        \toprule
                                  & Multi Detect & Motif Occur & Multi Count \\ \midrule
        openPangu-7B          & 10.71\%             & 0.75\%             & 0.19\%                       \\ 
        DeepSeek-Qwen-7B                   & 11.19\%              & 0.99\%              & 2.00\%               \\
        DeepSeek-Qwen-14B                  & 23.67\%              & 10.20\%             & 6.08\%               \\
        DeepSeek-Qwen-32B                  & 24.91\%              & 8.73\%              & 5.40\%               \\
        Qwen2.5-32B               & 23.88\%              & 11.48\%             & \textbf{19.69\%}         \\
        QwQ-32B                   & 23.19\%              & 4.02\%              & 0.40\%               \\
        GPT-4o-mini               & 18.80\%              & 8.44\%              & 13.07\%              \\
        DeepSeek-R1               & \underline{32.14\%}      & \underline{17.43\%}     & \underline{13.41\%}      \\
        o3                        & \textbf{39.64\%}         & \textbf{25.92\%}        & 1.94\%               \\
        
        \bottomrule
    \end{tabular}
    \caption{Performance comparison on Level 2 tasks.}
    \label{tab:performance_Level_3}
\end{table}

\noindent\textbf{Observation 6: DeepSeek-R1 achieves the best performance.} Across all tasks, DeepSeek-R1 essentially performs best, indicating superior long-range logical reasoning.

\begin{table*}[htbp]
\centering
\footnotesize
\setlength{\tabcolsep}{4pt} 
\begin{tabular}{l cc cc cc cc cc | cc}
\toprule
Backbone & \multicolumn{10}{c}{GPT-4o-mini} & \multicolumn{2}{c}{openPangu-7B} \\
\cmidrule(lr){1-11} \cmidrule(lr){12-13}

Task & \multicolumn{2}{c}{GPT-4o-mini} & \multicolumn{2}{c}{Random} & \multicolumn{2}{c}{Single} & \multicolumn{2}{c}{\textbf{Ours}} & \multicolumn{2}{c}{Agent} & \multicolumn{2}{c}{\textbf{Ours}} \\
\cmidrule(lr){2-3} \cmidrule(lr){4-5} \cmidrule(lr){6-7} \cmidrule(lr){8-9} \cmidrule(lr){10-11} \cmidrule(lr){12-13}
& Acc & Tokens & Acc & Tokens & Acc & Tokens & Acc & Tokens & Acc & Tokens & Acc & Tokens \\
\midrule

3-star & 68.5\% & \textbf{1685} & 80.0\% & 4430 & 84.5\% & \underline{3788} & \underline{88.0\%} & 4165 & \textbf{100.0\%} & 6689 & 78.0\% & 5906 \\
4-cycle & 57.5\% & \textbf{1669} & 75.5\% & 3893 & 88.5\% & \underline{3807} & \underline{92.5\%} & 4002 & \textbf{100.0\%} & 5651 & 77.0\% & 7042 \\
4-clique & 50.0\% & \textbf{1580} & 71.0\% & 3746 & 93.0\% & \underline{3653} & \underline{97.5\%} & 4141 & \textbf{100.0\%} & 5551 & 86.5\% & 7248 \\
bitriangle & 53.0\% & \textbf{1630} & 71.0\% & 3900 & 93.0\% & \underline{3776} & \underline{95.0\%} & 4105 & \textbf{100.0\%} & 5685 & 80.5\% & 7117 \\
4-chordalcycle & 51.0\% & \textbf{1647} & 72.5\% & 3970 & 90.5\% & \underline{3832} & \underline{96.0\%} & 4309 & \textbf{100.0\%} & 5835 & 86.0\% & 7452 \\
\midrule
4-tailedtriangle & 53.0\% & \textbf{1642} & 75.0\% & 3843 & 85.5\% & 3978 & \underline{88.6\%} & \underline{3705} & \textbf{99.8\%} & 5883 & 84.0\% & 6713 \\
triangle & 60.0\% & \textbf{1652} & 77.8\% & 3864 & \underline{86.5\%} & 3715 & 84.8\% & \underline{3405} & \textbf{100.0\%} & 5932 & 79.80\% & 6206 \\
\bottomrule
\end{tabular}
\caption{Performance comparison of the Structure-Aware Dispatcher against several baselines. "Random" refers to randomly assigning instances, "Single" refers to training and applying on the same temporal motif. Detailed results of the Structure-Aware Dispatcher using openPangu-7B as the LLM in the Appendix~\ref{app:pangu-dispatcher}. This highlights how our methods consistently achieve the second-best accuracy while being significantly more cost-effective than the top-performing "Agent". We held out the "4-tailedtriangle" and "triangle" motifs from the training set to specifically evaluate generalization.}
\label{tab:smart dispatcher}
\end{table*}

\begin{table}[t]
    \centering
    \small
    \setlength{\tabcolsep}{3pt} 

    \begin{tabular}{lcccc} 
        \toprule
        \multirow{2}{*}{Tasks} & \multicolumn{2}{c}{Agent} & \multicolumn{2}{c}{GPT-4o-mini} \\
        \cmidrule(lr){2-3} \cmidrule(lr){4-5}
                               & Acc. & Avg. Tokens & Acc & Avg. Tokens \\
        \midrule
        Multi Detect      & \textbf{98\%}  & 12298.63     & 18.80\%  & \textbf{2716.59} \\
        Motif Occur  & \textbf{100\%} & 11778.15     & 8.44\% & \textbf{3023.36}  \\
        Multi Count       & \textbf{99\%}  & 9190.19      & 13.07\% & \textbf{2790.10} \\
        \bottomrule
    \end{tabular}
    \caption{Performance comparison between our tool-augmented agent and a baseline LLM. We report accuracy (Acc.) and average token consumption (Avg. Tokens) as a measure of cost. The best accuracy and lowest token count in each row are marked in \textbf{bold}.}
    \label{tab:Performance of the Tool-augmented LLM agent}
\end{table}

\begin{figure}[htbp]
    \centering
    \begin{subfigure}{0.48\columnwidth}
        \centering
        \includegraphics[width=\linewidth]{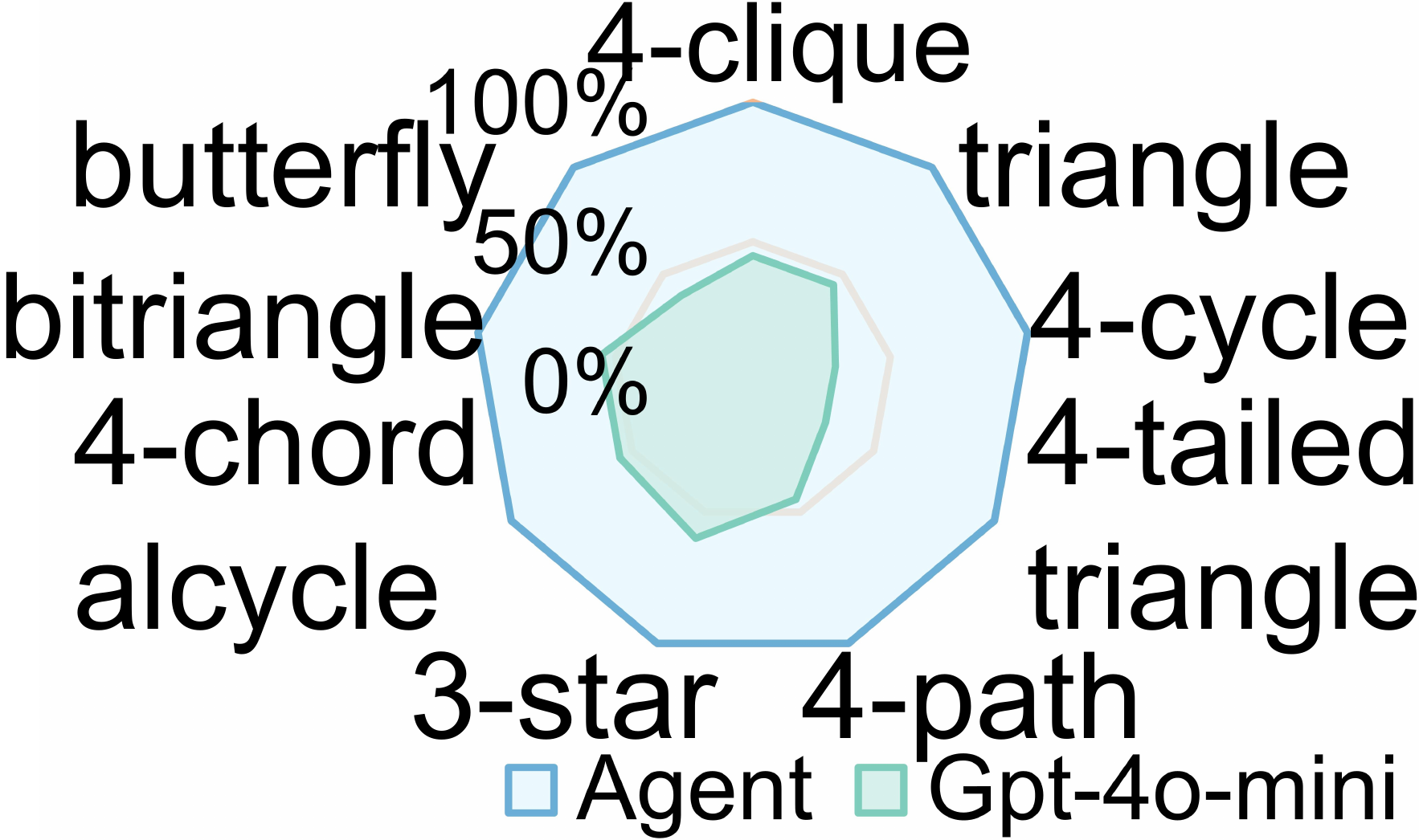} 
        \caption{accuracy}
        \label{fig:sub1_final}
    \end{subfigure}
    \hfill 
    \begin{subfigure}{0.48\columnwidth}
        \centering
        \includegraphics[width=\linewidth]{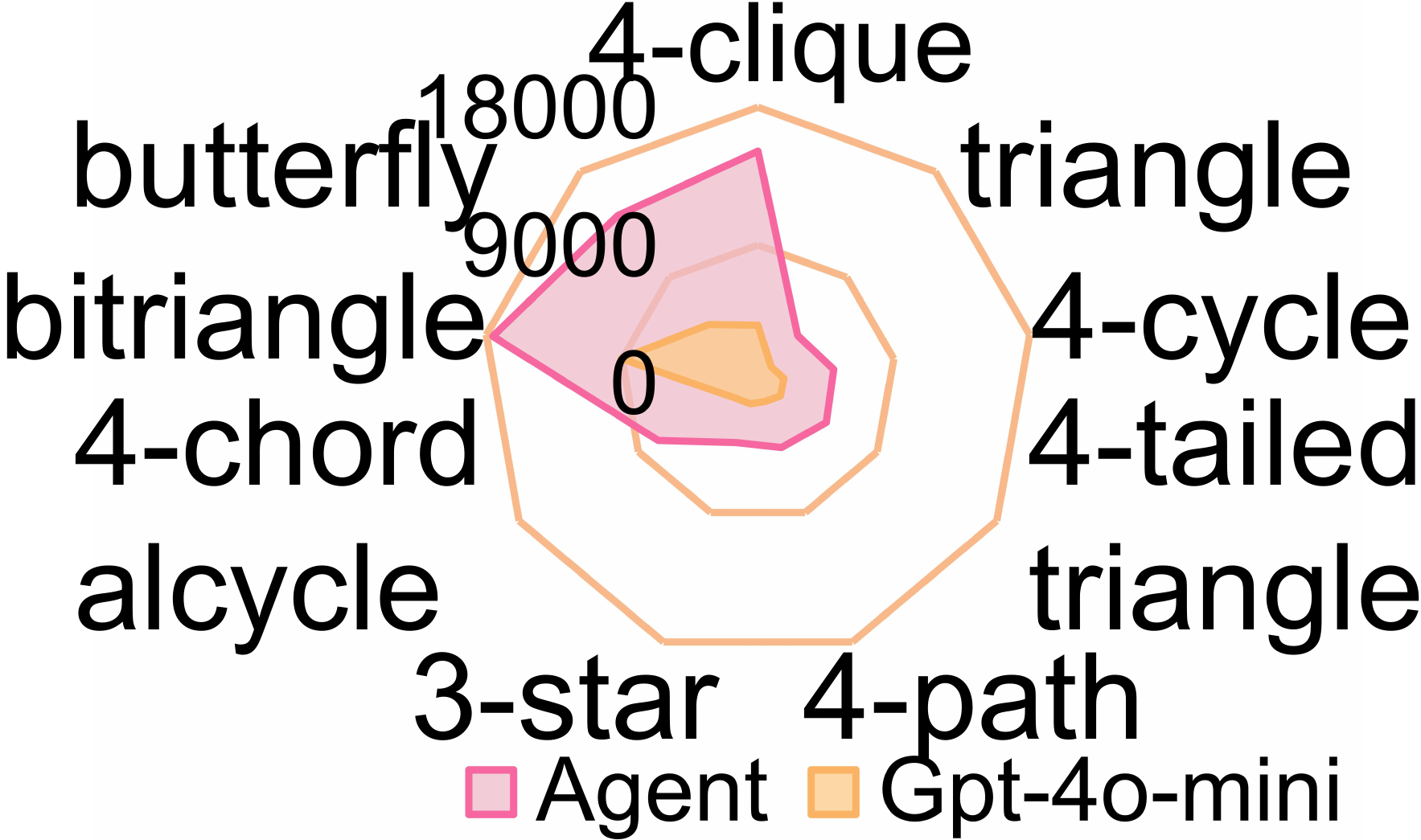} 
        \caption{average token consumption}
        \label{fig:sub2_final}
    \end{subfigure}
    
    \caption{Performance of the tool-augmented Agent versus GPT-4o-mini on the "Motif Detection" task, comparing (a) accuracy and (b) average token consumption. See Appendix~\ref{app:radas_graph} for results on all other tasks.}
    \label{fig:Performance of the Tool-augmented LLM agent}
\end{figure}

\noindent\textbf{Observation 7: openPangu-7B and DeepSeek-R1-Distill-Qwen-7B perform comparably.} The performance of openPangu-7B and DeepSeek-R1-Distill-Qwen-7B tends to be consistent, except on Motif Construction. Interestingly, on this task, openPangu-7B "smartly" outputs a complete motif, artificially inflating its performance.

Notes that we present selected experimental results in this section. The full results are available in Appendix~\ref{app:full_results}.

\subsection{Tool-Augmented LLM Agent (RQ2)}
We compare our tool-augmented LLM agent against the baseline GPT-4o-mini on the six LLMTM tasks in Table~\ref{tab:Performance of the Tool-augmented LLM agent} and Figure~\ref{fig:Performance of the Tool-augmented LLM agent}, which shows that while the agent solves temporal motif tasks with high accuracy, it incurs a significant computational cost.

\subsection{Structure-Aware Dispatcher(RQ3)}
We train a lightweight XGBoost classifier using a dataset of 1500 instances with varying scales and difficulties across five motif types (e.g., 3-star, 4-cycle), which tools LLMs to form an Agent acting as a \textbf{"Structure-Aware Dispatcher"}. Evaluated on a test set of 200 instances per motif, structure-aware dispatcher proves highly cost-effective, significantly outperforming baselines. To further assess its generalization, we created new test sets for two unseen motif types ("4-tailedtriangle" and "triangle"), each containing 500 instances. As shown in Table~\ref{tab:smart dispatcher}, the structure-aware dispatcher remains effective at balancing accuracy and cost even on these novel motifs, underscoring its strong generalizability.

\section{Related Work}
\noindent\textbf{LLMs on Dynamic Graphs.} The application of Large Language Models (LLMs) to dynamic graphs began with feasibility studies: for instance, LLM4DyG~\cite{zhang2024llm4dyglargelanguagemodels} benchmarked the spatio-temporal understanding of LLMs and GraphArena ~\cite{tang2025grapharenaevaluatingexploringlarge} benchmarked the graph computation of LLMs. Subsequent work developed more specialized frameworks: LLM-DA~\cite{wang2024largelanguagemodelsguideddynamic} extracts adaptable temporal logic rules; TGL-LLM~\cite{chang2025integratetemporalgraphlearning} aligns graph and language representations; and All in One ~\cite{sun2023onemultitaskpromptinggraph} leans towards machine learning.

\noindent\textbf{Temporal Motif Discovery.} Temporal motifs, critical local property of dynamic graphs, have a long history of study in fields from biology~\cite{Chechik2008-nv, faisal2013dynamicnetworksrevealkey} to social science~\cite{PhysRevE.91.052813, kosyfaki2018flowmotifsinteractionnetworks}, and possess classical algorithms for motif discovery and efficient counting~\cite{8691398, Sun2019NewAF, Liu2019SamplingMF}. 
More recently, deep learning methods have leveraged motifs for specific tasks, exemplified by a GCN-based approach that fuses local motif information with global properties for fraud detection~\cite{li2023motifawaretemporalgcnfraud}.

\noindent\textbf{Tool Learning.} Tool learning has emerged as a promising paradigm to augment LLMs for complex problems~\cite{Qu_2025}. Studies show that LLMs can effectively use tools via prompting without fine-tuning~\cite{paranjape2023artautomaticmultistepreasoning, zhang2023graphtoolformerempowerllmsgraph, li2024stridetoolassistedllmagent}. For example, HuggingGPT~\cite{shen2023hugginggptsolvingaitasks} leverages a sophisticated prompt design, while TPTU~\cite{ruan2023tptulargelanguagemodelbased} introduces a structured framework with one-step and sequential agents.

\section{Conclusion}
In this paper, we introduced the LLMTM benchmark to systematically evaluate the capabilities of Large Language Models (LLMs) on temporal motif problems in dynamic graphs, a previously unexplored domain. The benchmark comprises six tasks across two levels of increasing complexity: single-temporal motif recognition and multi-temporal motif identification. Our extensive experiments yielded seven fine-grained observations, illuminating the core reasoning mechanisms and performance bottlenecks of LLMs.

Building on these insights, we demonstrated that a tool-augmented agent can achieve strong performance across all benchmark tasks. To address the significant cost of this approach, we then introduced our primary contribution: a \textbf{Structure-Aware Dispatcher}. This framework uses a novel set of metrics to predict a problem's intrinsic difficulty, enabling the selective deployment of the powerful but expensive agent. Our results show that this method successfully balances the accuracy-cost trade-off and exhibits strong generalization.

\newpage
\section*{Acknowledgments}
This work is partially supported by NSFC program (No. 62272338) and Research Fund of Key Lab of Education Blockchain and Intelligent Technology, Ministry of Education (EBME25-F-06). Ruijie Wang is supported by the Fundamental Research Funds for the Central Universities.

\bibliography{reference}

@misc{asurveyofLLMandgraph,
      title={Advancing Graph Representation Learning with Large Language Models: A Comprehensive Survey of Techniques}, 
      author={Qiheng Mao and Zemin Liu and Chenghao Liu and Zhuo Li and Jianling Sun},
      year={2024},
      eprint={2402.05952},
      archivePrefix={arXiv},
      primaryClass={cs.LG},
      url={https://arxiv.org/abs/2402.05952}, 
}

@misc{zhang2024llm4dyglargelanguagemodels,
      title={LLM4DyG: Can Large Language Models Solve Spatial-Temporal Problems on Dynamic Graphs?}, 
      author={Zeyang Zhang and Xin Wang and Ziwei Zhang and Haoyang Li and Yijian Qin and Wenwu Zhu},
      year={2024},
      eprint={2310.17110},
      archivePrefix={arXiv},
      primaryClass={cs.LG},
      url={https://arxiv.org/abs/2310.17110}, 
}

@misc{huang2025largelanguagemodelsgood,
      title={Are Large Language Models Good Temporal Graph Learners?}, 
      author={Shenyang Huang and Ali Parviz and Emma Kondrup and Zachary Yang and Zifeng Ding and Michael Bronstein and Reihaneh Rabbany and Guillaume Rabusseau},
      year={2025},
      eprint={2506.05393},
      archivePrefix={arXiv},
      primaryClass={cs.CL},
      url={https://arxiv.org/abs/2506.05393}, 
}

@inproceedings{motif_in_temporal_network, series={WSDM 2017},
   title={Motifs in Temporal Networks},
   url={http://dx.doi.org/10.1145/3018661.3018731},
   DOI={10.1145/3018661.3018731},
   booktitle={Proceedings of the Tenth ACM International Conference on Web Search and Data Mining},
   publisher={ACM},
   author={Paranjape, Ashwin and Benson, Austin R. and Leskovec, Jure and et al.},
   year={2017},
   month=feb, pages={601–610},
   collection={WSDM 2017} }

@misc{liu2021temporalnetworkmotifsmodels,
      title={Temporal Network Motifs: Models, Limitations, Evaluation}, 
      author={Penghang Liu and Valerio Guarrasi and A. Erdem Sarıyüce and et al.},
      year={2021},
      eprint={2005.11817},
      archivePrefix={arXiv},
      primaryClass={cs.SI},
      url={https://arxiv.org/abs/2005.11817}, 
}

@misc{liu2025temporalmotifsfinancialnetworks,
      title={Temporal Motifs for Financial Networks: A Study on Mercari, JPMC, and Venmo Platforms}, 
      author={Penghang Liu and Bahadir Altun and Rupam Acharyya and Robert E. Tillman and Shunya Kimura and Naoki Masuda and Ahmet Erdem Sarıyüce},
      year={2025},
      eprint={2301.07791},
      archivePrefix={arXiv},
      primaryClass={cs.SI},
      url={https://arxiv.org/abs/2301.07791}, 
}

@inproceedings{Seshadhri_2013,
   title={Triadic Measures on Graphs: The Power of Wedge Sampling},
   url={http://dx.doi.org/10.1137/1.9781611972832.2},
   DOI={10.1137/1.9781611972832.2},
   booktitle={Proceedings of the 2013 SIAM International Conference on Data Mining},
   publisher={Society for Industrial and Applied Mathematics},
   author={Seshadhri, C. and Pinar, Ali and Kolda, Tamara G. and et al.},
   year={2013},
   month=may, pages={10–18} }

@misc{jha2014pathsamplingfastprovable,
      title={Path Sampling: A Fast and Provable Method for Estimating 4-Vertex Subgraph Counts}, 
      author={Madhav Jha and C. Seshadhri and Ali Pinar and et al.},
      year={2014},
      eprint={1411.4942},
      archivePrefix={arXiv},
      primaryClass={cs.DS},
      url={https://arxiv.org/abs/1411.4942}, 
}

@misc{pinar2016escapeefficientlycounting5vertex,
      title={ESCAPE: Efficiently Counting All 5-Vertex Subgraphs}, 
      author={Ali Pinar and C. Seshadhri and V. Vishal and et al.},
      year={2016},
      eprint={1610.09411},
      archivePrefix={arXiv},
      primaryClass={cs.SI},
      url={https://arxiv.org/abs/1610.09411}, 
}

@inproceedings{conf/wsdm/0002C00P17,
  author = {Bressan, Marco and Chierichetti, Flavio and Kumar, Ravi and Leucci, Stefano and Panconesi, Alessandro},
  booktitle = {WSDM},
  editor = {de Rijke, Maarten and Shokouhi, Milad and Tomkins, Andrew and Zhang, Min},
  ee = {https://doi.org/10.1145/3018661.3018732},
  isbn = {978-1-4503-4675-7},
  pages = {557-566},
  publisher = {ACM},
  title = {Counting Graphlets: Space vs Time.},
  url = {http://dblp.uni-trier.de/db/conf/wsdm/wsdm2017.html#0002C00P17},
  year = 2017
}

@misc{jain2018fastprovablemethodestimating,
      title={A Fast and Provable Method for Estimating Clique Counts Using Tur\'an's Theorem}, 
      author={Shweta Jain and C. Seshadhri},
      year={2018},
      eprint={1611.05561},
      archivePrefix={arXiv},
      primaryClass={cs.SI},
      url={https://arxiv.org/abs/1611.05561}, 
}

@misc{wei2023chainofthoughtpromptingelicitsreasoning,
      title={Chain-of-Thought Prompting Elicits Reasoning in Large Language Models}, 
      author={Jason Wei and Xuezhi Wang and Dale Schuurmans and Maarten Bosma and Brian Ichter and Fei Xia and Ed Chi and Quoc Le and Denny Zhou},
      year={2023},
      eprint={2201.11903},
      archivePrefix={arXiv},
      primaryClass={cs.CL},
      url={https://arxiv.org/abs/2201.11903}, 
}

@misc{deepseekai2025deepseekr1incentivizingreasoningcapability,
      title={DeepSeek-R1: Incentivizing Reasoning Capability in LLMs via Reinforcement Learning}, 
      author={DeepSeek-AI and Daya Guo and Dejian Yang and Haowei Zhang and Junxiao Song and Ruoyu Zhang and Runxin Xu and Qihao Zhu and Shirong Ma and Peiyi Wang and Xiao Bi and Xiaokang Zhang and Xingkai Yu and Yu Wu and Z. F. Wu and Zhibin Gou and Zhihong Shao and Zhuoshu Li and Ziyi Gao and Aixin Liu and Bing Xue and Bingxuan Wang and Bochao Wu and Bei Feng},
      year={2025},
      eprint={2501.12948},
      archivePrefix={arXiv},
      primaryClass={cs.CL},
      url={https://arxiv.org/abs/2501.12948}, 
}

@misc{yang2024qwen2technicalreport,
      title={Qwen2 Technical Report}, 
      author={An Yang and Baosong Yang and Binyuan Hui and Bo Zheng and Bowen Yu and Chang Zhou and Chengpeng Li and Chengyuan Li and Dayiheng Liu and Fei Huang and Guanting Dong and Haoran Wei and Huan Lin and Jialong Tang and Jialin Wang and Jian Yang and Jianhong Tu and Jianwei Zhang and Jianxin Ma and Jianxin Yang and Jin Xu and Jingren Zhou and Jinze Bai and Jinzheng He and Junyang Lin and Kai Dang and Keming Lu and Keqin Chen and Kexin Yang and Mei Li and Mingfeng Xue and Na Ni and Pei Zhang and Peng Wang and Ru Peng and Rui Men and Ruize Gao and Runji Lin and Shijie Wang and Shuai Bai and Sinan Tan and Tianhang Zhu and Tianhao Li and Tianyu Liu and Wenbin Ge and Xiaodong Deng and Xiaohuan Zhou and Xingzhang Ren and Xinyu Zhang and Xipin Wei and Xuancheng Ren and Xuejing Liu and Yang Fan and Yang Yao and Yichang Zhang and Yu Wan and Yunfei Chu and Yuqiong Liu and Zeyu Cui and Zhenru Zhang and Zhifang Guo and Zhihao Fan},
      year={2024},
      eprint={2407.10671},
      archivePrefix={arXiv},
      primaryClass={cs.CL},
      url={https://arxiv.org/abs/2407.10671}, 
}

@misc{qwen2025qwen25technicalreport,
      title={Qwen2.5 Technical Report}, 
      author={Qwen and An Yang and Baosong Yang and Beichen Zhang and Binyuan Hui and Bo Zheng and Bowen Yu and Chengyuan Li and Dayiheng Liu and Fei Huang and Haoran Wei and Huan Lin and Jian Yang and Jianhong Tu and Jianwei Zhang and Jianxin Yang and Jiaxi Yang and Jingren Zhou and Junyang Lin and Kai Dang and Keming Lu and Keqin Bao and Kexin Yang and Le Yu and Mei Li and Mingfeng Xue and Pei Zhang and Qin Zhu and Rui Men and Runji Lin and Tianhao Li and Tianyi Tang and Tingyu Xia and Xingzhang Ren and Xuancheng Ren and Yang Fan and Yang Su and Yichang Zhang and Yu Wan and Yuqiong Liu and Zeyu Cui and Zhenru Zhang and Zihan Qiu},
      year={2025},
      eprint={2412.15115},
      archivePrefix={arXiv},
      primaryClass={cs.CL},
      url={https://arxiv.org/abs/2412.15115}, 
}

@article{Qu_2025,
   title={Tool learning with large language models: a survey},
   volume={19},
   ISSN={2095-2236},
   url={http://dx.doi.org/10.1007/s11704-024-40678-2},
   DOI={10.1007/s11704-024-40678-2},
   number={8},
   journal={Frontiers of Computer Science},
   publisher={Springer Science and Business Media LLC},
   author={Qu, Changle and Dai, Sunhao and Wei, Xiaochi and Cai, Hengyi and Wang, Shuaiqiang and Yin, Dawei and Xu, Jun and Wen, Ji-rong},
   year={2025},
   month=jan }

@misc{chang2025integratetemporalgraphlearning,
      title={Integrate Temporal Graph Learning into LLM-based Temporal Knowledge Graph Model}, 
      author={He Chang and Jie Wu and Zhulin Tao and Yunshan Ma and Xianglin Huang and Tat-Seng Chua},
      year={2025},
      eprint={2501.11911},
      archivePrefix={arXiv},
      primaryClass={cs.IR},
      url={https://arxiv.org/abs/2501.11911}, 
}

@misc{wang2024largelanguagemodelsguideddynamic,
      title={Large Language Models-guided Dynamic Adaptation for Temporal Knowledge Graph Reasoning}, 
      author={Jiapu Wang and Kai Sun and Linhao Luo and Wei Wei and Yongli Hu and Alan Wee-Chung Liew and Shirui Pan and Baocai Yin},
      year={2024},
      eprint={2405.14170},
      archivePrefix={arXiv},
      primaryClass={cs.AI},
      url={https://arxiv.org/abs/2405.14170}, 
}

@misc{faisal2013dynamicnetworksrevealkey,
      title={Dynamic networks reveal key players in aging}, 
      author={Fazle Elahi Faisal and Tijana Milenkovic},
      year={2013},
      eprint={1307.3388},
      archivePrefix={arXiv},
      primaryClass={cs.CE},
      url={https://arxiv.org/abs/1307.3388}, 
}

@article{Chechik2008-nv,
  author = {Chechik, Gal and Oh, Eugene and Rando, Oliver and Weissman, Jonathan and Regev, Aviv and Koller, Daphne},
  journal = {Nat. Biotechnol.},
  pages = {1251--1259},
  title = {Activity motifs reveal principles of timing in transcriptional control of the yeast metabolic network},
  volume = 26,
  year = 2008
}

@article{PhysRevE.91.052813,
  title = {Temporal motifs reveal collaboration patterns in online task-oriented networks},
  author = {Xuan, Qi and Fang, Huiting and Fu, Chenbo and Filkov, Vladimir},
  journal = {Phys. Rev. E},
  volume = {91},
  issue = {5},
  pages = {052813},
  numpages = {9},
  year = {2015},
  month = {May},
  publisher = {American Physical Society},
  doi = {10.1103/PhysRevE.91.052813},
  url = {https://link.aps.org/doi/10.1103/PhysRevE.91.052813}
}

@misc{kosyfaki2018flowmotifsinteractionnetworks,
      title={Flow Motifs in Interaction Networks}, 
      author={Chrysanthi Kosyfaki and Nikos Mamoulis and Evaggelia Pitoura and Panayiotis Tsaparas},
      year={2018},
      eprint={1810.08408},
      archivePrefix={arXiv},
      primaryClass={cs.SI},
      url={https://arxiv.org/abs/1810.08408}, 
}

@ARTICLE{8691398,
  author={Sun, Xiaoli and Tan, Yusong and Wu, Qingbo and Chen, Baozi and Shen, Changxiang},
  journal={IEEE Access}, 
  title={TM-Miner: TFS-Based Algorithm for Mining Temporal Motifs in Large Temporal Network}, 
  year={2019},
  volume={7},
  number={},
  pages={49778-49789},
  doi={10.1109/ACCESS.2019.2911181}}

@article{Sun2019NewAF,
  title={New Algorithms for Counting Temporal Graph Pattern},
  author={Xiaoli Sun and Yusong Tan and Qingbo Wu and Jing Wang and Changxiang Shen},
  journal={Symmetry},
  year={2019},
  volume={11},
  pages={1188},
  url={https://api.semanticscholar.org/CorpusID:204194356}
}

@article{Liu2019SamplingMF,
  title={Sampling Methods for Counting Temporal Motifs},
  author={Paul Liu and Austin R and et al. Benson and Moses Charikar},
  journal={Proceedings of the Twelfth ACM International Conference on Web Search and Data Mining},
  year={2019},
  url={https://api.semanticscholar.org/CorpusID:59281321}
}

@misc{li2023motifawaretemporalgcnfraud,
      title={Motif-aware temporal GCN for fraud detection in signed cryptocurrency trust networks}, 
      author={Song Li and Jiandong Zhou and Chong MO and Jin LI and Geoffrey K. F. Tso and Yuxing Tian},
      year={2023},
      eprint={2211.13123},
      archivePrefix={arXiv},
      primaryClass={cs.LG},
      url={https://arxiv.org/abs/2211.13123}, 
}

@misc{paranjape2023artautomaticmultistepreasoning,
      title={ART: Automatic multi-step reasoning and tool-use for large language models}, 
      author={Bhargavi Paranjape and Scott Lundberg and Sameer Singh and Hannaneh Hajishirzi and Luke Zettlemoyer and Marco Tulio Ribeiro},
      year={2023},
      eprint={2303.09014},
      archivePrefix={arXiv},
      primaryClass={cs.CL},
      url={https://arxiv.org/abs/2303.09014}, 
}

@misc{zhang2023graphtoolformerempowerllmsgraph,
      title={Graph-ToolFormer: To Empower LLMs with Graph Reasoning Ability via Prompt Augmented by ChatGPT}, 
      author={Jiawei Zhang},
      year={2023},
      eprint={2304.11116},
      archivePrefix={arXiv},
      primaryClass={cs.AI},
      url={https://arxiv.org/abs/2304.11116}, 
}

@misc{li2024stridetoolassistedllmagent,
      title={STRIDE: A Tool-Assisted LLM Agent Framework for Strategic and Interactive Decision-Making}, 
      author={Chuanhao Li and Runhan Yang and Tiankai Li and Milad Bafarassat and Kourosh Sharifi and Dirk Bergemann and Zhuoran Yang},
      year={2024},
      eprint={2405.16376},
      archivePrefix={arXiv},
      primaryClass={cs.CL},
      url={https://arxiv.org/abs/2405.16376}, 
}

@misc{shen2023hugginggptsolvingaitasks,
      title={HuggingGPT: Solving AI Tasks with ChatGPT and its Friends in Hugging Face}, 
      author={Yongliang Shen and Kaitao Song and Xu Tan and Dongsheng Li and Weiming Lu and Yueting Zhuang},
      year={2023},
      eprint={2303.17580},
      archivePrefix={arXiv},
      primaryClass={cs.CL},
      url={https://arxiv.org/abs/2303.17580}, 
}

@misc{ruan2023tptulargelanguagemodelbased,
      title={TPTU: Large Language Model-based AI Agents for Task Planning and Tool Usage}, 
      author={Jingqing Ruan and Yihong Chen and Bin Zhang and Zhiwei Xu and Tianpeng Bao and Guoqing Du and Shiwei Shi and Hangyu Mao and Ziyue Li and Xingyu Zeng and Rui Zhao},
      year={2023},
      eprint={2308.03427},
      archivePrefix={arXiv},
      primaryClass={cs.AI},
      url={https://arxiv.org/abs/2308.03427}, 
}

@misc{brown2020languagemodelsfewshotlearners,
      title={Language Models are Few-Shot Learners}, 
      author={Tom B. Brown and Benjamin Mann and Nick Ryder and Melanie Subbiah and Jared Kaplan and Prafulla Dhariwal and Arvind Neelakantan and Pranav Shyam and Girish Sastry and Amanda Askell and Sandhini Agarwal and Ariel Herbert-Voss and Gretchen Krueger and Tom Henighan and Rewon Child and Aditya Ramesh and Daniel M. Ziegler and Jeffrey Wu and Clemens Winter and Christopher Hesse and Mark Chen and Eric Sigler and Mateusz Litwin and Scott Gray and Benjamin Chess and Jack Clark and Christopher Berner and Sam McCandlish and Alec Radford and Ilya Sutskever and Dario Amodei},
      year={2020},
      eprint={2005.14165},
      archivePrefix={arXiv},
      primaryClass={cs.CL},
      url={https://arxiv.org/abs/2005.14165}, 
}

@misc{dai2025largelanguagemodelsunderstand,
      title={How Do Large Language Models Understand Graph Patterns? A Benchmark for Graph Pattern Comprehension}, 
      author={Xinnan Dai and Haohao Qu and Yifen Shen and Bohang Zhang and Qihao Wen and Wenqi Fan and Dongsheng Li and Jiliang Tang and Caihua Shan},
      year={2025},
      eprint={2410.05298},
      archivePrefix={arXiv},
      primaryClass={cs.LG},
      url={https://arxiv.org/abs/2410.05298}, 
}

@ARTICLE{1323804,
  author={Cordella, L.P. and Foggia, P. and Sansone, C. and Vento, M.},
  journal={IEEE Transactions on Pattern Analysis and Machine Intelligence}, 
  title={A (sub)graph isomorphism algorithm for matching large graphs}, 
  year={2004},
  volume={26},
  number={10},
  pages={1367-1372},
  doi={10.1109/TPAMI.2004.75}}

@misc{tang2025grapharenaevaluatingexploringlarge,
      title={GraphArena: Evaluating and Exploring Large Language Models on Graph Computation}, 
      author={Jianheng Tang and Qifan Zhang and Yuhan Li and Nuo Chen and Jia Li},
      year={2025},
      eprint={2407.00379},
      archivePrefix={arXiv},
      primaryClass={cs.AI},
      url={https://arxiv.org/abs/2407.00379}, 
}

@misc{shi2025deepdiveradaptivesearchintensity,
      title={DeepDiver: Adaptive Search Intensity Scaling via Open-Web Reinforcement Learning}, 
      author={Wenxuan Shi and Haochen Tan and Chuqiao Kuang and Xiaoguang Li and Xiaozhe Ren and Chen Zhang and Hanting Chen and Yasheng Wang and Lu Hou and Lifeng Shang},
      year={2025},
      eprint={2505.24332},
      archivePrefix={arXiv},
      primaryClass={cs.CL},
      url={https://arxiv.org/abs/2505.24332}, 
}

@inproceedings{Shetty2004TheEE,
  title={The Enron Email Dataset Database Schema and Brief Statistical Report},
  author={Jitesh Shetty and Jafar Adibi},
  year={2004},
  url={https://api.semanticscholar.org/CorpusID:59919272}
}

@misc{rossi2020temporalgraphnetworksdeep,
      title={Temporal Graph Networks for Deep Learning on Dynamic Graphs}, 
      author={Emanuele Rossi and Ben Chamberlain and Fabrizio Frasca and Davide Eynard and Federico Monti and Michael Bronstein},
      year={2020},
      eprint={2006.10637},
      archivePrefix={arXiv},
      primaryClass={cs.LG},
      url={https://arxiv.org/abs/2006.10637}, 
}

@misc{xu2019powerfulgraphneuralnetworks,
      title={How Powerful are Graph Neural Networks?}, 
      author={Keyulu Xu and Weihua Hu and Jure Leskovec and Stefanie Jegelka},
      year={2019},
      eprint={1810.00826},
      archivePrefix={arXiv},
      primaryClass={cs.LG},
      url={https://arxiv.org/abs/1810.00826}, 
}

@misc{sun2023onemultitaskpromptinggraph,
      title={All in One: Multi-task Prompting for Graph Neural Networks}, 
      author={Xiangguo Sun and Hong Cheng and Jia Li and Bo Liu and Jihong Guan},
      year={2023},
      eprint={2307.01504},
      archivePrefix={arXiv},
      primaryClass={cs.SI},
      url={https://arxiv.org/abs/2307.01504}, 
}

@misc{cai2024efficienttemporalbutterflycounting,
      title={Efficient Temporal Butterfly Counting and Enumeration on Temporal Bipartite Graphs}, 
      author={Xinwei Cai and Xiangyu Ke and Kai Wang and Lu Chen and Tianming Zhang and Qing Liu and Yunjun Gao},
      year={2024},
      eprint={2306.00893},
      archivePrefix={arXiv},
      primaryClass={cs.DS},
      url={https://arxiv.org/abs/2306.00893}, 
}

@article{kgreasoning_survey_kdd,
author = {Liu, Lihui and Wang, Zihao and Tong, Hanghang},
title = {Neural-Symbolic Reasoning over Knowledge Graphs: A Survey from a Query Perspective},
year = {2025},
issue_date = {June 2025},
publisher = {Association for Computing Machinery},
address = {New York, NY, USA},
volume = {27},
number = {1},
issn = {1931-0145},
url = {https://doi.org/10.1145/3748239.3748249},
doi = {10.1145/3748239.3748249},
journal = {SIGKDD Explor. Newsl.},
month = jul,
pages = {124–136},
numpages = {13}
}

@techreport{unifying_knowledge,
  title        = {Unifying Knowledge in Agentic LLMs: Concepts, Methods, and Recent Advancements},
  author       = {Liu, Lihui and Shu, Kai},
  institution  = {TechRxiv},
  year         = {2025},
  note         = {Available at \url{https://www.techrxiv.org/users/964779/articles/1333370-unifying-knowledge-in-agentic-llms-concepts-methods-and-recent-advancements?commit=bc73a26e68087ea82a763ea741af036a2ee2e0c5}}
}

@inproceedings{liu-2025-hyperkgr,
    title = "{H}yper{KGR}: Knowledge Graph Reasoning in Hyperbolic Space with Graph Neural Network Encoding Symbolic Path",
    author = "Liu, Lihui",
    editor = "Christodoulopoulos, Christos  and
      Chakraborty, Tanmoy  and
      Rose, Carolyn  and
      Peng, Violet",
    booktitle = "Proceedings of the 2025 Conference on Empirical Methods in Natural Language Processing",
    month = nov,
    year = "2025",
    address = "Suzhou, China",
    publisher = "Association for Computational Linguistics",
    url = "https://aclanthology.org/2025.emnlp-main.1279/",
    pages = "25188--25199",
    ISBN = "979-8-89176-332-6",
}

@inproceedings{wang-metatkgr,
author = {Wang, Ruijie and Li, Zheng and Sun, Dachun and Liu, Shengzhong and Li, Jinning and Yin, Bing and Abdelzaher, Tarek},
title = {Learning to sample and aggregate: few-shot reasoning over temporal knowledge graphs},
year = {2022},
booktitle = {Proceedings of the 36th International Conference on Neural Information Processing Systems},
series = {NeurIPS '22}
}

@inproceedings{wang-mpkd,
author = {Wang, Ruijie and Li, Zheng and Yang, Jingfeng and Cao, Tianyu and Zhang, Chao and Yin, Bing and Abdelzaher, Tarek},
title = {Mutually-paced Knowledge Distillation for Cross-lingual Temporal Knowledge Graph Reasoning},
year = {2023},
booktitle = {Proceedings of the ACM Web Conference 2023},
series = {WWW '23}
}

@inproceedings{wang-metahkg,
author = {Wang, Ruijie and Zhang, Yutong and Li, Jinyang and Liu, Shengzhong and Sun, Dachun and Wang, Tianchen and Wang, Tianshi and Chen, Yizhuo and Kara, Denizhan and Abdelzaher, Tarek},
title = {MetaHKG: Meta Hyperbolic Learning for Few-shot Temporal Reasoning},
year = {2024},
booktitle = {Proceedings of the 47th International Conference on Research and Development in Information Retrieval},
series = {SIGIR '24}
}

@misc{zhang2025atmgadadaptivetemporalmotif,
      title={ATM-GAD: Adaptive Temporal Motif Graph Anomaly Detection for Financial Transaction Networks}, 
      author={Zeyue Zhang and Lin Song and Erkang Bao and Xiaoling Lv and Xinyue Wang},
      year={2025},
      eprint={2508.20829},
      archivePrefix={arXiv},
      primaryClass={cs.LG},
      url={https://arxiv.org/abs/2508.20829}, 
}

@ARTICLE{9525268,
  author={Qiu, Zhenyu and Wu, Jia and Hu, Wenbin and Du, Bo and Yuan, Guocai and Yu, Philip S.},
  journal={IEEE Transactions on Knowledge and Data Engineering}, 
  title={Temporal Link Prediction With Motifs for Social Networks}, 
  year={2023},
  volume={35},
  number={3},
  pages={3145-3158},
  doi={10.1109/TKDE.2021.3108513}}

\newpage
\appendix
\setcounter{secnumdepth}{2} %
\renewcommand\thesubsection{\thesection.\arabic{subsection}}

\begin{center}
    {\Large\bfseries LLMTM: Benchmarking and Optimizing LLMs for Temporal Motif Analysis in Dynamic Graphs}
    \\[10pt] 
\end{center}

\textbf{Appendix}
The appendix provides further details on our methodology, experimental setup, and complete results.

\begin{itemize}[leftmargin=*]
    \item Appendix~\ref{app:all_prompts} provides the full text and structure of our prompts, detailing the six sequential components used across all tasks.
    
    \item Appendix~\ref{app:cot_prompts} presents a detailed example of the Chain-of-Thought prompt, illustrating the two-step reasoning process designed to guide the LLM.

    \item Appendix~\ref{app:agent_prompt} contains the complete, universal prompt used to control the Tool-Augmented LLM Agent's ReAct workflow.

    \item Appendix~\ref{app:analysis} presents the full 3D visualizations from our preliminary analysis, showing how graph scale ($N$), time span ($T$), and window size ($W$) influence the distribution of all nine motif types.

    \item Appendix~\ref{app:exp_settings} details the specific data generation settings for each of the six benchmark tasks.

     \item Appendix~\ref{app:algorithms} describes the algorithmic logic for our suite of five algorithmic tools, and provides the complete pseudocode for the foundational Motif\_Detection algorithm.

    \item Appendix~\ref{app:full_results} contains the complete experimental results, including performance tables for all five open-source LLMs across the four prompting strategies.

    \item Appendix~\ref{app:radas_graph} provides supplementary radar charts that compare the tool-augmented agent and the GPT-4o-mini baseline on the "Motif Classification" and "Motif Construction" tasks.
    
    \item Appendix~\ref{app:real-world} provides the experimental results on the real-world dataset.

    \item Appendix~\ref{app:deep_learning} presents the experimental results for the Deep Learning-based Methods on Motif Detection.

    \item Appendix~\ref{app:pangu-dispatcher} 
    provides the complete experimental results for openPangu-7B as the LLM for the Structure-Aware Dispatcher.
    
    \item Appendix~\ref{sec:level_0_results} provides a detailed analysis of the "Level 0" foundational dynamic graph experiments, including full results and observations on the general dynamic graph understanding of LLMs.
\end{itemize}

\section{Prompt Design}
\label{sec:appendix_prompts_algos}

\subsection{General Prompt Structure For All Tasks}
\label{app:all_prompts}
Our prompts for solving temporal motif problems on dynamic graphs are constructed from six sequential components: a dynamic graph instruction, a temporal motif instruction, a temporal motifs description, a task-specific instruction, an answer format instruction, and the final question. To prevent ambiguity, we use a distinct notation for motifs (e.g., $(u_0, u_1, t_0, a)$), even though both are represented by quadruplets. The detailed prompts for all tasks are provided in Table~\ref{tab:prompt_example_1} through~\ref{tab:prompt_example_6}. For brevity, since the initial instruction set (dynamic graph, temporal motif, and motif description) is identical for all tasks within a given level, we only present it once per level: in the “Motif Classification” prompt for Level 1, and in the “Multi-Motif Detection” prompt for Level 2.

\begin{table}[htbp]
    \centering
    \small 
    \setlength{\tabcolsep}{4pt} 
    \begin{tabularx}{\columnwidth}{lX}
        \toprule
        \textbf{Motif Classification} & \textbf{Prompt} \\
        \midrule
        
        \begin{tabular}[t]{@{}l@{}}DyG \\ Instruction\end{tabular} & 
        In an undirected dynamic graph, ($u, v, t, a$) means that node $u$ and node $v$ are linked with an undirected edge at time $t$, ($u, v, t, d$) means that node $u$ and node $v$ are deleted with an undirected edge at time $t$. \\
        \midrule
        
        \begin{tabular}[t]{@{}l@{}}Motif \\ Instruction\end{tabular} & 
        A $k$-node, $l$-edge, $\delta$-temporal motif is a time-ordered sequence of $k$ nodes and $l$ distinct edges within $\delta$ duration, i.e., $M = (u1, v1, t1, a), (u2, v2, t2, a) \dots , $\newline$(ul, vl, tl, a)$. These edges $(u1, v1), (u2, v2) \dots , (ul, vl)$ form a specific pattern, $t1 < t2 < \dots < tl$ are strictly increasing, and $tl - t1 \leq \delta$. That is each edge that forms a specific pattern occurs in specific order within a fixed time window. Each consecutive event shares at least one common node. \\
        \midrule
        
        \begin{tabular}[t]{@{}l@{}}Task \\ Instruction\end{tabular} & 
        Your task is to answer whether the given undirected dynamic graph is the given temporal motif?\\
        \midrule
        
        \begin{tabular}[t]{@{}l@{}}Answer \\ Instruction\end{tabular} & 
        Give the answer as “Yes” or “No” at the end of your response after 'Answer:'. \\
        \midrule
        Question & 
        Here is what you need to answer: Question: Given an undirected dynamic graph with the edges [(1, 2, 0, a), (0, 2, 1, a), (0, 1, 4, a)]. Given a triangle temporal motif which is a 3-node, 3-edge, 5-temporal motif with the edges[($u0, u1, t0, a$), ($u1, u2, t1, a$), ($u2, u0, t2, a$)]. Whether the given undirected dynamic graph is the given motif? \\
        \bottomrule
    \end{tabularx}
    \caption{Prompt construction for the "Motif Classification" task.}
    \label{tab:prompt_example_1}
\end{table}

\begin{table}[htbp]
    \centering
    \small 
    \setlength{\tabcolsep}{4pt} 
    \begin{tabularx}{\columnwidth}{lX}
        \toprule
        \textbf{Motif Detection} & \textbf{Prompt} \\
        \midrule

        \begin{tabular}[t]{@{}l@{}}Task \\ Instruction\end{tabular} & 
        Your task is to determine whether the given undirected dynamic graph contains the given temporal motif? This means that there exists a subgraph within the undirected dynamic graph that matches both the structure and the temporal constraints of the temporal motif.\\
        \midrule
        
        \begin{tabular}[t]{@{}l@{}}Answer \\ Instruction\end{tabular} & 
        Give the answer as “Yes” or “No” at the end of your response after 'Answer:'. \\
        \midrule
        
        Question & 
        Here is what you need to answer: Question: Given an undirected dynamic graph with the edges [(3, 8, 0, a), (0, 5, 1, a), (3, 9, 1, a), (4, 6, 1, a), (8, 9, 1, a), (0, 4, 2, a), (2, 5, 2, a), (3, 4, 2, a), (1, 3, 3, a), (4, 9, 3, a), (0, 5, 4, d), (1, 3, 4, d), (1, 9, 4, a), (4, 9, 4, d)].Given a triangle temporal motif which is a 3-node, 3-edge, 4-temporal motif with the edges[($u0, u1, t0, a$), ($u1, u2, t1, a$), ($u2, u0, t2, a$)]. Whether the given undirected dynamic graph contains the given temporal motif? \\
        \bottomrule
    \end{tabularx}
    \caption{Prompt construction for the "Motif Detection" task.}
    \label{tab:prompt_example_2}
\end{table}

\begin{table}[htbp]
    \centering
    \small 
    \setlength{\tabcolsep}{4pt} 
    \begin{tabularx}{\columnwidth}{lX}
        \toprule
        \textbf{Motif Construction} & \textbf{Prompt} \\
        \midrule

        \begin{tabular}[t]{@{}l@{}}Task \\ Instruction\end{tabular} & 
        Your task is to answer How to modify the given undirected dynamic graph so that it contains the given temporal motif? The number of modifications should be as small as possible.\\
        \midrule
        
        \begin{tabular}[t]{@{}l@{}}Answer \\ Instruction\end{tabular} & 
        Give the answer as a list of 4-tuples at the end of your response after 'Answer:' \\
        \midrule
        
        Question & 
        Here is what you need to answer: Question: Given an undirected dynamic graph with the edges [(7, 8, 1, a), (1, 8, 2, a), (1, 5, 4, a), (2, 9, 4, a), (7, 8, 6, d), (1, 8, 7, d), (7, 8, 7, a), (0, 4, 8, a), (4, 9, 8, a), (2, 9, 9, d)]. Given a 4-cycle temporal motif which is a 4-node, 4-edge, 5-temporal motif with the edges[($u0, u1, t0, a$), ($u1, u2, t1, a$), ($u2, u3, t2, a$), ($u3, u0, t3, a$)]. How to modify the given undirected dynamic graph so that it contains the given temporal motif? \\
        \bottomrule
    \end{tabularx}
    \caption{Prompt construction for the "Motif Construction" task.}
    \label{tab:prompt_example_3}
\end{table}

\begin{table}[htbp]
    \centering
    \small 
    \setlength{\tabcolsep}{4pt} 
    \begin{tabularx}{\columnwidth}{lX}
        \toprule
        \textbf{Multi-Motif Detection} & \textbf{Prompt} \\
        \midrule
        
        \begin{tabular}[t]{@{}l@{}}DyG \\ Instruction\end{tabular} & 
        In an undirected dynamic graph, ($u, v, t, a$) means that node $u$ and node $v$ are linked with an undirected edge at time $t$, ($u, v, t, d$) means that node $u$ and node $v$ are deleted with an undirected edge at time $t$. \\
        \midrule
        
        \begin{tabular}[t]{@{}l@{}}Motif \\ Instruction\end{tabular} & 
        A $k$-node, $l$-edge, $\delta$-temporal motif is a time-ordered sequence of $k$ nodes and $l$ distinct edges within $\delta$ duration, i.e., $M = (u1, v1, t1, a), (u2, v2, t2, a) \dots ,$\newline$(ul, vl, tl, a)$. These edges $(u1, v1), (u2, v2) \dots , (ul, vl)$ form a specific pattern, $t1 < t2 < \dots < tl$ are strictly increasing and $tl - t1 \leq \delta$. That is each edge that forms a specific pattern occurs in specific order within a fixed time window. Each consecutive event shares at least one common node. \\
        \midrule
        \begin{tabular}[t]{@{}l@{}}Temporal \\ Motifs\end{tabular} & 
        Possible temporal motifs in the dynamic graph include:
 
    triangle: a 3-node, 3-edge, 3-temporal motif with the edges [($u0, u1, t0, a$), ($u1, u2, t1, a$), ($u2, u0, t2, a$)]
    
    3-star: a 4-node, 3-edge, 3-temporal motif with the edges [($u0, u1, t0, a$), ($u0, u2, t1, a$), ($u0, u3, t2, a$)]
    
    4-path: a 4-node, 3-edge, 3-temporal motif with the edges [($u0, u1, t0, a$), ($u1, u2, t1, a$), ($u2, u3, t2, a$)]\newline
    ...\newline
    
    bitriangle: 6-node, 6-edge, 15-temporal motif with the edges [($u0, u1, t0, a$), ($u1, u3, t1, a$), ($u3, u5, t2, a$), ($u5, u4, t3, a$), ($u4, u2, t4, a$), ($u2, u0, t5, a$)]
     \\
        \midrule
        \begin{tabular}[t]{@{}l@{}}Task \\ Instruction\end{tabular} & 
        Your task is to identify What temporal motifs present in the given undirected dynamic graph?\\
        \midrule
        
        \begin{tabular}[t]{@{}l@{}}Answer \\ Instruction\end{tabular} & 
        Give the answer by listing the names of temporal motifs at the end of your response after 'Answer:' \\
        \midrule
        Question & 
        Here is what you need to answer: Question: Given an undirected dynamic graph with the edges [(1, 4, 0, a), (4, 10, 0, a), (7, 9, 1, a), (9, 16, 1, a), (12, 13, 1, a), (16, 17, 1, a), (8, 11, 2, a), (8, 12, 2, a), (9, 16, 2, d) ...]. What temporal motifs present in the given undirected dynamic graph?\\
        \bottomrule
    \end{tabularx}
    \caption{Prompt construction for the "Multi-Motif Detection" task.}
    \label{tab:prompt_example_4}
\end{table}

\begin{table}[htbp]
    \centering
    \small 
    \setlength{\tabcolsep}{4pt} 
    \begin{tabularx}{\columnwidth}{lX}
        \toprule
        \textbf{Motif Occurrence Prediction} & \textbf{Prompt} \\
        \midrule
        \begin{tabular}[t]{@{}l@{}}Task \\ Instruction\end{tabular} & 
        Your task is to answer When does each of the above temporal motifs first appear in the given undirected dynamic graph? The appearance time of a temporal motif refers to the time when the last edge of the temporal motif occurs. Omit temporal motifs that do not appear in the undirected dynamic graph.\\
        \midrule
        
        \begin{tabular}[t]{@{}l@{}}Answer \\ Instruction\end{tabular} & 
        Give the answer as a list of tuples consisting of the name of a temporal motif and an integer at the end of your response after 'Answer:' \\
        \midrule
        Question & 
        Here is what you need to answer: Question: Given an undirected dynamic graph with the edges[(1, 3, 0, a), (1, 7, 0, a), (8, 12, 0, a), (10, 16, 0, a), (1, 9, 1, a), (7, 19, 1, a), (10, 13, 1, a), (10, 18, 1, a), (2, 5, 2, a) ...]. When does each of the above temporal motifs first appear in the given undirected dynamic graph?\\
        \bottomrule
    \end{tabularx}
    \caption{Prompt construction for the "Motif Occurrence Prediction" task.}
    \label{tab:prompt_example_5}
\end{table}

\begin{table}[htbp]
    \centering
    \small 
    \setlength{\tabcolsep}{4pt} 
    \begin{tabularx}{\columnwidth}{lX}
        \toprule
        \textbf{Multi-Motif Count} & \textbf{Prompt} \\
        \midrule
        \begin{tabular}[t]{@{}l@{}}Task \\ Instruction\end{tabular} & 
        Your task is to answer How many times does each of the above temporal motifs appear in the given undirected dynamic graph? Omit temporal motifs that do not appear in the undirected dynamic graph. \\
        \midrule
        
        \begin{tabular}[t]{@{}l@{}}Answer \\ Instruction\end{tabular} & 
        Give the answer as a list of tuples consisting of the name of a temporal motif and an integer at the end of your response after 'Answer:' \\
        \midrule
        Question & 
        Here is what you need to answer: Question: Given an undirected dynamic graph with the edges [(0, 16, 0, a), (9, 10, 0, a), (9, 17, 0, a), (14, 17, 0, a), (6, 10, 1, a), (7, 13, 1, a), (11, 14, 1, a), (16, 19, 1, a), (2, 18, 2, a), ...]. How many times does each of the above temporal motifs appear in the given undirected dynamic graph?\\
        \bottomrule
    \end{tabularx}
    \caption{Prompt construction for the "Multi-Motif Count" task.}
    \label{tab:prompt_example_6}
\end{table}

\subsection{Chain-of-Thought Prompt}
\label{app:cot_prompts}

We designed CoT prompts to guide the LLM through a two-step reasoning process for solving temporal motif problems: first, describing a motif's structural characteristics in words to activate semantic understanding, and second, sequentially checking each of the four key constraints (\textbf{Structural}, \textbf{Connectivity}, \textbf{Temporal}, and \textbf{Duration}). An example for the "Motif Detection" task is provided in Figure~\ref{fig:cot_prompt}.




\begin{figure*}[t] 
    \begin{tcolorbox}[
      colback=gray!10,
      colframe=black,
      arc=3mm,
      boxrule=0.5pt,
      title=Chain of Thought Prompt
      
    ]
    In an undirected dynamic graph, ($u, v, t, a$) means that node $u$ and node $v$ are linked with an undirected edge at time $t$, ($u, v, t, d$) means that node $u$ and node $v$ are deleted with an undirected edge at time $t$.

    A $k$-node, $l$-edge, $\delta$-temporal motif is a time-ordered sequence of $k$ nodes and $l$ distinct edges within $\delta$ duration, i.e., $M = (u1, v1, t1, a), (u2, v2, t2, a) \dots ,(ul, vl, tl, a)$. These edges $(u1, v1), (u2, v2) \dots , (ul, vl)$ form a specific pattern, $t1 < t2 < \dots < tl$ are strictly increasing, and $tl - t1 \leq \delta$. That is each edge that forms a specific pattern occurs in specific order within a fixed time window. Each consecutive event shares at least one common node. When searching for a specific temporal motif in the undirected dynamic graph, it is necessary to match pattern, edge order and time window. Node IDs and exact timestamps are irrelevant. Meanwhile,you should only focus on added edges. Note that some patterns are symmetric, so the order of the corresponding timestamps may be unimportant.

    Your task is to determine whether the given undirected dynamic graph contains the given temporal motif? This means that there exists a subgraph within the undirected dynamic graph that matches both the structure and the temporal constraints of the temporal motif.

    Give the answer as “Yes” or “No” at the end of your response after 'Answer:'.
    
    Here is an example:
    
    \textbf{Question:} Given an undirected dynamic graph with the edges [(1, 4, 0, a), (2, 3, 1, a), (4, 2, 2, a), (2, 1, 3, a), (0, 3, 4, a), (0, 3, 5, d)]. Given a triangle temporal motif which is a 3-node, 3-edge, 4-temporal motif with the edges [(u0, u1, t0, a), (u1, u2, t1, a), (u2, u0, t2, a)]. Whether the given undirected dynamic graph contains the given temporal motif?

    \vspace{1em}

    \textbf{**Chain of Thought:**}
    
    \begin{enumerate}[label=\arabic*.]
        \item My goal is to determine if the given dynamic graph contains a triangle motif.
        
        \item I'll start by identifying the nodes and all "a" operation edges in the graph.
        \begin{itemize}[label=*, leftmargin=2.5em]
            \item Nodes: 0, 1, 2, 3, 4
            \item Edges: (1, 4, 0, 'a'), (2, 3, 1, 'a'), (4, 2, 2, 'a'), (2, 1, 3, 'a'), (0, 3, 4, 'a')
        \end{itemize}
        
        \item Next, I'll check if the graph contains a triangle motif.
        \begin{itemize}[label=*, leftmargin=2.5em]
            \item A triangle motif is a sequence of \textbf{three edges forming a closed cycle}, where timestamps are strictly increasing, and the difference between the maximum and minimum timestamp is at most 4.
            \item In this undirected graph, the edges (1, 4, 0, 'a'), (4, 2, 2, 'a'), (2, 1, 3, 'a') form a \textbf{closed cycle}.
            \item The timestamps 0, 2, 3 are \textbf{strictly increasing}.
            \item The difference between the maximum and minimum timestamp is $3 - 0 = 3 < 4$, which satisfies the \textbf{time window} constraint.
        \end{itemize}
        
        \item Therefore, the graph contains a triangle motif.
    \end{enumerate}
    
    \textbf{Answer:} Yes
    
    Here is what you need to answer:
    
    Question: Given an undirected dynamic graph with the edges [(1, 5, 0, a), (5, 6, 1, a), (7, 9, 2, a), (1, 8, 3, a), (3, 5, 3, a), (4, 7, 3, a), (6, 8, 3, a), (0, 4, 4, a), (3, 5, 4, d)].Given a triangle temporal motif which is a 3-node, 3-edge, 4-temporal motif with the edges[(u0, u1, t0, a), (u1, u2, t1, a), (u2, u0, t2, a)]. Whether the given undirected dynamic graph contains the given temporal motif?
    
    Answer:
    \end{tcolorbox}
    \caption{An example of a CoT prompt for the “Motif Detection” task.}
    \label{fig:cot_prompt} 
\end{figure*}

\subsection{Tool-Augmented Agent Prompt}
\label{app:agent_prompt}
We designed a single, universal prompt that enables our Tool-Augmented LLM Agent to effectively solve all tasks in the LLMTM benchmark. As illustrated in Figure~\ref{fig:agent_prompt}, this unified prompt compels the model to follow two key principles: (1) a structured reasoning framework for interpretability and (2) precise formatting for robust tool calls.

\section{Experimental and Implementation Details}
\label{sec:appendix_exp_analysis}

\subsection{Parameter Analysis for Data Generation}
\label{app:analysis}
\begin{figure*}[t]
    \centering
    \includegraphics[width=0.9\textwidth]{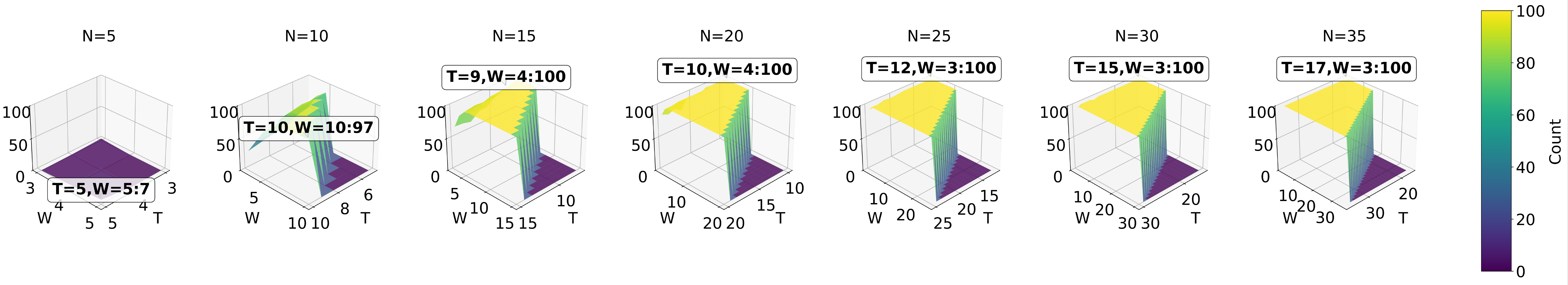}
    \caption{A 3D analysis of how dynamic graph scale ($N$), time span ($T$), and window size ($W$) influence the distribution of the "3-star" motif. The peak count and its corresponding settings are annotated.}
    \label{fig:3d_3-star}
\end{figure*}

\begin{figure*}[t]
    \centering
    \includegraphics[width=0.9\textwidth]{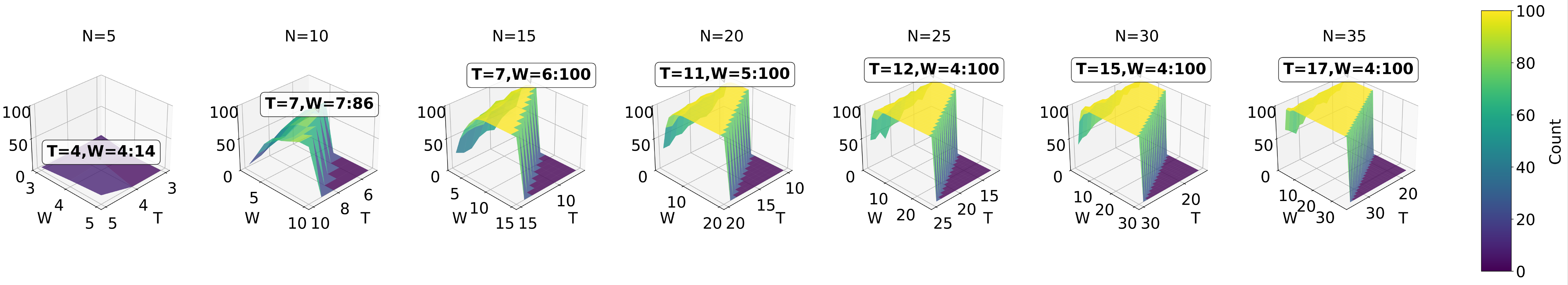}
    \caption{A 3D analysis of how dynamic graph scale ($N$), time span ($T$), and window size ($W$) influence the distribution of the "triangle" motif. The peak count and its corresponding settings are annotated.}
    \label{fig:3d_triangle}
\end{figure*}

\begin{figure*}[t]
    \centering
    \includegraphics[width=0.9\textwidth]{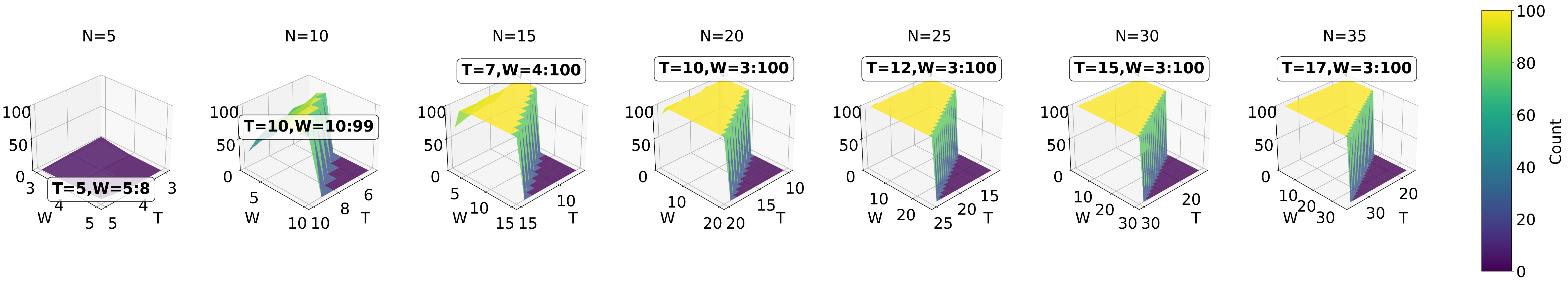}
    \caption{A 3D analysis of how dynamic graph scale ($N$), time span ($T$), and window size ($W$) influence the distribution of the "4-path" motif. The peak count and its corresponding settings are annotated.}
    \label{fig:3d_4-path}
\end{figure*}

\begin{figure*}[t]
    \centering
    \includegraphics[width=0.9\textwidth]{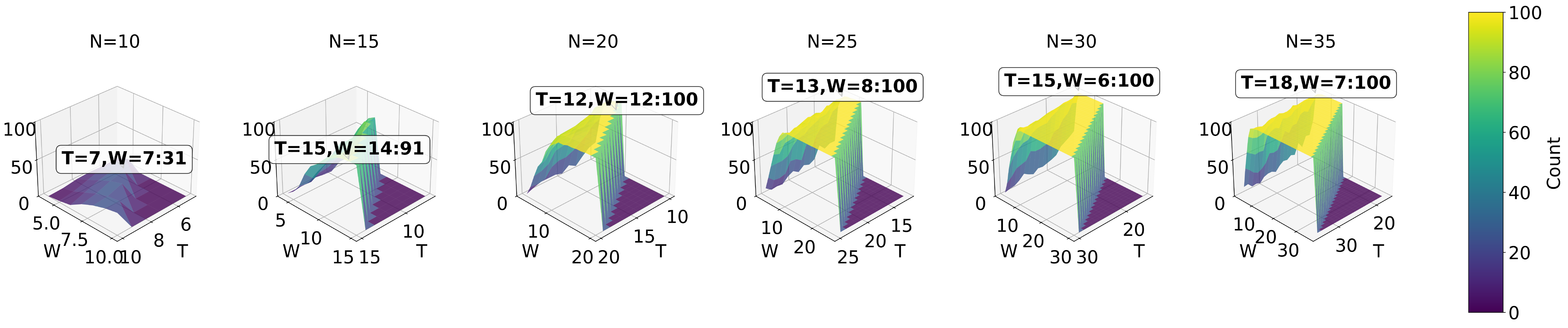}
    \caption{A 3D analysis of how dynamic graph scale ($N$), time span ($T$), and window size ($W$) influence the distribution of the "4-tailedtriangle" motif. The peak count and its corresponding settings are annotated.}
    \label{fig:3d_4-tailedtriangle}
\end{figure*}

\begin{figure*}[t]
    \centering
    \includegraphics[width=0.9\textwidth]{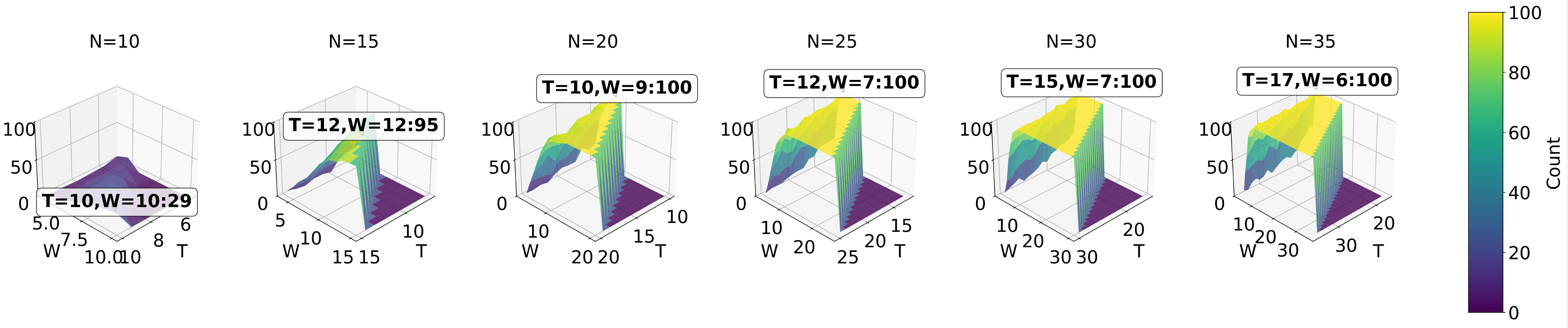}
    \caption{A 3D analysis of how dynamic graph scale ($N$), time span ($T$), and window size ($W$) influence the distribution of the "4-cycle" motif. The peak count and its corresponding settings are annotated.}
    \label{fig:3d_4-cycle}
\end{figure*}

\begin{figure*}[t]
    \centering
    \includegraphics[width=0.9\textwidth]{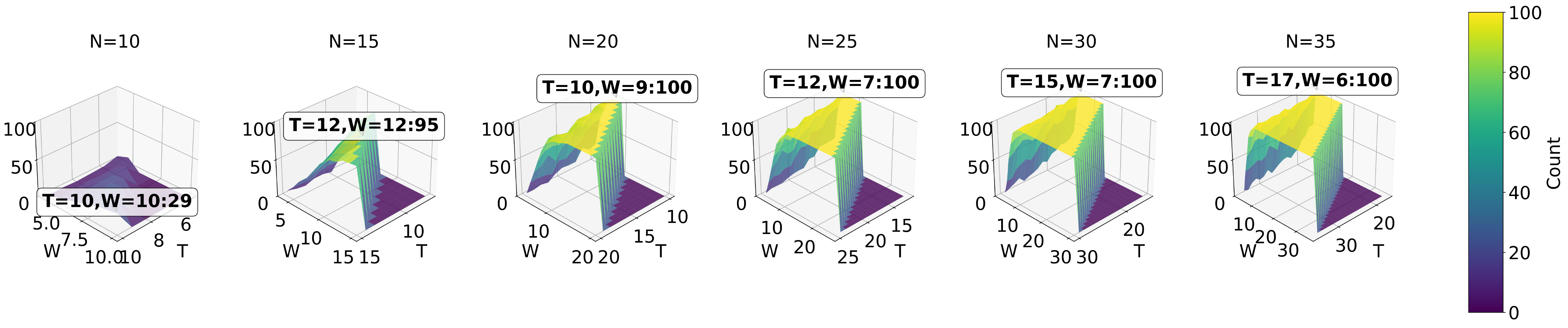}
   \caption{A 3D analysis of how dynamic graph scale ($N$), time span ($T$), and window size ($W$) influence the distribution of the "butterfly" motif. The peak count and its corresponding settings are annotated.}
    \label{fig:3d_butterfly}
\end{figure*}

\begin{figure*}[t]
    \centering
    \includegraphics[width=0.9\textwidth]{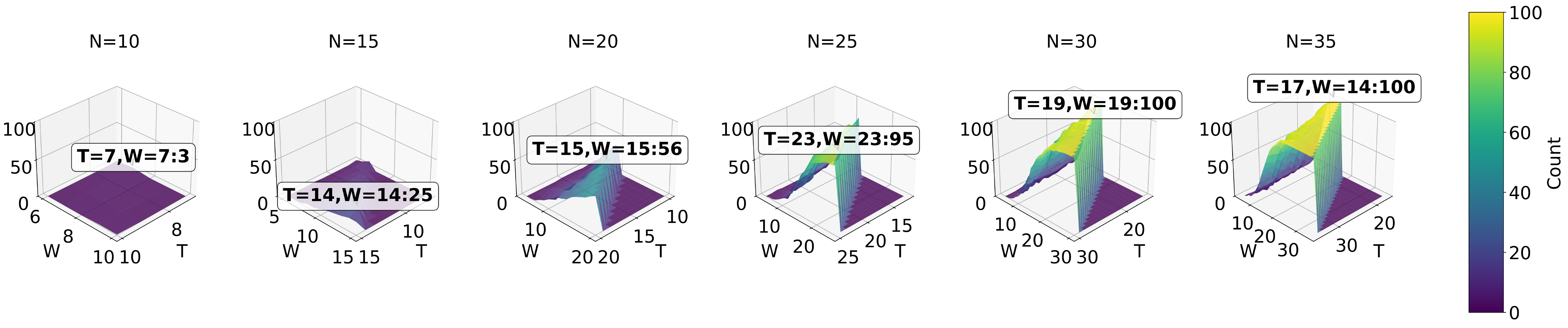}
    \caption{A 3D analysis of how dynamic graph scale ($N$), time span ($T$), and window size ($W$) influence the distribution of the "4-chordalcycle” motif. The peak count and its corresponding settings are annotated.}
    \label{fig:3d_4-chordalcycle}
\end{figure*}

\begin{figure*}[t]
    \centering
    \includegraphics[width=0.9\textwidth]{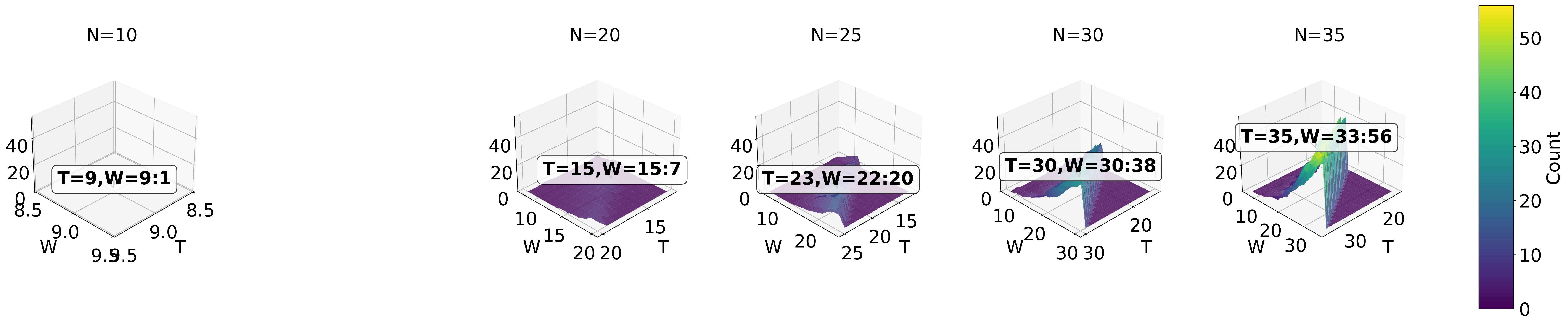}
    \caption{A 3D analysis of how dynamic graph scale ($N$), time span ($T$), and window size ($W$) influence the distribution of the "4-clique” motif. The peak count and its corresponding settings are annotated.}
    \label{fig:3d_4-clique}
\end{figure*}

\begin{figure*}[t]
    \centering
    \includegraphics[width=0.9\textwidth]{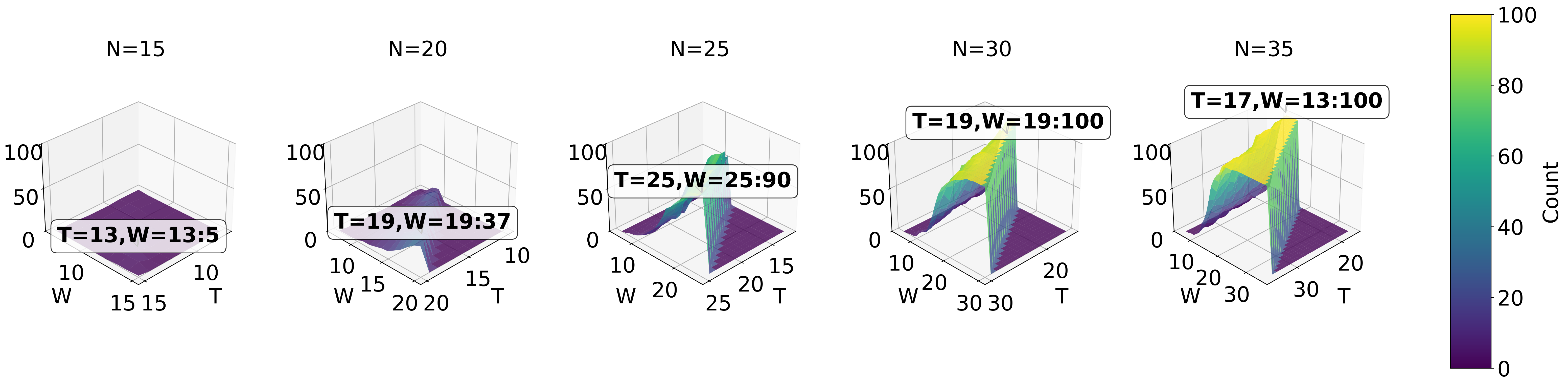}
   \caption{A 3D analysis of how dynamic graph scale ($N$), time span ($T$), and window size ($W$) influence the distribution of the "bitriangle” motif. The peak count and its corresponding settings are annotated.}
    \label{fig:3d_bitriangle}
\end{figure*}

To generate datasets with a balanced distribution of positive and negative temporal motif instances for a fair evaluation, we first conducted a preliminary analysis. We investigated how graph scale ($N$), time span ($T$), and window size ($W$) collectively influence the distribution of all nine temporal motifs. The 3D visualizations of this analysis are presented in Figures~\ref{fig:3d_3-star} through~\ref{fig:3d_bitriangle}.

\begin{figure*}[t]
    \begin{tcolorbox}[
        colback=gray!10,
        colframe=black,
        arc=3mm,
        boxrule=0.5pt,
        title=Agent Prompt Template
    ]
        
    You are a tool-using assistant for dynamic graph analysis.
    \vspace{1em}
    
    You have access to the following tools: \{tools\}
    
    \vspace{1em}
    Use the following format:
    \begin{itemize}[label={}, leftmargin=1.5em, itemsep=0pt, topsep=2pt, parsep=0pt]
        \item \textbf{Question:} the input question you must answer
        \item \textbf{Thought:} you should always think about what to do
        \item \textbf{Action:} the action to take, should be one of [\{tool\_names\}]. The action name must be shown.
        \item \textbf{Action Input:} the input to the action. The complete input parameters must be shown (Even if it has shown before, copy it here exactly). 
        \item \textbf{Observation:} the result of the action
        \item \textbf{Thought:} I now know the final answer
        \item \textbf{Final Answer:} the final answer to the original input question
    \end{itemize}

    CRITICAL RULES FOR TOOL USAGE:
    \begin{itemize}[label={}, leftmargin=1.5em, itemsep=0pt, topsep=2pt, parsep=0pt]
        \item - Answer ONLY what the user asked - NOTHING MORE.
        \item - Use EXACTLY ONE tool - NEVER call multiple tools.
        \item - Action Input MUST be a dictionary that maps the parameters of the tool to the data in the question.
        \item - The "edge\_list" parameter MUST be a list of 4-element arrays: ($u, v, t, operation$), $u, v$ and $t$ are integers, operation is a string like "a" or "d".
        \item - You MUST copy the edge list from the user's question EXACTLY as it appears. Do not add, remove, or modify any edges. 
        \item - For the "motif\_list" parameter MUST be a dictionary that maps motif name to a nested object containing its edge\_pattern and time\_window.
    \end{itemize}
        
    

\begin{itemize}[label={}, leftmargin=3em, itemsep=3pt, topsep=2pt, parsep=0pt]
    \item 
    - Motif\_list Parameter Format: \{\{ \\ 
    \hspace*{1em}"motif\_name": \{\{ \\ 
    \hspace*{2em}"edge\_pattern": [($u0, v0, t0, operation$), ($u1, v1, t1, operation$), ...], \\
    \hspace*{2em}"time\_window": time\_window \\
    \hspace*{1em}\}\} \\
    \}\}
    
    \item - time\_window is the number extracted from "X-temporal motif" in the question.
    \item - edge\_pattern is a list of tuples, each tuple is a 4-element array: ($u0, v0, t0, operation$), $u0, v0, t0$ are strings with u, u, t as the initials, extracted from $"u0", "u1", "t0"$ in the question, operation is a string like "a" or "d".
\end{itemize}     

\begin{itemize}[label={}, leftmargin=1.5em, itemsep=3pt, topsep=2pt, parsep=0pt]
\item - For the "motif\_definitions" parameter MUST be is a dictionary that maps each motif name to a nested object containing its edge\_pattern and time\_window.
\end{itemize}  

\begin{itemize}[label={}, leftmargin=3em, itemsep=3pt, topsep=2pt, parsep=0pt]        
            \item
            - Motif\_definitions Parameter Format: \{\{ \par
            \hspace*{1.5em}"motif\_name1": \{\{ \par
            \hspace*{3em}"edge\_pattern": [($u0, v0, t0, operation$), ($u1, v1, t1, operation$), ...], \par
            \hspace*{3em}"time\_window": time\_window \par
            \hspace*{1.5em}\}\}, \par 
            \hspace*{1.5em}"motif\_name2": \{\{ \par
            \hspace*{3em}"edge\_pattern": [($u0, v0, t0, operation$), ($u1, v1, t1, operation$), ...], \par
            \hspace*{3em}"time\_window": time\_window \par
            \hspace*{1.5em}\}\}, \par 
            \hspace*{1.5em}... \par
            \}\} \par

        \item - time\_window is the number extracted from "X-temporal motif" in the question.
        \item - edge\_pattern is a list of tuples, each tuple is a 4-element array: ($u0, v0, t0, operation$), directly copied from the question.
\end{itemize}

\begin{itemize}[label={}, leftmargin=1.5em, itemsep=3pt, topsep=2pt, parsep=0pt] 
        \item - After you receive an Observation, IMMEDIATELY output Final Answer.
        \item - The Final Answer MUST be the EXACT content from the last Observation, without any modification or conversational filler.
    \end{itemize}
    
    \vspace{1em}
    \textbf{Begin!}
    
    \par\noindent
    Question: \{input\}
    \par\noindent
    Thought:\texttt\{agent\_scratchpad\}
    \end{tcolorbox}
    
    \caption{The prompt template for the Tool-Augmented LLM Agent.}
    \label{fig:agent_prompt}
\end{figure*}

\subsection{Task-Specific Data Settings}
\label{app:exp_settings}

Based on the analysis above, we tailor the data generation settings for specific tasks, as detailed in the table~\ref{tab:motif_params} through ~\ref{tab:motif_params_example}. For "Motif Classification", the graph scale is matched to the size of the query motif. For "Motif Detection", we select settings that ensure an approximate 50\% presence rate for each motif type. For the Level 2 multi-motif tasks, settings are chosen to ensure a non-zero occurrence probability for all motif types.

\begin{table}[htbp]
    \centering
    
    \begin{tabularx}{\columnwidth}{
        X
        S[table-format=1.0]
        S[table-format=1.1]
        S[table-format=2.0]
        S[table-format=2.0]
    }
        \toprule
        \multicolumn{1}{c}{\textbf{Motif Classification}} & 
        {\textbf{N}} & 
        {\textbf{M}} & 
        {\textbf{T}} & 
        {\textbf{W}} \\
        \midrule
        
        3-star           & 4 & 3 & 5  & 5  \\
        triangle         & 3 & 3 & 5  & 5  \\
        4-path           & 4 & 3 & 5  & 5  \\
        4-cycle          & 4 & 4 & 5  & 5  \\
        4-chordalcycle   & 4 & 5 & 10 & 10 \\
        4-tailedtriangle & 4 & 4 & 5  & 5  \\
        4-clique         & 4 & 6 & 10 & 10 \\
        bitriangle       & 6 & 6 & 10 & 10 \\
        butterfly        & 4 & 4 & 5  & 5  \\
        
        \bottomrule
    \end{tabularx}
    \caption{Settings for different motif types in the "Motif Classification" task, where $M$ is the number of edges.}
    \label{tab:motif_params}

\end{table}

\begin{table}[htbp]
    \centering
    
    \begin{tabularx}{\columnwidth}{
        >{\raggedright\arraybackslash}X 
        *{3}{S[table-format=2.0]}
    }
        \toprule
        \multicolumn{1}{c}{\textbf{Motif Detection}} & 
        {\textbf{N}} & 
        {\textbf{T}} & 
        {\textbf{W}} \\
        \midrule
        
        3-star           & 10 & 5  & 3  \\
        triangle         & 10 & 5  & 4  \\
        4-path           & 10 & 5  & 3  \\
        4-cycle          & 15 & 10 & 6  \\
        4-chordalcycle   & 20 & 15 & 14 \\
        4-tailedtriangle & 15 & 10 & 7  \\
        4-clique         & 35 & 30 & 27 \\
        bitriangle       & 25 & 20 & 14 \\
        butterfly        & 15 & 10 & 6  \\
        
        \bottomrule
    \end{tabularx}
    \caption{Settings for the "Motif Detection" task. The parameters for each motif type are chosen to ensure an approximate 50\% presence rate, while the edge probability is fixed at $p=0.3$.}
    \label{tab:motif_params_contain}
\end{table}

\begin{table}[htbp]
    \centering
    
    \begin{tabularx}{\columnwidth}{
        >{\raggedright\arraybackslash}X 
        *{3}{S[table-format=2.0]}
    }
        \toprule
        \multicolumn{1}{c}{\textbf{Motif Construction}} & 
        {\textbf{N}} & 
        {\textbf{T}} & 
        {\textbf{W}} \\
        \midrule
        
        4-cycle          & 10 & 10 & 5  \\
        4-tailedtriangle & 10 & 10 & 5  \\
        4-chordalcycle   & 10 & 15 & 10 \\
        4-clique         & 10 & 15 & 10 \\
        bitriangle       & 10 & 15 & 10 \\
        
        \bottomrule
    \end{tabularx}
    \caption{Settings for different motif types in the "Motif Construction" task, where edge probability is fixed at $p=0.3$.}
    \label{tab:motif_params_modify}
\end{table}

\begin{table}[t]
    \centering
    
    \begin{tabularx}{\columnwidth}{
        >{\raggedright\arraybackslash}X 
        *{3}{S[table-format=2.0]}
    }
        \toprule
        \multicolumn{1}{c}{\textbf{Level 2 Tasks}} & 
        {\textbf{N}} & 
        {\textbf{T}} & 
        {\textbf{W}} \\
        \midrule
        
        3-star           & 20 & 15 & 3  \\
        triangle         & 20 & 15 & 3  \\
        4-path           & 20 & 15 & 3  \\
        4-cycle          & 20 & 15 & 6  \\
        4-chordalcycle   & 20 & 15 & 14 \\
        4-tailedtriangle & 20 & 15 & 6  \\
        4-clique         & 20 & 15 & 15 \\
        bitriangle       & 20 & 15 & 15 \\
        butterfly        & 20 & 15 & 6  \\
        
        \bottomrule
    \end{tabularx}
    \caption{Settings for all Level 2 multi-motif tasks. The parameters are chosen to ensure a non-zero occurrence probability for all motif types, where the edge probability is fixed at $p=0.3$.}
    \label{tab:motif_params_example}
\end{table}

\subsection{Algorithmic Tools}
\label{app:algorithms}

To construct the tool-augmented LLM agent, we developed a suite of five algorithmic tools with key components for motif reasoning, as detailed in Table~\ref{tab:algorithmic_logic}. The most fundamental of these is the Motif\_Detection algorithm, whose pseudocode is presented in Algorithm~\ref{alg:judge_motif_code}.

\begin{algorithm}[tbh!]
\caption{Motif\_Detection Algorithm}
\label{alg:judge_motif_code}

\textbf{Input}: 

\begin{tabular}[t]{l p{0.9\columnwidth}}
    $G_{dy}:$ & A dynamic graph of 4-tuples ($u, v, t, op$). \\
    $M_{tem}:$ & A temporal motif of 4-tuples ($u0, u1, t0, a$). \\
    $\Delta T:$ & The time window of temporal motif. \\
\end{tabular}
\vspace{1mm}
\textbf{Output}: True if a valid instance of $M_{tem}$ exists in $G_{dy}$, False otherwise.
\vspace{2mm}

\begin{algorithmic}[1] 
\STATE // Step 1: Parse inputs into static graphs and temporal data structures
\STATE $G_S, T_{map} \leftarrow \text{ParseDynamicGraph}(G_{dy})$
\STATE $M_S, E_M^{sorted} \leftarrow \text{ParseTemporalMotif}(M_{tem})$
\STATE // Step 2: Find all potential topological matches
\STATE $\Phi \leftarrow \text{FindSubgraphIsomorphisms}(M_S, G_S)$
\STATE // Step 3: For each match, verify temporal constraints
\FOR{each mapping $\phi \in \Phi$}
    \IF{a valid timestamp assignment $T_{instance}$ exists and \newline within $\Delta T:$ for $\phi$}
        \STATE \textbf{return} True \quad // A valid instance is found
    \ENDIF
\ENDFOR
\STATE \textbf{return} False \quad // No valid instance found
\end{algorithmic}
\end{algorithm}

\section{Experimental Results}
\label{sec:appendix_results}

\subsection{Detailed Experimental Results}
\label{app:full_results}
The following tables (Tables~\ref{tab:prompting_qwen7b} through~\ref{tab:level2_qwq32b}) present the detailed experimental results for all six tasks. We evaluate five open-source LLMs (including the DeepSeek-R1-Distill-Qwen series, Qwen2.5-32B-Instruct, and QwQ-32B) across four distinct prompting techniques (zero/one-shot and zero/one-shot CoT). These results are provided to facilitate further research in this area.

\setlength{\dbltextfloatsep}{10pt} 
\begin{table*}[htbp]
    \centering
    
    \begin{tabularx}{\textwidth}{>{\raggedright\arraybackslash}l >{\raggedright\arraybackslash}X}
        \toprule
        \textbf{Tools} & \textbf{Algorithmic Logic} \\
        \midrule
        
        Motif\_Detection & 
        \begin{enumerate}[label=(\arabic*), nosep, leftmargin=*]
            \item Use the classic GraphMatcher algorithm to find all subgraph isomorphisms satisfying the Structural and Connectivity constraints.
            \item Each resulting topological mapping is verified against the Temporal and Duration constraints.
        \end{enumerate} \\
        \midrule
        Motif\_Construction & 
        \begin{enumerate}[label=(\arabic*), nosep, leftmargin=*]
            \item Construct the temporal motif pattern that is missing only its final edge.
            \item Use the Motif\_Detection function to find all instances of this incomplete motif within the dynamic graph.
            \item Add the missing final edge, ensuring the correct timestamp.
            \item Verify that a complete temporal motif now exists after the addition.
        \end{enumerate} \\
        \midrule
        Multi\_Motif\_Detection & Iteratively use the Motif\_Detection function to determine if an instance of each temporal motif type exists. \\
        \midrule
        Motif\_Occurrence\_Prediction & Iteratively use the Motif\_Detection function to identify if each temporal motif type is present, and if so, records the timestamp of its first(minimum) occurrence. \\
        \midrule
        Multi\_Motif\_Count & Iteratively use the Motif\_Detection function to find all instances of each temporal motif type and records the total count. \\
        \bottomrule
    \end{tabularx}
    \caption{The algorithmic logic for each tool.}
    \label{tab:algorithmic_logic}
\end{table*}

\begin{table*}[t]
    \centering
    \small 
    \setlength{\tabcolsep}{5pt} 
    \begin{tabular}{@{}lccccccccc@{}}
        \toprule
        \textbf{DeepSeek-Qwen-7B} & \textbf{3-star} & \textbf{triangle} & \textbf{4-path} & \textbf{4-cycle} & \textbf{4-chordalcycle} & \textbf{4-tailedtriangle} & \textbf{4-clique} & \textbf{bitriangle} & \textbf{butterfly} \\ 
        \midrule
        Zero-shot                 & 50\%             & \underline{50\%}             & \underline{65\%}             & \underline{50\%}             & \textbf{65\%}           & \underline{60\%}                 & \textbf{65\%}           & \underline{65\%}             & 50\%             \\
        One-shot                  & \underline{60\%}             & \underline{50\%}             & \underline{65\%}             & \underline{50\%}             & \textbf{65\%}           & \underline{60\%}                 & 35\%             & 50\%             & \underline{55\%}             \\
        Zero-shot+CoT             & 50\%             & \underline{50\%}             & \underline{65\%}             & \underline{50\%}             & \textbf{65\%}           & \underline{60\%}                 & \textbf{65\%}           & \underline{65\%}             & 50\%             \\
        One-shot+CoT              & \textbf{90\%}            & \textbf{95\%}           & \textbf{95\%}           & \textbf{65\%}           & \textbf{65\%}           & \textbf{90\%}            & \underline{45\%}             & \textbf{80\%}           & \textbf{90\%}            \\ 
        \bottomrule
    \end{tabular}
    \caption{Performance of different prompting strategies for the DeepSeek-R1-Distill-Qwen-7B (DeepSeek-Qwen-7B) model on the "Motif Classification" task. For each motif type (column), the best performance is marked in \textbf{bold} and the second best is \underline{underlined}.}
    \label{tab:prompting_qwen7b}
\end{table*}

\begin{table*}[p]
    \centering
    \small 
    \setlength{\tabcolsep}{5pt} 
    \begin{tabular}{@{}lccccccccc@{}}
        \toprule
        \textbf{DeepSeek-Qwen-14B} & \textbf{3-star} & \textbf{triangle} & \textbf{4-path} & \textbf{4-cycle} & \textbf{4-chordalcycle} & \textbf{4-tailedtriangle} & \textbf{4-clique} & \textbf{bitriangle} & \textbf{butterfly} \\ 
        \midrule
        Zero-shot                 & \textbf{100\%}          & \underline{90\%}            & \underline{95\%}            & \textbf{100\%}          & \underline{90\%}                 & \textbf{100\%}                 & \textbf{100\%}          & \textbf{100\%}          & 90\%             \\
        One-shot                  & \textbf{100\%}          & \textbf{100\%}          & \textbf{100\%}          & \textbf{100\%}          & \underline{95\%}                 & \textbf{100\%}                 & \textbf{100\%}          & \textbf{100\%}          & \underline{95\%}             \\
        Zero-shot+Cot             & \textbf{100\%}          & 85\%             & \textbf{100\%}          & \textbf{100\%}          & \textbf{100\%}          & \textbf{100\%}                 & \textbf{100\%}          & \textbf{100\%}          & \textbf{100\%}          \\
        One-shot+Cot              & \textbf{100\%}          & \textbf{100\%}          & \textbf{100\%}          & \textbf{100\%}          & \textbf{100\%}          & \textbf{100\%}                 & \textbf{100\%}          & \textbf{100\%}          & \underline{95\%}             \\ 
        \bottomrule
    \end{tabular}
    \caption{Performance of different prompting strategies on the "Motif Classification" task for the DeepSeek-R1-Distill-Qwen-14B (DeepSeek-Qwen-14B) model. For each motif type (column), the best performance is in \textbf{bold}, and the second best is \underline{underlined}.}
    \label{tab:prompting_qwen14b}
\end{table*}

\begin{table*}[b!]
    \centering
    \small 
    \setlength{\tabcolsep}{5pt} 
    \begin{tabular}{@{}lccccccccc@{}}
        \toprule
        \textbf{DeepSeek-Qwen-32B} & \textbf{3-star} & \textbf{triangle} & \textbf{4-path} & \textbf{4-cycle} & \textbf{4-chordalcycle} & \textbf{4-tailedtriangle} & \textbf{4-clique} & \textbf{bitriangle} & \textbf{butterfly} \\ 
        \midrule
        Zero-shot                 & \underline{95\%}            & \underline{95\%}            & \underline{95\%}            & 90\%             & \underline{95\%}                 & \textbf{100\%}                 & \textbf{100\%}          & \textbf{100\%}          & \underline{85\%}             \\
        One-shot                  & \textbf{100\%}          & 90\%             & \underline{95\%}            & \underline{95\%}            & \textbf{100\%}          & \textbf{100\%}                 & \textbf{100\%}          & \textbf{100\%}          & \textbf{95\%}            \\
        Zero-shot+CoT             & \textbf{100\%}          & \underline{95\%}            & 90\%             & \textbf{100\%}          & \underline{95\%}                 & \textbf{100\%}                 & \textbf{100\%}          & \textbf{100\%}          & \underline{85\%}             \\
        One-shot+CoT              & \textbf{100\%}          & \textbf{100\%}          & \textbf{100\%}          & \textbf{100\%}          & \textbf{100\%}          & \textbf{100\%}                 & \textbf{100\%}          & \textbf{100\%}          & \textbf{95\%}            \\ 
        \bottomrule
    \end{tabular}
    \caption{Performance of different prompting strategies on the "Motif Classification" task for the DeepSeek-R1-Distill-Qwen-32B (DeepSeek-Qwen-32B) model.}
    \label{tab:prompting_qwen32b}
\end{table*}

\begin{table*}[t!]
    \centering
    \small 
    \setlength{\tabcolsep}{5pt} 
    \begin{tabular}{@{}lccccccccc@{}}
        \toprule
        \textbf{Qwen2.5-32B-Instruct} & \textbf{3-star} & \textbf{triangle} & \textbf{4-path} & \textbf{4-cycle} & \textbf{4-chordalcycle} & \textbf{4-tailedtriangle} & \textbf{4-clique} & \textbf{bitriangle} & \textbf{butterfly} \\ 
        \midrule
        Zero-shot                 & \textbf{100\%}          & 80\%             & \underline{90\%}            & \underline{90\%}            & \textbf{95\%}           & \underline{90\%}                 & \underline{75\%}             & 70\%             & \underline{90\%}             \\
        One-shot                  & \textbf{100\%}          & 75\%             & 80\%             & 85\%             & \underline{80\%}                 & \underline{90\%}                 & 65\%             & 70\%             & 55\%             \\
        Zero-shot+CoT             & \textbf{100\%}          & \underline{85\%}             & \textbf{100\%}          & \textbf{100\%}          & \textbf{95\%}           & \underline{90\%}                 & \underline{75\%}             & \underline{85\%}             & \underline{90\%}             \\
        One-shot+CoT              & \textbf{100\%}          & \textbf{100\%}          & \textbf{100\%}          & \textbf{100\%}          & \textbf{95\%}           & \textbf{100\%}            & \textbf{90\%}            & \textbf{100\%}          & \textbf{100\%}           \\ 
        \bottomrule
    \end{tabular}
    \caption{Performance of different prompting strategies on the "Motif Classification" task for the Qwen2.5-32B-Instruct model. For each motif type (column), the best performance is in \textbf{bold}, and the second best is \underline{underlined}.}
    \label{tab:prompting_qwen2.5_32b}
\end{table*}

\begin{table*}[p!]
    \centering
    \small 
    \setlength{\tabcolsep}{5pt} 
    \begin{tabular}{@{}lccccccccc@{}}
        \toprule
        \textbf{QwQ-32B} & \textbf{3-star} & \textbf{triangle} & \textbf{4-path} & \textbf{4-cycle} & \textbf{4-chordalcycle} & \textbf{4-tailedtriangle} & \textbf{4-clique} & \textbf{bitriangle} & \textbf{butterfly} \\ 
        \midrule
        Zero-shot                 & \textbf{100\%}          & \underline{95\%}            & \underline{95\%}            & 90\%             & \textbf{100\%}          & 90\%              & \underline{95\%}             & \underline{90\%}             & 85\%             \\
        One-shot                  & \textbf{100\%}          & \textbf{100\%}          & \underline{95\%}            & \underline{95\%}             & \textbf{100\%}          & \underline{95\%}              & \underline{95\%}             & \underline{90\%}             & \underline{90\%}             \\
        Zero-shot+CoT             & \textbf{100\%}          & \underline{95\%}            & \textbf{100\%}          & \textbf{100\%}          & \textbf{100\%}          & \textbf{100\%}           & 85\%             & 85\%             & 85\%             \\
        One-shot+CoT              & \textbf{100\%}          & \textbf{100\%}          & \textbf{100\%}          & \textbf{100\%}          & \textbf{100\%}          & \textbf{100\%}           & \textbf{100\%}           & \textbf{100\%}          & \textbf{95\%}            \\ 
        \bottomrule
    \end{tabular}
    \caption{Performance of different prompting strategies on the "Motif Classification" task for the QwQ-32B model. For each motif type (column), the best performance is in \textbf{bold}, and the second best is \underline{underlined}.}
    \label{tab:prompting_qwq32b}
\end{table*}

\begin{table*}[t]
    \centering
    \small
    \setlength{\tabcolsep}{5pt}
    \begin{tabular}{@{}lccccc@{}}
        \toprule
        \textbf{DeepSeek-R1-Distill-Qwen-7B} & 
        \textbf{4-chordalcycle} & 
        \textbf{4-clique} & 
        \textbf{4-cycle} & 
        \textbf{4-tailedtriangle} & 
        \textbf{bitriangle} \\ 
        \midrule
        Zero-shot                 & 5\%              & 15\%             & 10\%             & 5\%              & \underline{25\%}             \\
        One-shot                  & 15\%             & 5\%              & \underline{15\%}             & \textbf{25\%}          & 10\%             \\
        Zero-shot+CoT             & \underline{30\%}             & \underline{20\%}             & 10\%             & \underline{20\%}             & \textbf{40\%}            \\
        One-shot+CoT              & \textbf{35\%}            & \textbf{25\%}          & \textbf{20\%}          & \underline{20\%}             & 15\%             \\ 
        \bottomrule
    \end{tabular}
    \caption{Performance of different prompting strategies on the "Motif Construction" task for the DeepSeek-R1-Distill-Qwen-7B model. For each motif type (column), the best performance is in \textbf{bold}, and the second best is \underline{underlined}.}
    \label{tab:modify_motif_qwen7b}
\end{table*}

\begin{table*}[t]
    \centering
    \small
    \setlength{\tabcolsep}{5pt}
    \begin{tabular}{@{}lccccc@{}}
        \toprule
        \textbf{DeepSeek-R1-Distill-Qwen-14B} & 
        \textbf{4-chordalcycle} & 
        \textbf{4-clique} & 
        \textbf{4-cycle} & 
        \textbf{4-tailedtriangle} & 
        \textbf{bitriangle} \\ 
        \midrule
        Zero-shot                 & 40\%             & \underline{65\%}             & 70\%             & 35\%             & \textbf{80\%}            \\
        One-shot                  & \underline{50\%}             & \underline{65\%}             & \underline{75\%}             & \underline{65\%}             & \textbf{80\%}            \\
        Zero-shot+CoT             & 40\%             & \underline{65\%}             & \textbf{85\%}          & 55\%             & \underline{75\%}             \\
        One-shot+CoT              & \textbf{65\%}            & \textbf{70\%}          & \textbf{85\%}          & \textbf{75\%}          & 70\%             \\ 
        \bottomrule
    \end{tabular}
    \caption{Performance of different prompting strategies on the "Motif Construction" task for the DeepSeek-R1-Distill-Qwen-14B model. For each motif type (column), the best performance is in \textbf{bold}, and the second best is \underline{underlined}.}
    \label{tab:modify_motif_qwen14b}
\end{table*}

\begin{table*}[t]
    \centering
    \small
    \setlength{\tabcolsep}{5pt}
    \begin{tabular}{@{}lccccc@{}}
        \toprule
        \textbf{DeepSeek-R1-Distill-Qwen-32B} & 
        \textbf{4-chordalcycle} & 
        \textbf{4-clique} & 
        \textbf{4-cycle} & 
        \textbf{4-tailedtriangle} & 
        \textbf{bitriangle} \\ 
        \midrule
        Zero-shot                 & 55\%             & 65\%             & \underline{80\%}             & 60\%             & 50\%             \\
        One-shot                  & 40\%             & \underline{75\%}             & 40\%             & 35\%             & \underline{75\%}             \\
        Zero-shot+CoT             & \textbf{80\%}            & \textbf{90\%}          & \textbf{95\%}          & \underline{75\%}             & 45\%             \\
        One-shot+CoT              & \underline{70\%}             & \textbf{90\%}          & 65\%             & \textbf{90\%}          & \textbf{85\%}            \\ 
        \bottomrule
    \end{tabular}
    \caption{Performance of different prompting strategies on the "Motif Construction" task for the DeepSeek-R1-Distill-Qwen-32B model. For each motif type (column), the best performance is in \textbf{bold}, and the second best is \underline{underlined}.}
    \label{tab:modify_motif_qwen32b}
\end{table*}

\begin{table*}[t]
    \centering
    \small
    \setlength{\tabcolsep}{5pt}
    \begin{tabular}{@{}lccccc@{}}
        \toprule
        \textbf{Qwen2.5-32B-Instruct} & 
        \textbf{4-chordalcycle} & 
        \textbf{4-clique} & 
        \textbf{4-cycle} & 
        \textbf{4-tailedtriangle} & 
        \textbf{bitriangle} \\ 
        \midrule
        Zero-shot                 & \textbf{65\%}            & 20\%             & 30\%             & \underline{60\%}             & 55\%             \\
        One-shot                  & 35\%             & 30\%             & 25\%             & 35\%             & \textbf{90\%}            \\
        Zero-shot+CoT             & \underline{50\%}             & \underline{75\%}             & \underline{55\%}             & 55\%             & \underline{85\%}             \\
        One-shot+CoT              & \underline{50\%}             & \textbf{90\%}          & \textbf{60\%}          & \textbf{65\%}          & 75\%             \\ 
        \bottomrule
    \end{tabular}
    \caption{Performance of different prompting strategies on the "Motif Construction" task for the Qwen2.5-32B-Instruct model. For each motif type (column), the best performance is in \textbf{bold}, and the second best is \underline{underlined}.}
    \label{tab:modify_motif_qwen2.5_32b}
\end{table*}

\begin{table*}[t]
    \centering
    \small
    \setlength{\tabcolsep}{5pt}
    \begin{tabular}{@{}lccccc@{}}
        \toprule
        \textbf{QwQ-32B} & 
        \textbf{4-chordalcycle} & 
        \textbf{4-clique} & 
        \textbf{4-cycle} & 
        \textbf{4-tailedtriangle} & 
        \textbf{bitriangle} \\ 
        \midrule
        Zero-shot                 & 45\%             & 45\%             & 70\%             & 40\%             & 30\%             \\
        One-shot                  & \underline{70\%}             & \underline{65\%}             & 50\%             & 70\%             & \underline{45\%}             \\
        Zero-shot+CoT             & \textbf{80\%}            & \textbf{90\%}          & \textbf{85\%}          & \textbf{95\%}          & \textbf{80\%}            \\
        One-shot+CoT              & \underline{70\%}             & 60\%             & \underline{80\%}             & \underline{85\%}             & \textbf{80\%}            \\ 
        \bottomrule
    \end{tabular}
    \caption{Performance of different prompting strategies on the "Motif Construction" task for the QwQ-32B model. For each motif type (column), the best performance is in \textbf{bold}, and the second best is \underline{underlined}.}
    \label{tab:modify_motif_qwq32b}
\end{table*}

\begin{table*}[t]
    \centering
    \small 
    \setlength{\tabcolsep}{5pt} 
    \begin{tabular}{@{}lccccccccc@{}}
        \toprule
        \textbf{DeepSeek-Qwen-7B} & \textbf{3-star} & \textbf{triangle} & \textbf{4-path} & \textbf{4-cycle} & \textbf{4-chordalcycle} & \textbf{4-tailedtriangle} & \textbf{4-clique} & \textbf{bitriangle} & \textbf{butterfly} \\ 
        \midrule
        Zero-shot                 & 50\%           & 20\%             & 35\%             & 30\%             & 15\%              & 20\%              & \underline{30\%}             & 5\%             & 0\%             \\
        One-shot                  & \textbf{85\%}  & \textbf{70\%}    & \textbf{80\%}    & \textbf{55\%}    & \textbf{65\%}     & \textbf{60\%}     & 25\%             & 5\%             & \underline{10\%}            \\
        Zero-shot+CoT             & \underline{60\%} & 45\%             & 50\%             & \underline{50\%} & 15\%              & 10\%              & \underline{30\%}             & \underline{20\%}             & \textbf{20\%}            \\
        One-shot+CoT              & \textbf{85\%}           & \underline{65\%} & \underline{65\%} & \textbf{55\%}             & \underline{45\%}  & \underline{35\%}  & \textbf{35\%}    & \textbf{35\%}    & \textbf{20\%}             \\ 
        \bottomrule
    \end{tabular}
    \caption{Performance of different prompting strategies on the "Motif Detection" task for the DeepSeek-R1-Distill-Qwen-7B (DeepSeek-Qwen-7B) model. For each motif type (column), the best performance is in \textbf{bold}, and the second best is \underline{underlined}.}
    \label{tab:contain_motif_qwen7b}
\end{table*}

\begin{table*}[t]
    \centering
    \small 
    \setlength{\tabcolsep}{5pt} 
    \begin{tabular}{@{}lccccccccc@{}}
        \toprule
        \textbf{DeepSeek-Qwen-14B} & \textbf{3-star} & \textbf{triangle} & \textbf{4-path} & \textbf{4-cycle} & \textbf{4-chordalcycle} & \textbf{4-tailedtriangle} & \textbf{4-clique} & \textbf{bitriangle} & \textbf{butterfly} \\ 
        \midrule
        Zero-shot                 & \textbf{90\%}           & 80\%             & 75\%             & 30\%             & \underline{30\%}              & 35\%              & 25\%             & \underline{35\%}             & \underline{40\%}             \\
        One-shot                  & \underline{85\%}            & \textbf{90\%}          & \underline{85\%}             & 30\%             & 25\%              & 35\%              & \underline{40\%}             & 20\%             & \textbf{45\%}            \\
        Zero-shot+CoT             & \textbf{90\%}           & 65\%             & 80\%             & \underline{35\%}             & 10\%              & \underline{45\%}              & 20\%             & 25\%             & 25\%             \\
        One-shot+CoT              & \underline{85\%}            & \underline{85\%}             & \textbf{90\%}          & \textbf{45\%}          & \textbf{50\%}           & \textbf{55\%}           & \textbf{45\%}          & \textbf{45\%}          & 30\%             \\ 
        \bottomrule
    \end{tabular}
    \caption{Performance of different prompting strategies on the "Motif Detection" task for the DeepSeek-R1-Distill-Qwen-14B (DeepSeek-Qwen-14B) model. For each motif type (column), the best performance is in \textbf{bold}, and the second best is \underline{underlined}.}
    \label{tab:contain_motif_qwen14b}
\end{table*}

\begin{table*}[t]
    \centering
    \small 
    \setlength{\tabcolsep}{5pt} 
    \begin{tabular}{@{}lccccccccc@{}}
        \toprule
        \textbf{DeepSeek-Qwen-32B} & \textbf{3-star} & \textbf{triangle} & \textbf{4-path} & \textbf{4-cycle} & \textbf{4-chordalcycle} & \textbf{4-tailedtriangle} & \textbf{4-clique} & \textbf{bitriangle} & \textbf{butterfly} \\ 
        \midrule
        Zero-shot                 & \textbf{90\%}           & 85\%             & \underline{80\%}             & 45\%             & 10\%              & 55\%              & 40\%             & 35\%             & 35\%             \\
        One-shot                  & \textbf{90\%}           & \underline{90\%}             & 70\%             & \underline{50\%}             & 20\%              & \underline{60\%}              & \underline{45\%}             & \underline{40\%}             & 45\%             \\
        Zero-shot+CoT             & \underline{80\%}             & 75\%             & \textbf{85\%}          & \textbf{80\%}          & \textbf{40\%}           & 35\%              & \underline{45\%}             & \underline{40\%}             & \textbf{70\%}            \\
        One-shot+CoT              & \textbf{90\%}           & \textbf{95\%}          & \textbf{85\%}          & \underline{50\%}             & \underline{35\%}              & \textbf{65\%}           & \textbf{55\%}          & \textbf{45\%}          & \underline{55\%}             \\ 
        \bottomrule
    \end{tabular}
    \caption{Performance of different prompting strategies on the "Motif Detection" task for the DeepSeek-R1-Distill-Qwen-32B (DeepSeek-Qwen-32B) model. For each motif type (column), the best performance is in \textbf{bold}, and the second best is \underline{underlined}.}
    \label{tab:contain_motif_qwen32b}
\end{table*}

\begin{table*}[t]
    \centering
    \small 
    \setlength{\tabcolsep}{5pt} 
    \begin{tabular}{@{}lccccccccc@{}}
        \toprule
        \textbf{Qwen2.5-32B-Instruct} & \textbf{3-star} & \textbf{triangle} & \textbf{4-path} & \textbf{4-cycle} & \textbf{4-chordalcycle} & \textbf{4-tailedtriangle} & \textbf{4-clique} & \textbf{bitriangle} & \textbf{butterfly} \\ 
        \midrule
        Zero-shot                 & \underline{75\%}             & \underline{70\%}             & \underline{75\%}             & \underline{50\%}             & \textbf{55\%}           & \underline{45\%}              & 20\%             & \underline{45\%}             & \underline{35\%}             \\
        One-shot                  & \textbf{85\%}           & 65\%             & \underline{75\%}             & \textbf{60\%}           & \textbf{55\%}           & \underline{45\%}              & 10\%             & 30\%             & \textbf{50\%}            \\
        Zero-shot+CoT             & \textbf{85\%}           & \underline{70\%}             & \underline{75\%}             & \textbf{60\%}           & \underline{40\%}              & \underline{45\%}              & \underline{55\%}             & \textbf{60\%}          & \underline{35\%}             \\
        One-shot+CoT              & \textbf{85\%}           & \textbf{80\%}          & \textbf{80\%}          & \underline{50\%}             & \textbf{55\%}           & \textbf{65\%}           & \textbf{60\%}          & \underline{45\%}             & \textbf{50\%}            \\ 
        \bottomrule
    \end{tabular}
    \caption{Performance of different prompting strategies on the "Motif Detection" task for the Qwen2.5-32B-Instruct model. For each motif type (column), the best performance is in \textbf{bold}, and the second best is \underline{underlined}.}
    \label{tab:contain_motif_qwen2.5_32b}
\end{table*}

\begin{table*}[t]
    \centering
    \small 
    \setlength{\tabcolsep}{5pt} 
    \begin{tabular}{@{}lccccccccc@{}}
        \toprule
        \textbf{QwQ-32B} & \textbf{3-star} & \textbf{triangle} & \textbf{4-path} & \textbf{4-cycle} & \textbf{4-chordalcycle} & \textbf{4-tailedtriangle} & \textbf{4-clique} & \textbf{bitriangle} & \textbf{butterfly} \\ 
        \midrule
        Zero-shot                 & \underline{85\%}            & \underline{85\%}             & 70\%             & \underline{50\%}             & \underline{5\%}               & 25\%              & \underline{5\%}              & \underline{10\%}             & 15\%             \\
        One-shot                  & \underline{85\%}            & 80\%             & \underline{75\%}             & 30\%             & \textbf{10\%}           & \textbf{45\%}           & \underline{5\%}              & \textbf{15\%}          & \underline{25\%}             \\
        Zero-shot+CoT             & \underline{85\%}            & 80\%             & \textbf{80\%}          & 35\%             & \underline{5\%}               & \underline{35\%}              & \textbf{10\%}           & \underline{10\%}             & \textbf{30\%}            \\
        One-shot+CoT              & \textbf{90\%}           & \textbf{95\%}          & \underline{75\%}             & \textbf{60\%}          & \textbf{10\%}           & \textbf{45\%}           & \underline{5\%}              & \textbf{15\%}          & \textbf{30\%}            \\ 
        \bottomrule
    \end{tabular}
    \caption{Performance of different prompting strategies on the "Motif Detection" task for the QwQ-32B model. For each motif type (column), the best performance is in \textbf{bold}, and the second best is \underline{underlined}.}
    \label{tab:contain_motif_qwq32b}
\end{table*}

\setlength{\intextsep}{20pt}
\setlength{\textfloatsep}{20pt}
\begin{table}[t!]
    \centering
    \small 
    \setlength{\tabcolsep}{3pt} 
    \begin{tabular}{@{}lccc@{}}
        \toprule
        \textbf{DeepSeek-Qwen-7B} & 
        \textbf{\begin{tabular}[c]{@{}c@{}}Motif Detect\end{tabular}} & 
        \textbf{\begin{tabular}[c]{@{}c@{}}Motif Occur\end{tabular}} & 
        \textbf{\begin{tabular}[c]{@{}c@{}}Motif Count\end{tabular}} \\ 
        \midrule
        Zero-shot                 & \underline{11.19\%}      & 0.99\%              & 2.00\%               \\
        One-shot                  & 3.60\%               & \textbf{4.22\%}           & \textbf{15.64\%}         \\
        Zero-shot+CoT             & 9.86\%               & 1.78\%              & 3.43\%               \\
        One-shot+CoT              & \textbf{12.00\%}         & \underline{3.00\%}              & \underline{15.46\%}      \\ 
        \bottomrule
    \end{tabular}
    \caption{Performance of different prompting strategies on Level 2 tasks for the DeepSeek-R1-Distill-Qwen-7B (DeepSeek-Qwen-7B) model. We use
“Multi Detect,” “Motif Occur,” and “Multi Count” as shorthand for the ”Multi-Motif Detection”, ”Motif Occurrence
Prediction”, and ”Multi-Motif Count” tasks, respectively. For each task (column), the best performance is in \textbf{bold}, and the second best is \underline{underlined}.}
    \label{tab:level2_qwen7b}
\end{table}

\begin{table}[hp]
    \centering
    \small 
    \setlength{\tabcolsep}{5pt} 
    \begin{tabular}{@{}lccc@{}}
        \toprule
        \textbf{DeepSeek-Qwen-14B} & 
        \textbf{\begin{tabular}[c]{@{}c@{}}Motif Detect\end{tabular}} & 
        \textbf{\begin{tabular}[c]{@{}c@{}}Motif Occur\end{tabular}} & 
        \textbf{\begin{tabular}[c]{@{}c@{}}Motif Count\end{tabular}} \\ 
        \midrule
        Zero-shot                 & \textbf{23.67\%}         & \textbf{10.20\%}        & \textbf{6.08\%}          \\
        One-shot                  & 14.88\%              & 8.19\%              & 4.48\%               \\
        Zero-shot+CoT             & 19.36\%              & \underline{9.95\%}              & \underline{6.07\%}               \\
        One-shot+CoT              & \underline{20.09\%}      & 8.21\%              & 4.35\%               \\ 
        \bottomrule
    \end{tabular}
    \caption{Performance of different prompting strategies on Level 2 tasks for the DeepSeek-R1-Distill-Qwen-14B (DeepSeek-Qwen-14B) model. For each task (column), the best performance is in \textbf{bold}, and the second best is \underline{underlined}.}
    \label{tab:level2_qwen14b}
\end{table}

\begin{table}[hp]
    \centering
    \small 
    \setlength{\tabcolsep}{5pt} 
    \begin{tabular}{@{}lccc@{}}
        \toprule
        \textbf{DeepSeek-Qwen-32B} & 
        \textbf{\begin{tabular}[c]{@{}c@{}}Motif Detect\end{tabular}} & 
        \textbf{\begin{tabular}[c]{@{}c@{}}Motif Occur\end{tabular}} & 
        \textbf{\begin{tabular}[c]{@{}c@{}}Motif Count\end{tabular}} \\ 
        \midrule
        Zero-shot                 & 24.91\%              & 8.73\%              & 5.40\%               \\
        One-shot                  & \underline{28.79\%}      & \underline{10.50\%}     & 3.97\%               \\
        Zero-shot+CoT             & \textbf{29.83\%}         & 9.44\%              & \underline{6.19\%}               \\
        One-shot+CoT              & 27.87\%              & \textbf{10.66\%}        & \textbf{9.20\%}          \\ 
        \bottomrule
    \end{tabular}
    \caption{Performance of different prompting strategies on Level 2 tasks for the DeepSeek-R1-Distill-Qwen-32B (DeepSeek-Qwen-32B) model. For each task (column), the best performance is in \textbf{bold}, and the second best is \underline{underlined}.}
    \label{tab:level2_qwen32b}
\end{table}

\begin{table}[hbp!]
    \centering
    \small 
    \setlength{\tabcolsep}{5pt} 
    \begin{tabular}{@{}lccc@{}}
        \toprule
        \textbf{Qwen2.5-32B} & 
        \textbf{\begin{tabular}[c]{@{}c@{}}Motif Detect\end{tabular}} & 
        \textbf{\begin{tabular}[c]{@{}c@{}}Motif Occur\end{tabular}} & 
        \textbf{\begin{tabular}[c]{@{}c@{}}Motif Count\end{tabular}} \\ 
        \midrule
        Zero-shot                 & 23.88\%              & \underline{11.48\%}     & \underline{19.69\%}      \\
        One-shot                  & \underline{24.99\%}      & \textbf{13.52\%}        & \textbf{25.05\%}         \\
        Zero-shot+CoT             & 21.82\%              & 9.34\%              & 7.94\%               \\
        One-shot+CoT              & \textbf{30.25\%}         & 10.90\%             & 6.00\%               \\ 
        \bottomrule
    \end{tabular}
    \caption{Performance of different prompting strategies on Level 2 tasks for the Qwen2.5-32B-Instruct model (Qwen2.5-32B). For each task (column), the best performance is in \textbf{bold}, and the second best is \underline{underlined}.}
    \label{tab:level2_qwen2.5_32b}
\end{table}

\begin{table}[hbpt]
    \centering
    \small 
    \setlength{\tabcolsep}{5pt} 
    \begin{tabular}{@{}lccc@{}}
        \toprule
        \textbf{QwQ-32B} & 
        \textbf{\begin{tabular}[c]{@{}c@{}}Motif Detect\end{tabular}} & 
        \textbf{\begin{tabular}[c]{@{}c@{}}Motif Occur\end{tabular}} & 
        \textbf{\begin{tabular}[c]{@{}c@{}}Motif Count\end{tabular}} \\ 
        \midrule
        Zero-shot                 & \underline{23.19\%}      & 4.02\%              & \underline{0.40\%}               \\
        One-shot                  & 22.35\%              & 3.74\%              & 0.32\%                     \\
        Zero-shot+CoT             & \textbf{29.69\%}         & \underline{6.08\%}              & 0.25\%                    \\
        One-shot+CoT              & 11.51\%              & \textbf{6.84\%}         & \textbf{0.51\%}                   \\ 
        \bottomrule
    \end{tabular}
    \caption{Performance of different prompting strategies on Level 2 tasks for the QwQ-32B model. For each task (column), the best performance is in \textbf{bold}, and the second best is \underline{underlined}.}
    \label{tab:level2_qwq32b}
\end{table}

\subsection{Agent Performance Visualized}
\label{app:radas_graph}
\begin{figure}[htbp]
    \centering
    
    \begin{subfigure}{0.48\columnwidth}
        \centering
        \includegraphics[width=\linewidth]{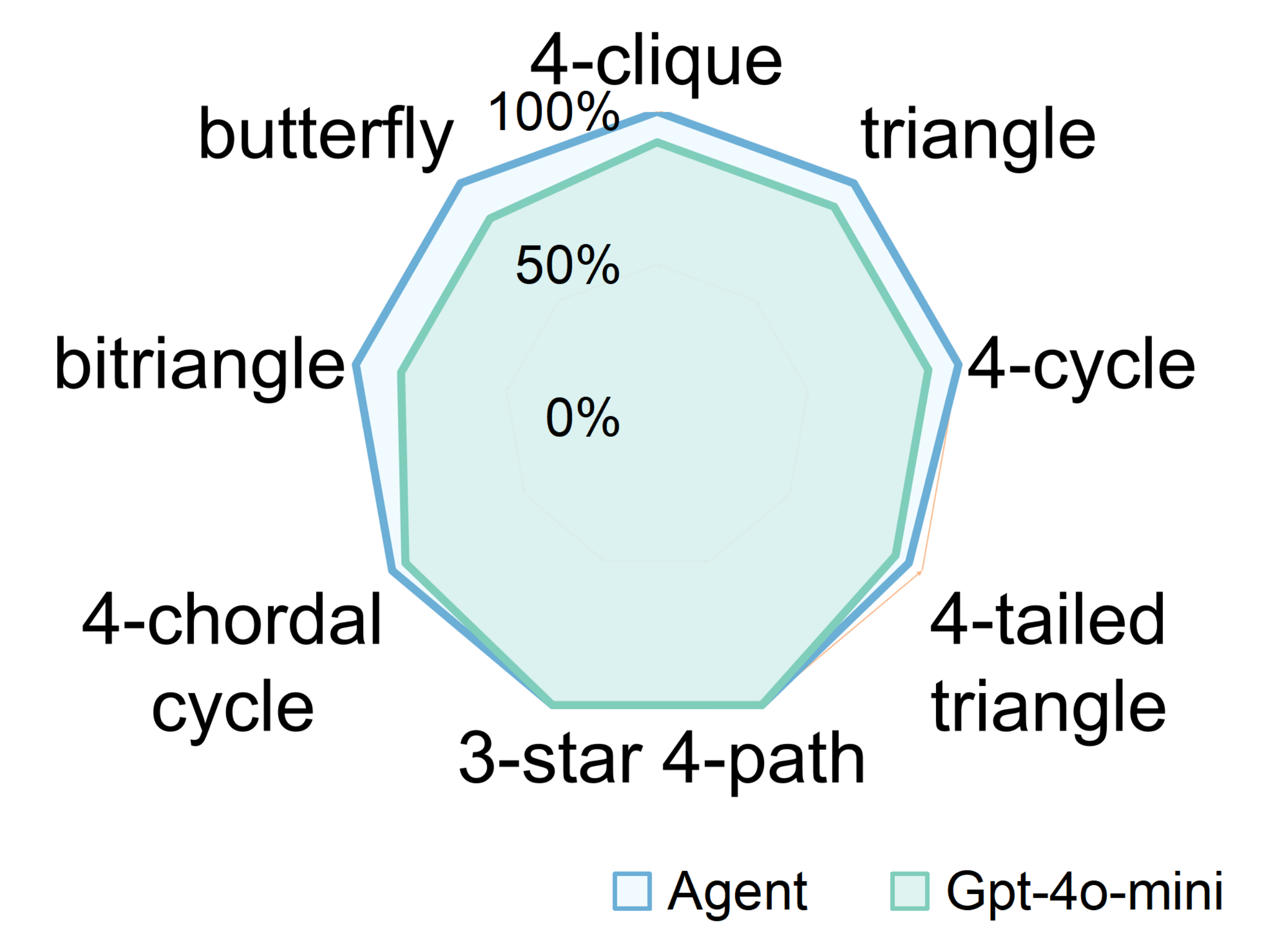} 
        \caption{accuracy}
        \label{fig:sub1_final}
    \end{subfigure}
    \hfill 
    \begin{subfigure}{0.48\columnwidth}
        \centering
        \includegraphics[width=\linewidth]{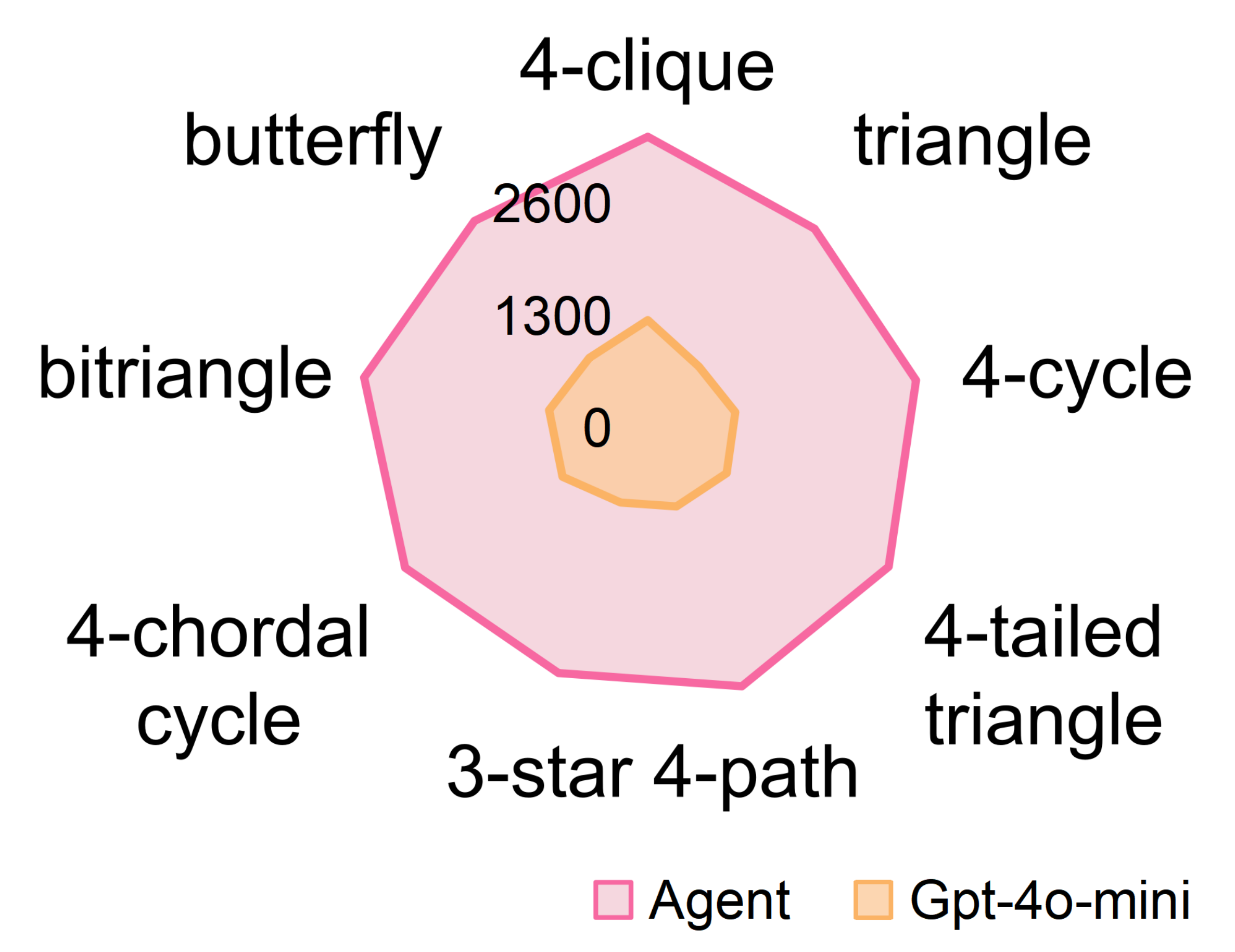} 
        \caption{average token consumption}
        \label{fig:sub2_final}
    \end{subfigure}
    
    \caption{Performance of the tool-augmented Agent versus GPT-4o-mini on the "Motif Classification" task, comparing (a) accuracy and (b) token consumption.}
    \label{fig:Performance of the Tool-augmented LLM agent on Judge Is Motif task}
\end{figure}

\begin{figure}[htbp]
    \centering
    
    \begin{subfigure}{0.48\columnwidth}
        \centering
        \includegraphics[width=\linewidth]{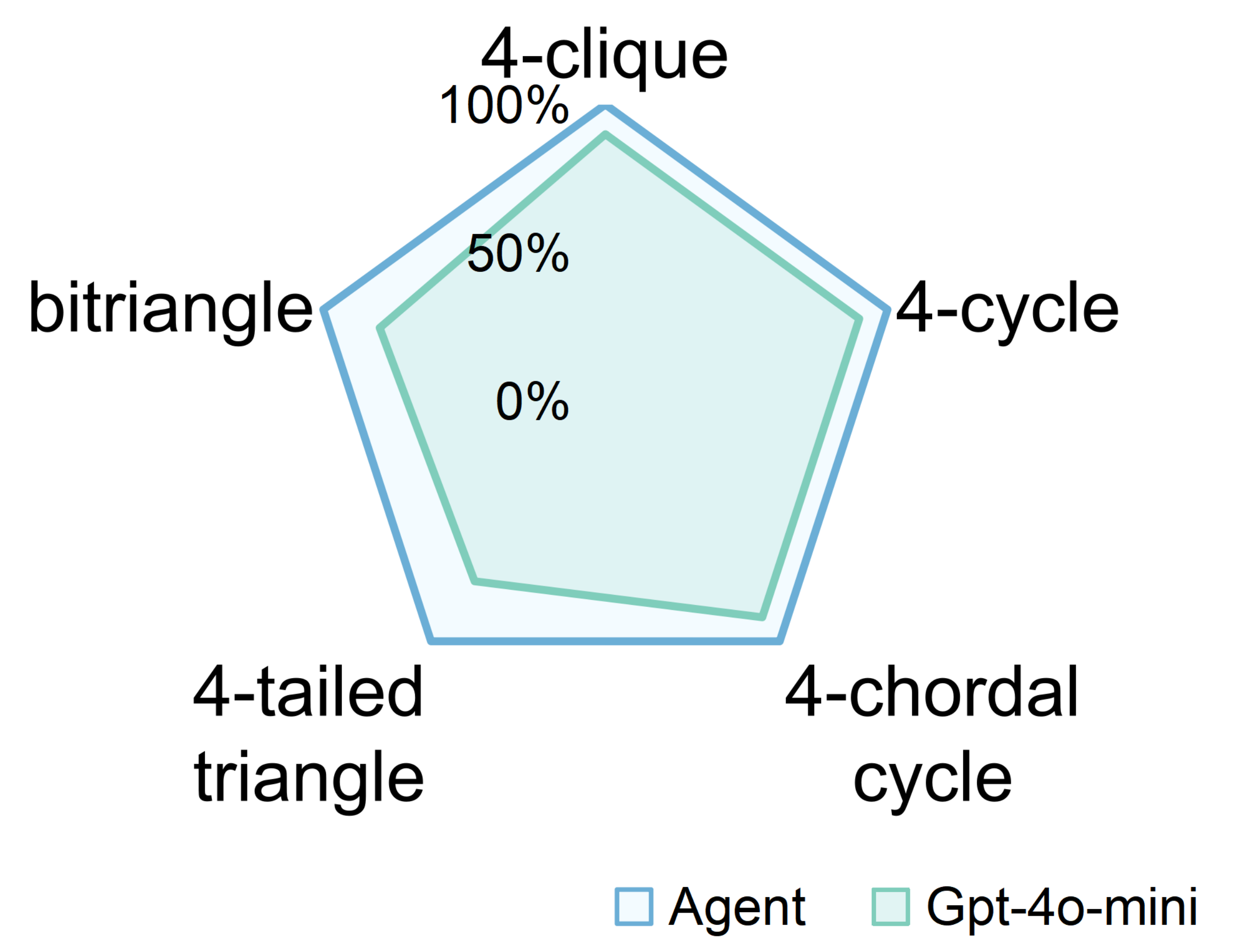} 
        \caption{accuracy}
        \label{fig:sub1_final}
    \end{subfigure}
    \hfill 
    \begin{subfigure}{0.48\columnwidth}
        \centering
        \includegraphics[width=\linewidth]{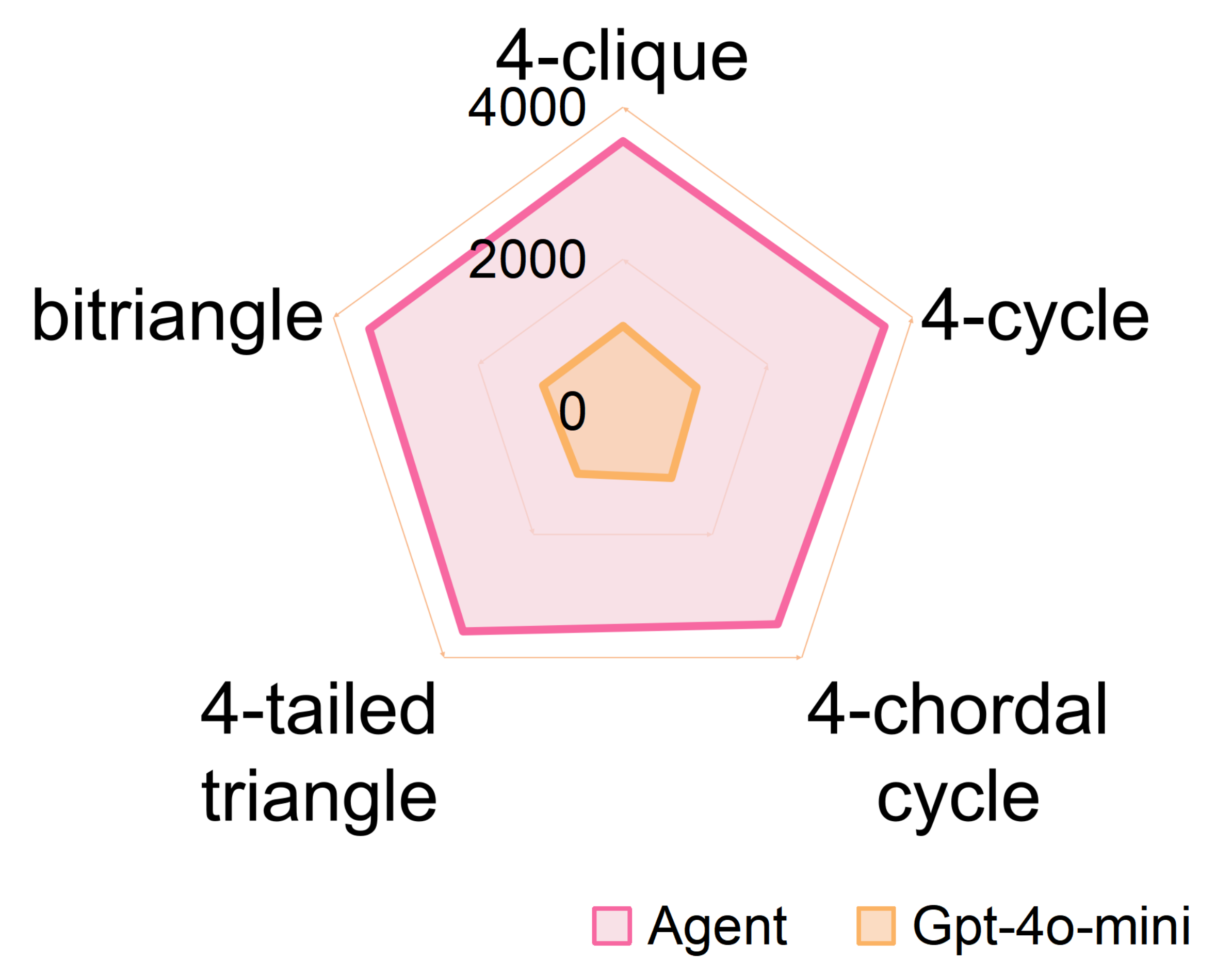} 
        \caption{average token consumption}
        \label{fig:sub2_final}
    \end{subfigure}
    
    \caption{Performance of the tool-augmented Agent versus GPT-4o-mini on the "Motif Construction" task, comparing (a) accuracy and (b) token consumption.}
    \label{fig:Performance of the Tool-augmented LLM agent on Modify Motif task}
\end{figure}

We supplement the comparison between our tool-augmented LLM agent and the baseline GPT-4o-mini with results on the "Motif Classification" and "Motif Construction" tasks, as shown in Figures~\ref{fig:Performance of the Tool-augmented LLM agent on Judge Is Motif task} and~\ref{fig:Performance of the Tool-augmented LLM agent on Modify Motif task}. The comparison shows that while the agent solves these tasks with high accuracy, it does so at a significant computational cost.

\subsection{Real-world Dataset}
\label{app:real-world}
 \textbf{Enron} ~\cite{Shetty2004TheEE}: is an email correspondence dataset containing around 50K emails exchanged among employees of the ENRON energy company over a three-year period.

We conducted the "Motif Detection" experiment on the Enron dataset. As illustrated in Table~\ref{tab:motif_detection_enron}, our structure-aware dispatcher pipeline remains effective at balancing accuracy and cost on real-world dataset.
\begin{table*}[htbp]
\centering


\small 
\setlength{\tabcolsep}{5pt} 

\begin{tabular}{l rr rr rr}
\toprule

Motif & \multicolumn{2}{c}{Gpt-4o-mini} & \multicolumn{2}{c}{\textbf{Ours}} & \multicolumn{2}{c}{Agent} \\
\cmidrule(lr){2-3} \cmidrule(lr){4-5} \cmidrule(lr){6-7}
 & Acc & Avg\_tokens & Acc & Avg\_tokens & Acc & Avg\_tokens \\
\midrule

3-star & 75\% & \textbf{1135.65} & \underline{85\%} & \underline{1697.25} & \textbf{100\%} & 3038.35 \\
4-cycle & 55\% & \textbf{1248.15} & \underline{90\%} & \underline{2701.85} & \textbf{100\%} & 3436.40 \\
4-clique & 70\% & \textbf{1633.55} & \underline{85\%} & \underline{3596.05} & \textbf{100\%} & 5966.70 \\
bitrangle & 50\% & \textbf{1486.70} & \underline{95\%} & \underline{3707.70} & \textbf{100\%} & 4330.85 \\
4-chordalcycle & \underline{55\%} & \textbf{1474.50} & \textbf{100\%} & \underline{3402.90} & \textbf{100\%} & 4283.40 \\
4-tailedtriangle & 45\% & \textbf{1220.10} & \underline{90\%} & \underline{2841.55} & \textbf{100\%} & 3478.55 \\
triangle & 70\% & \textbf{1048.60} & \underline{85\%} & \underline{1596.55} & \textbf{100\%} & 3073.80 \\
\bottomrule
\end{tabular}
\caption{Performance comparison of our Structure-Aware Dispatcher pipeline against several baselines on the real-world Enron dataset. For each task (row), the best performance is in bold, and the second-best is \underline{underlined}.}
\label{tab:motif_detection_enron} 
\end{table*}

\subsection{Performance of Deep Learning-based Methods on Motif Detection}
\label{app:deep_learning}
We applied TGN \cite{rossi2020temporalgraphnetworksdeep} and GIN \cite{xu2019powerfulgraphneuralnetworks} on the Motif Detection task. As shown in Table~\ref{tab:motif detection_dl_baselines}, their performance was unsatisfactory even with 20\% of the dataset allocated for training.

\subsection{openPangu-7B as the LLM for Structure-Aware Dispatcher}
\label{app:pangu-dispatcher}

We conducted experiments using openPangu-7B-DeepDive as the LLM for the Structure-Aware Dispatcher. As shown in Table~\ref{tab:pangu_dis}, this further demonstrates that our Structure-Aware Dispatcher pipeline can effectively balance accuracy and cost, exhibiting high cost-effectiveness and generalizability.

\begin{table*}[t] 
\centering
\begin{tabular}{l cc cc cc cc}
\toprule
Task & \multicolumn{2}{c}{Random} & \multicolumn{2}{c}{openPangu-7B} & \multicolumn{2}{c}{\textbf{Ours}} & \multicolumn{2}{c}{Agent} \\
\cmidrule(lr){2-3} \cmidrule(lr){4-5} \cmidrule(lr){6-7} \cmidrule(lr){8-9}
& Acc & Avg\_tokens & Acc & Avg\_tokens & Acc & Avg\_tokens & Acc & Avg\_tokens \\
\midrule
3-star & 66.0\% & \underline{4846.76} & 46.0\% & \textbf{1489.74} & \underline{78.0\%} & 5905.50 & \textbf{96.5\%} & 8377.64 \\
4-cycle & 66.0\% & \underline{5011.14} & 44.5\% & \textbf{2316.68} & \underline{77.0\%} & 7041.69 & \textbf{95.5\%} & 8942.82 \\
4-clique & 67.0\% & \underline{5681.81} & 43.0\% & \textbf{2552.86} & \underline{86.5\%} & 7247.80 & \textbf{94.5\%} & 8436.14 \\
bitriangle & 66.5\% & \underline{5371.50} & 45.0\% & \textbf{2261.39} & \underline{80.5\%} & 7117.05 & \textbf{92.5\%} & 8607.33 \\
4-chordalcycle & 70.0\% & \underline{5623.86} & 45.5\% & \textbf{2428.70} & \underline{86.0\%} & 7452.23 & \textbf{94.5\%} & 8963.83 \\
4-tailedtriangle & 69.4\% & \underline{5184.09} & 53.5\% & \textbf{1812.77} & \underline{82.5\%} & 6855.78 & \textbf{94.0\%} & 9178.86 \\
triangle & 69.0\% & \underline{5038.77} & 48.4\% & \textbf{1817.19} & \underline{84.0\%} & 6712.74 & \textbf{93.4\%} & 8957.78 \\
\bottomrule
\end{tabular}
\caption{Performance comparison of the Structure-Aware Dispatcher using openPangu-7B as the LLM against several baselines. The ”Random” baseline refers to randomly
assigning instances. This highlights how our method
consistently achieves the second-best accuracy while being significantly more cost-effective than the top-performing ”Agent” baseline.
We held out the ”4-tailedtriangle” and ”triangle” motifs from the training set to specifically evaluate generalization.}
\label{tab:pangu_dis} 
\end{table*}

\section{Metrics Formula}
\label{formula}

The formulas for the Structure-Aware Dispatcher's metrics are as follows:

\textbf{Cyclomatic Complexity}
measures structural complexity by calculating the number of independent cycles in the graph.
\[
C = E - N + P
\]
where $E$ is the number of edges, $N$ is the number of nodes, and $P$ is the number of connected components.

\textbf{Node Degree Ratios} also measures structural complexity
\begin{itemize}
    \item ratio\_nodes\_eq\_2: The ratio of nodes with a degree of 2, representing low-complexity parts.
    \item ratio\_nodes\_ge\_3: The ratio of nodes with a degree of 3 or more, representing high-complexity "intersections".
\end{itemize}

\textbf{Edge Locality Score} measures the dispersion of edges by calculating the standard deviation of edge indices connected to core nodes.
\[
\text{Edge\_Locality\_Score} = \frac{1}{|V'_{core}|} \sum_{v \in V'_{core}} \text{Std}(S_v)
\]
where $V'_{core}$ is the set of core nodes (degree $\ge$ 2), and $S_v$ is the set of index for edges connected to node $v$.

\begin{table*}[htbp]
\centering


\small 
\setlength{\tabcolsep}{4.5pt} 

\begin{tabular}{@{}l ccccccc @{}}
\toprule
Model & 3-star & 4-cycle & 4-clique & bitriangle & 4-chordalcycle & 4-tailedtriangle & triangle \\ 
\midrule

TGN & \underline{65.0\%} & 67.2\% & \underline{71.0\%} & \underline{50.0\%} & 56.3\% & 73.0\% & \underline{68.8\%} \\
GIN & 60.0\% & \underline{69.2\%} & 37.5\% & \underline{50.0\%} & \underline{68.8\%} & \underline{87.5\%} & 43.8\% \\
\textbf{Ours} & \textbf{88.0\%} & \textbf{92.5\%} & \textbf{97.5\%} & \textbf{95.0\%} & \textbf{96.0\%} & \textbf{88.6\%} & \textbf{84.8\%} \\
\bottomrule
\end{tabular}
\caption{Performance comparison on the "Motif Detection" task against deep learning baselines (TGN, GIN). 
The best performance in each column is in \textbf{bold}, and the second best is \underline{underlined}.}
\label{tab:motif detection_dl_baselines} 
\end{table*}

\begin{table*}[t]
    \centering
    \small 
    \setlength{\tabcolsep}{5pt} 
    \begin{tabular}{l cccc}
        \toprule
        \textbf{Model} & 
        \textbf{Sort Edge} & 
        \textbf{\begin{tabular}[c]{@{}c@{}}When Link \\ and Dislink\end{tabular}} & 
        \textbf{What Edges} & 
        \textbf{Reverse Graph} \\
        \midrule

        openPangu-7B          & 21\%             & 45\%             & 32\%             & 31\%             \\ 
        DeepSeek-R1-Distill-Qwen-7B & 39\% & 57\% & 34\% & 59\% \\
        DeepSeek-R1-Distill-Qwen-14B & 80\% & 90\% & 94\% & 71\% \\
        DeepSeek-R1-Distill-Qwen-32B & 92\% & 100\% & 93\% & 75\% \\
        Qwen2.5-32B-Instruct & 46\% & 93\% & 38\% & 50\% \\
        QwQ-32B & 87\% & 100\% & 93\% & 94\% \\
        GPT-4o-mini & 17\% & 98\% & 71\% & 20\% \\
        o3 & 100\% & 100\% & 96\% & 93\% \\
        Deepseek-R1 & 94\% & 100\% & 100\% & 100\% \\

        \midrule

        Random Baseline & 
        $\frac{1}{M!} \approx 1.147 \times 10^{-11}$ &
        $\binom{T}{2} = 10$ &
        $\frac{1}{\sum_{i=1}^{M} i!} \approx 1.065 \times 10^{-11}$ &
        $\frac{1}{M!} \approx 1.147 \times 10^{-11}$ \\
        
        \bottomrule
    \end{tabular}
     \caption{Performance comparison on Level 0 tasks and the random baseline. For each task (column), the best performance is in \textbf{bold}, and the second best is \underline{underlined}.}
    \label{tab:level0_performance}
\end{table*}

\section{General Dynamic Graph Understanding}
\label{sec:level_0_results}
Before investigating temporal motifs, we first conducted a set of foundational experiments on dynamic graphs to assess whether LLMs can correctly interpret our quadruplet representation. We designate these as Level 0 tasks. Unless otherwise specified, we use a default data generation setting of $N=10, T=5,$ and $p=0.3$ to generate 100 instances for each task. The corresponding results are presented in Tables~\ref{tab:level0_performance} through~\ref{tab:level0_qwq32b}, and our observations are detailed below.

\noindent\textbf{Level 0: Fundamental Dynamic Graph Understanding}

\begin{itemize}
    \item \noindent\textbf{Sort Edge.} Requires the model to chronologically sort a shuffled sequence of temporal edge events. The relative order of events with identical timestamps is ignored during evaluation.
    
    \item \noindent\textbf{When Link and Dislink.} Asks the model to identify the first "link" (add) and "dislink" (delete) timestamps for a given node pair. The query only considers direct connections and is sampled to ensure both event types exist.
    
    \item \noindent\textbf{What Edges.} Tasks the model with identifying all active edges at a query timestamp $t$. An edge is considered active if its most recent event at or before $t$ was an "add" operation.
    
    \item \noindent\textbf{Reverse Graph.} Requires the model to reverse the graph's temporal evolution by swapping the operation type for every event ("add" becomes "delete" and vice-versa). Evaluation uses a chronological sort with a tie-breaking rule that processes additions before deletions.
\end{itemize}

\noindent\textbf{Observation 1: LLMs possess a fundamental understanding of dynamic graphs, with performance dictated by model scale and characteristics.} As shown in Table~\ref{tab:level0_performance}, all models achieve competent results on fundamental tasks such as `When Link and Dislink` and `Sort Edge`, confirming their ability to comprehend temporal order and graph evolution. Performance is primarily influenced by two factors: (1) \textbf{Model Scale}, with larger models showing a stronger ability to handle complex logic, and (2) \textbf{Model Characteristics}, where differences in training and architecture cause significant performance gaps, even among models of the same 32B scale (DeepSeek-R1-Distill-Qwen-32B, Qwen2.5-32B-Instruct, and QwQ-32B).

\begin{figure}[htbp]
    \centering
    \includegraphics[width=1.0\columnwidth]{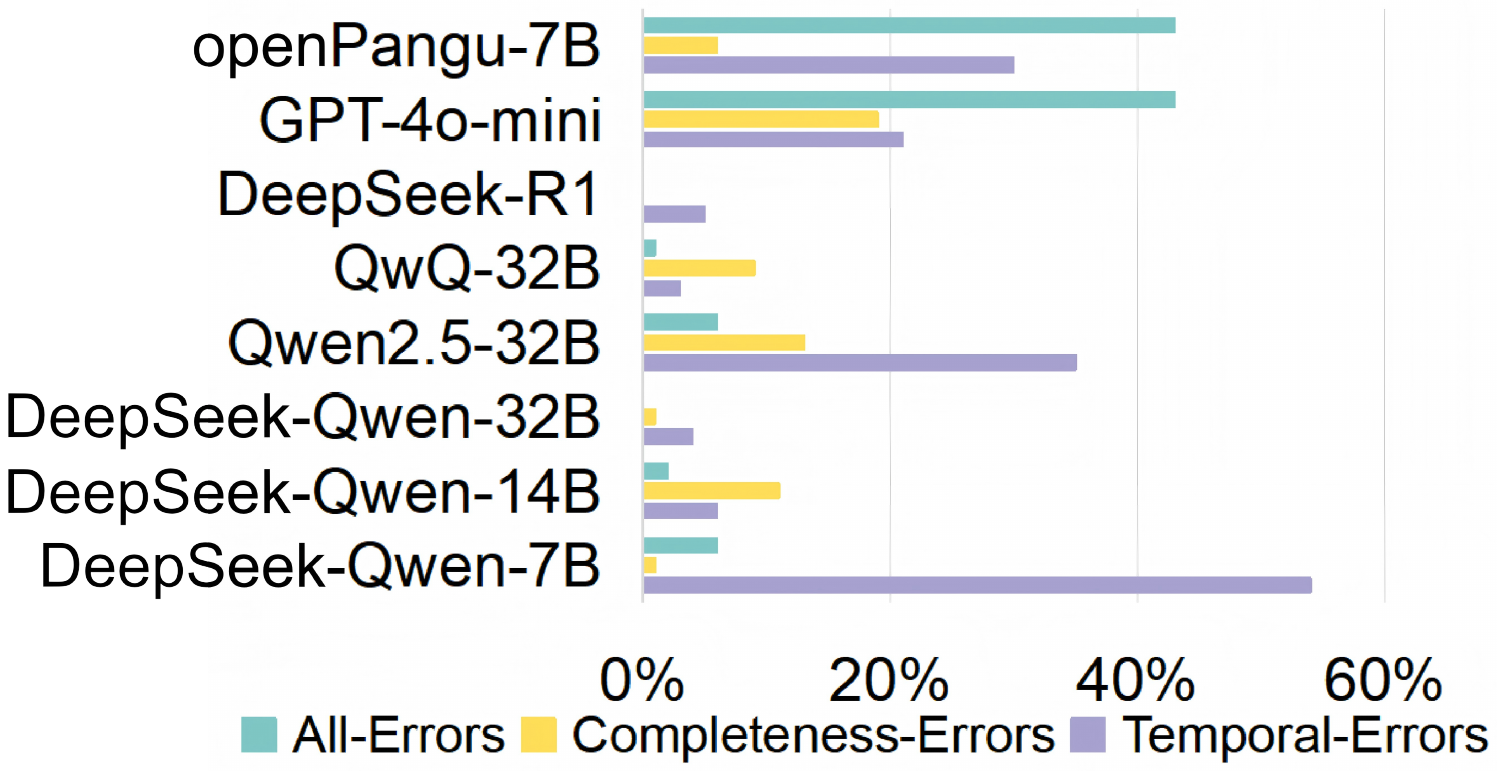}
    \caption{A fine-grained error analysis for the "Sort Edge" task, breaking down failures into three categories: temporal errors, completeness errors, and all errors.}
    \label{fig:error}
\end{figure}

\noindent\textbf{Observation 2: The core bottleneck for LLMs is processing global information across long sequences.} This is supported by a fine-grained error analysis of the “Sort Edge” task (Figure~\ref{fig:error}), where "temporal errors" are the primary failure mode. While models exhibit high completeness by recalling most individual events, their limitations in handling long-range dependencies prevent them from maintaining the strict logical consistency required for global tasks, such as constructing a perfect timeline. This bottleneck also explains why nearly all models excel on the “When Link and Dislink” task (Table~\ref{tab:level0_performance}), as it demands only local reasoning and thus bypasses this fundamental limitation.

\begin{table*}[t]
    \centering
    \small
    \setlength{\tabcolsep}{5pt}
    \begin{tabular}{@{}lcccc@{}}
        \toprule
        \textbf{DeepSeek-R1-Distill-Qwen-7B} & 
        \textbf{Sort Edge (N=5)} & 
        \textbf{When Link and Dislink} & 
        \textbf{What Edges} & 
        \textbf{Reverse Edge (N=5)} \\ 
        \midrule
        Zero-shot                 & 39\%             & 57\%             & \underline{34\%} & 59\%             \\
        One-shot                  & \underline{47\%}             & \underline{58\%}             & 27\%             & \underline{73\%}             \\
        Zero-shot+CoT             & 38\%             & 52\%             & 33\%             & 60\%             \\
        One-shot+CoT              & \textbf{67\%}            & \textbf{76\%}          & \textbf{40\%}          & \textbf{95\%}            \\ 
        \bottomrule
    \end{tabular}
    \caption{Performance of different prompting strategies on Level 0 tasks for the DeepSeek-R1-Distill-Qwen-7B model. For each task (column), the best performance is in \textbf{bold}, and the second best is \underline{underlined}.}
    \label{tab:level0_qwen7b}
\end{table*}

\begin{table*}[t]
    \centering
    \small
    \setlength{\tabcolsep}{5pt}
    \begin{tabular}{@{}lcccc@{}}
        \toprule
        \textbf{DeepSeek-R1-Distill-Qwen-14B} & 
        \textbf{Sort Edge} & 
        \textbf{When Link and Dislink} & 
        \textbf{What Edges} & 
        \textbf{Reverse Edge} \\ 
        \midrule
        Zero-shot                 & 80\%             & 90\%             & \underline{94\%}             & \underline{84\%}             \\
        One-shot                  & \underline{89\%}             & 95\%             & \underline{94\%}             & 77\%\textsuperscript{\emph{a}} \\
        Zero-shot+CoT             & 88\%             & \underline{98\%}             & 90\%             & 77\%             \\
        One-shot+CoT              & \textbf{96\%}            & \textbf{100\%}         & \textbf{98\%}          & \textbf{98\%}            \\ 
        \bottomrule
    \end{tabular}
    \begin{flushleft}
    \end{flushleft}
    \caption{Performance of different prompting strategies on Level 0 tasks for the DeepSeek-R1-Distill-Qwen-14B model. For each task (column), the best performance is in \textbf{bold}, and the second best is \underline{underlined}.}
    \label{tab:level0_qwen14b}
\end{table*}

\begin{table*}[t]
    \centering
    \small
    \setlength{\tabcolsep}{5pt}
    \begin{tabular}{@{}lcccc@{}}
        \toprule
        \textbf{DeepSeek-R1-Distill-Qwen-32B} & 
        \textbf{Sort Edge} & 
        \textbf{When Link and Dislink} & 
        \textbf{What Edges} & 
        \textbf{Reverse Edge} \\ 
        \midrule
        Zero-shot                 & 92\%             & \textbf{100\%}         & \underline{93\%}             & 94\%             \\
        One-shot                  & \underline{97\%}             & \textbf{100\%}         & \underline{93\%}             & \underline{99\%}             \\
        Zero-shot+CoT             & 95\%             & \textbf{100\%}         & 91\%             & 92\%             \\
        One-shot+CoT              & \textbf{98\%}            & \textbf{100\%}         & \textbf{97\%}          & \textbf{100\%}           \\ 
        \bottomrule
    \end{tabular}
    \caption{Performance of different prompting strategies on Level 0 tasks for the DeepSeek-R1-Distill-Qwen-32B model. For each task (column), the best performance is in \textbf{bold}, and the second best is \underline{underlined}.}
    \label{tab:level0_qwen32b}
\end{table*}

\begin{table*}[t]
    \centering
    \small
    \setlength{\tabcolsep}{5pt}
    \begin{tabular}{@{}lcccc@{}}
        \toprule
        \textbf{Qwen2.5-32B-Instruct} & 
        \textbf{Sort Edge} & 
        \textbf{When Link and Dislink} & 
        \textbf{What Edges} & 
        \textbf{Reverse Edge} \\ 
        \midrule
        Zero-shot                 & 46\%             & 93\%             & 38\%             & 50\%             \\
        One-shot                  & \underline{54\%}             & \underline{99\%}             & 42\%             & \textbf{76\%}          \\
        Zero-shot+CoT             & 45\%             & \textbf{100\%}         & \underline{56\%}             & \underline{61\%}             \\
        One-shot+CoT              & \textbf{63\%}            & \underline{99\%}             & \textbf{80\%}          & 59\%             \\ 
        \bottomrule
    \end{tabular}
    \caption{Performance of different prompting strategies on Level 0 tasks for the Qwen2.5-32B-Instruct model. For each task (column), the best performance is in \textbf{bold}, and the second best is \underline{underlined}.}
    \label{tab:level0_qwen2.5_32b}
\end{table*}

\begin{table*}[t]
    \centering
    \small
    \setlength{\tabcolsep}{5pt}
    \begin{tabular}{@{}lcccc@{}}
        \toprule
        \textbf{QwQ-32B} & 
        \textbf{Sort Edge} & 
        \textbf{When Link and Dislink} & 
        \textbf{What Edges} & 
        \textbf{Reverse Edge} \\ 
        \midrule
        Zero-shot                 & 87\%             & \textbf{100\%}         & 93\%             & 94\%             \\
        One-shot                  & 90\%             & \textbf{100\%}         & 93\%             & \underline{97\%}             \\
        Zero-shot+CoT             & \underline{94\%}             & \textbf{100\%}         & \underline{97\%}             & 92\%             \\
        One-shot+CoT              & \textbf{95\%}            & \textbf{100\%}         & \textbf{98\%}          & \textbf{100\%}           \\ 
        \bottomrule
    \end{tabular}
    \caption{Performance of different prompting strategies on Level 0 tasks for the QwQ-32B model. For each task (column), the best performance is in \textbf{bold}, and the second best is \underline{underlined}.}
    \label{tab:level0_qwq32b}
\end{table*}

\end{document}